%% file: main.tex
\documentclass[lettersize,journal]{IEEEtran}

\ifCLASSINFOpdf
\else
\fi

\usepackage[numbers,sort&compress]{natbib}
\usepackage[normalem]{ulem}
\usepackage{amssymb}
\usepackage{amsmath}
\usepackage{longtable}
\usepackage{array}
\usepackage{booktabs}
\usepackage{calc}
\usepackage[hidelinks]{hyperref}
\usepackage{xurl}
\usepackage{graphicx}
\usepackage{soul}
\usepackage{textcomp}
\usepackage{rotating}
\usepackage{tabularx}
\newcolumntype{L}{>{\raggedright\arraybackslash}X}
\newcolumntype{P}[1]{>{\raggedright\arraybackslash}p{#1}}
\newcolumntype{Y}{>{\raggedright\arraybackslash}X}
\usepackage{xcolor}
\usepackage{gensymb}
\usepackage{standalone}
\usepackage{fontawesome5}
\usepackage{multirow}
\usepackage{tikz}
\usepackage[inkscapelatex=false]{svg}
\usetikzlibrary{arrows.meta}
\usetikzlibrary{shapes.geometric, arrows.meta, positioning, calc, patterns, decorations.pathreplacing}

\usepackage[dvipsnames]{xcolor}

\DeclareRobustCommand{\statusYes}{\textcolor{ForestGreen}{\faCheckCircle}}
\DeclareRobustCommand{\statusNo}{\textcolor{red!70!black}{\faTimesCircle}}
\DeclareRobustCommand{\statusPartial}{\textcolor{orange!90!black}{\faAdjust}}

\providecolor{RevRed}{RGB}{198,32,40}
\definecolor{patchA}{RGB}{74,144,217}
\definecolor{patchB}{RGB}{232,168,56}
\definecolor{kept}{RGB}{46,139,87}
\definecolor{cropred}{RGB}{226,85,85}
\definecolor{patchC}{RGB}{91,174,91}
\definecolor{patchD}{RGB}{199,90,155}

\definecolor{trainblue}{RGB}{74,144,217}
\definecolor{testred}{RGB}{226,85,85}
\definecolor{foldA}{RGB}{74,144,217}
\definecolor{foldB}{RGB}{232,168,56}
\definecolor{foldC}{RGB}{91,174,91}
\definecolor{foldD}{RGB}{199,90,155}
\definecolor{goodgreen}{RGB}{46,139,87}
\definecolor{midorange}{RGB}{255, 143, 0}

\definecolor{accent}{RGB}{80,80,180}
\definecolor{boxfill}{RGB}{245,245,248}
\definecolor{boxstroke}{RGB}{200,200,210}
\definecolor{arrowgray}{RGB}{180,180,190}
\definecolor{subtextgray}{RGB}{130,130,140}

\usepackage{float}

\usepackage[most]{tcolorbox}

\input{boxes}

\begin{document}
\bstctlcite{IEEEexample:BSTcontrol}

\title{From Machine Learning to Large-Scale EO products: Good Practices and Guidelines for Making Maps}

\author{
  Ghjulia Sialelli,
  Robin Young,
  Cesar Aybar,
  Yuchang Jiang,
  Linus Scheibenreif,
  Damien Robert,
  Clemens Mosig,
  Kristof Van Tricht,
  Mikolaj Czerkawski,
  Nandika Tsendbazar,
  Adam J. Stewart,
  Jan D. Wegner,
  Aleksis Pirinen,
  Olof Mogren,
  and Konrad Schindler
\thanks{The work of G.~Sialelli was supported by an ETH AI Center doctoral fellowship. The work of D.~Roberts was supported by the EU Horizon Europe program under Grant Agreement No.~101131841 (Embed2Scale), with additional funding from SERI and UKRI.}
  \thanks{Corresponding author: G. Sialelli (gsialelli@ethz.ch) is with the Photogrammetry and Remote Sensing group, ETH Zurich, 8093 Zurich Switzerland and with the ETH AI Center, 8092 Zurich, Switzerland.}
  \thanks{R. Young is with the Dept.\ of Computer Science and Technology, University of Cambridge, UK.}
  \thanks{C. Aybar and M. Czerkawski are with Asterisk Labs, UK.}
  \thanks{L. Scheibenreif, and K. Schindler are with the Photogrammetry and Remote Sensing group, ETH Zurich, Switzerland.}
  \thanks{Y. Jiang is with Land Change Science, Swiss Federal Research Institute WSL, Birmensdorf, Switzerland.}
  \thanks{D. Robert and J. D. Wegner are with the EcoVision Lab, Dept.\ of Mathematical Modeling and Machine Learning, University of Zurich, Switzerland.}
  \thanks{C. Mosig is with the Institute for Earth System Science and Remote Sensing, Leipzig University, Germany.}
  \thanks{K. Van Tricht is with the Vlaamse Instelling voor Technologisch Onderzoek (VITO), Belgium.}
  \thanks{N. Tsendbazar is with the Laboratory of Geo-information Science and Remote Sensing at Wageningen University, Netherlands.}
  \thanks{A. J. Stewart is with the Chair of Data Science in Earth Observation, Technical University of Munich, Germany.}
  \thanks{A. Pirinen and O. Mogren are with RISE Research Institutes of Sweden, the Swedish Centre for Impacts of Climate Extremes, and Climate AI Nordics.}
}

\maketitle

\begin{abstract}
The past decade has seen an explosion of large-scale geospatial products derived from Earth observation (EO) data using machine learning (ML). Driven by expanding model capacity and compute availability, the domain has reached a level of operational maturity where practitioners can generate high-resolution global products spanning thematic areas as diverse as human settlement, vegetation structure, and natural hazards. Deep learning has expanded the paradigm beyond locally calibrated models toward singular, globally applicable models trained on heterogeneous data, attracting a broad and growing community of practitioners from both the EO and ML sides. Yet, the resulting workflow introduces challenges that fall between established EO practice and mainstream ML: every stage presents tightly coupled technical challenges that recur across applications, and much of the practical knowledge needed to navigate them remains siloed within individual application domains. This guide brings together map-makers across the EO and ML communities, distilling their common practices, recurring pitfalls, and recommended approaches into a single end-to-end account, from raw satellite data to a published map product. Crucially, it treats the pipeline as an integrated workflow rather than a sequence of independent steps. We focus on wall-to-wall raster maps at moderate resolution (10–30m) and continental to global extent, produced with ML from open satellite archives, mainly Sentinel-1, Sentinel-2 and Landsat.  The article is written for a deliberately broad audience, from those just entering the EO domain to ML experts building ML-ready EO datasets or generating large-scale maps.
\end{abstract}


\section{Introduction}
\label{sec:introduction}

Maps, of varying scales and extents, are accessible and powerful tools for policy-makers and other stakeholders to visualize spatial patterns, and inform planning and decision making. Forest carbon claims made under the UNFCCC REDD+ mechanism rest on mapped estimates of extent and change \citep{Nasimiyu2026}. The humanitarian sector relies on rapidly generated maps to assess damage extent \citep{UNOSAT_2024_gaza_damage}, and on population maps to determine which communities are at risk \citep{Metzger2024}.

The past decade has seen a rapid expansion in the production of large-scale geospatial products derived from Earth observation (EO) data using machine learning (ML). Driven by expanding model capacity and compute availability, the domain has reached a level of operational maturity where practitioners can routinely generate high-resolution global products, including maps of building footprints \citep{glazer2025tempo}, land cover \citep{Brown_2022}, canopy height \citep{lang2023canopyheight,Tolan2024}, renewable energy installations \citep{microsoft2024renewables} and flood inundation \citep{Misra2025}. The traditional EO toolkit (spectral indices, radiative transfer and other physical models, spectral unmixing, and classical ML pipelines built on hand-crafted features) historically delivered maps by leveraging locally calibrated models tuned to specific regions, sensors, or biomes; global products created this way were mosaics of local solutions. Deep learning has expanded the paradigm toward singular, globally applicable models trained on heterogeneous data, promising consistent worldwide products but also introducing challenges with no clear analog in the traditional workflow. Scaled inference, for instance, now demands specific engineering skills (distributed compute, efficient I/O, cloud-native data formats) that traditional EO scientists rarely needed in the past. At the same time, the apparent ease with which such maps seem to be produced hides the complexity of the underlying pipeline, and the lowered barrier to entry has enabled a proliferation of map products released without adequate validation. Crucially, ``maps are being developed by different organizations across the research-policy-practice interface, [...] without a shared understanding of best practices'' \citep{Singh2026}.
When considering the full workflow for a global map, every stage of the process presents domain-specific, tightly coupled challenges that have no direct equivalent in conventional machine learning. The failures we most often see in published maps are not failures of any single stage in isolation, but rather mismatches; e.g., train/test splits that rely on Sentinel-2 tile grids, introducing spatial overlap and hidden data leakage. Most reviews and tutorials on ML for EO are organized around methods or applications \citep{zhang2016deep,zhu2017deep,Persello2022,tuia2025}. Individual stages have received dedicated attention \citep{olofsson2014good,stehman2019key,ploton2020spatialvalidation,meyer2022machine}, but a comprehensive account of the workflow as a whole is lacking, particularly for the steps between a trained model and a released product. 

This article is addressed to a broad audience of researchers and practitioners, from those just entering the EO domain to ML experts looking to build ML-ready EO datasets or generate large-scale maps. It presents an end-to-end account of the good practices, challenges, and common pitfalls that span the entire pipeline, from the collection of satellite data to a published, validated map product. Throughout the article, we refer to many published products; two of them act as a common thread. \textsc{WorldCereal} \citep{Van_Tricht_2023} is an operational, categorical cropland and crop-type mapping system: it consumes per-pixel multi-sensor time series, classifies each pixel independently with gradient-boosted trees, runs inference on CPU, releases its 2021 products through Zenodo and Earth Engine, and validates its cropland layer through design-based accuracy assessment against a probability sample. \textsc{Canopy height map (CHM)} refers to ECHOSAT \citep{pauls2026echosat}, the most recent of three canopy height maps from the same group; where ECHOSAT leaves a step undocumented, we take it from its predecessors \citep{pauls2024estimating,pauls2025capturing}. It is a research-oriented continuous regression product: it consumes multi-temporal image patches, is trained on GPU with encoder--decoder networks, is released through Earth Engine, and is evaluated against its own test set. We contrast their choices in boxes throughout the article.

It would be misleading to present map-making sequentially: acquire, preprocess, train, infer, validate, publish. Several constraints run backwards through the workflow. For example, to minimize domain shift, training data preparation should ideally align with the data sources and processing baselines available on the global inference platform. As such, Figure \ref{fig:overview} presents the workflow as a directed graph, emphasizing that choices made at each stage carry consequences for other ones, and awareness of these dependencies is one of the factors that separate a technically sound map from a misleading one.

\begin{figure*}[!t]
    \centering
    \resizebox{\textwidth}{!}{\input{overview.tex}}
    \caption{The mapping workflow as a constraint network. Solid arrows are the familiar forward flow; dotted arrows indicate that every stage is conditioned by the intended map product. Dashed arrows are constraints that run \emph{backwards} through the workflow, and are responsible for the most expensive and least recoverable mistakes in large-scale mapping: the training and inference data must match, or the model meets a distribution shift at deployment that it was never tested on; data must be acquired over the regions where validation is intended, since without inference there, there is nothing to validate against; the uncertainty the product has to report constrains which models are admissible; the compute budget available for deployment shapes model development; and the validation regions must be excluded from the training and model-selection sets.}
\label{fig:overview}
\end{figure*}

\section{Overview of the EO data infrastructure landscape}
\label{sec:eo-data-landscape}

Researchers and practitioners must navigate a complex and rapidly evolving ecosystem of satellite data sources and access platforms. The choices made in the early stages (which mission to use, which processing level to start from, and which platform to retrieve the data through) have downstream consequences that are easy to underestimate. This section maps the landscape of available EO data, and discusses the practical considerations of accessing them.

\begin{definition}{Processing levels, analysis-ready, ML-ready}
\term{Processing levels} How far a product is from the raw signal. Level~0 is unprocessed instrument data; Level~1 adds time referencing, calibration and geolocation (e.g., Sentinel-1 GRD, Sentinel-2 L1C top-of-atmosphere reflectance); Level~2 holds geophysical variables at the resolution and location of the Level~1 source (e.g., Sentinel-2 L2A surface reflectance, Landsat Collection~2 Level-2); Level~3 maps variables onto a uniform space-time grid, often by compositing (e.g., the Sentinel-1 monthly mosaics) \citep{nasa_levels}.
\term{Analysis-ready data (ARD)} Data processed to standardized physical specifications (calibrated, geometrically orthorectified, radiometrically corrected, with per-pixel quality masks and documented provenance) so that it can be used for analysis with a minimum of additional user effort.
\term{ML-ready} Data formatted into standardized tensors (e.g., spatial or temporal patches) and organized in high-throughput layouts to optimize I/O performance during model training.
\smallskip \\
While institutional standards (e.g., NASA, ESIP) often use \textbf{AI-ready} as a broader term encompassing FAIR metadata, licensing, and documentation, \textbf{ML-ready} specifically denotes the structural formatting required for efficient GPU ingestion.
\end{definition}

EO data spans a range of modalities and thematic focuses, presenting distinct trade-offs in spatial detail, temporal revisit, and access regimes. To ground the subsequent discussion of processing baselines and retrieval pipelines, Table~\ref{tab:eo-missions} establishes a concrete taxonomy of the missions that recur most often in large-scale mapping, covering both open-access constellations and commercial platforms, alongside their core specifications and typical applications.

\begin{table*}[!t]
\centering
\caption{Earth observation missions that recur in large-scale mapping.}
\label{tab:eo-missions}
\renewcommand{\arraystretch}{1.2}
\footnotesize
\begin{tabular}{@{}lllllll@{}}
\toprule
\textbf{Mission} & \textbf{Modality} & \textbf{Spatial res.} & \textbf{Revisit} & \textbf{Cov.} & \textbf{Active} & \textbf{Typical use cases} \\
\midrule
\multicolumn{7}{@{}l}{\textit{Open access}} \\[2pt]
Sentinel-2        & Optical        & 10--60\,m       & 5 days      & \faLeaf  & 2015-- & Land cover, agriculture \\
Landsat           & Optical        & 15--120\,m      & 8--18 days  & \faLeaf  & 1972-- & Land cover change \\
Sentinel-3        & Optical        & 300--1200\,m    & $<$2 days   & \faGlobe & 2016-- & Ocean color, SST \\
MODIS             & Optical        & 250--1000\,m    & 1--2 days   & \faGlobe & 1999-- & Vegetation, fire, albedo \\
VIIRS             & Optical        & 375--750\,m     & $\sim$1 day & \faGlobe & 2011-- & Nighttime lights, fire \\
Sentinel-1        & SAR, C-band    & 5--20\,m        & 6 days      & \faGlobe\textsuperscript{\faAsterisk} & 2014-- & Flood, oil spill mapping \\
ALOS PALSAR/-2    & SAR, L-band    & 6--100\,m       & 14--46 days & \faGlobe & 2006-- & Forest mapping \\
NISAR             & SAR, L/S-band  & 10--20\,m       & 12 days     & \faLeaf  & 2026-- & Land, forest, ice change \\
BIOMASS           & SAR, P-band    & 50--200\,m      & $\sim$17 days & \faLeaf & 2025-- & Forest biomass \\
GEDI              & LiDAR          & 25\,m footprint & Sparse      & $\pm$51.6\textdegree & 2019-- & Canopy height, biomass \\
ICESat-2          & LiDAR          & 11\,m footprint & 91 days     & \faGlobe & 2018-- & Ice sheet elevation \\
GRACE / GRACE-FO  & Gravimetry     & $\sim$300\,km   & 30 days     & \faGlobe & 2002-- & Groundwater, ice mass \\
Gaofen-1/6\textsuperscript{$\dagger$} & Optical & 16\,m & 2–4 days & \faGlobe & 2013– & Agriculture, land cover \\
\addlinespace
\multicolumn{7}{@{}l}{\textit{Commercial}} \\[2pt]
Planet (Dove)     & Optical        & 3--5\,m         & Daily\textsuperscript{$\ddagger$} & \faLeaf & 2017-- & Change detection \\
Planet (SkySat)   & Optical        & 0.5\,m          & \faCrosshairs & \faCrosshairs & 2016-- & Sub-daily site monitoring \\
Vantor (WorldView)& Optical        & 0.3--2\,m       & \faCrosshairs & \faCrosshairs & 2001-- & Change detection, defense \\
Airbus (Pl\'eiades/SPOT) & Optical & 0.3--6\,m       & \faCrosshairs & \faCrosshairs & 2002-- & Large-area mapping, 3D \\
ICEYE             & SAR, X-band    & 0.25--3\,m      & \faCrosshairs & \faCrosshairs & 2018-- & Flood, disaster response \\
Capella           & SAR, X-band    & 0.5--1.2\,m     & \faCrosshairs & \faCrosshairs & 2020-- & Infrastructure monitoring \\
TerraSAR-X / TanDEM-X & SAR, X-band & 1--18\,m       & 11 days     & \faCrosshairs & 2007-- & Elevation models, InSAR \\
\bottomrule
\end{tabular}

\vspace{1ex}
  \begin{minipage}{\textwidth}
    \footnotesize
    Revisit times assume full constellation availability; spatial resolution ranges reflect different acquisition modes or spectral bands. Coverage: \faGlobe\,=\,global (land \& ocean), \faLeaf\,=\,land surfaces and coastal areas only, \faCrosshairs\,=\,tasked/on-demand. \textsuperscript{\faAsterisk}\,The failure of Sentinel-1B induced significant data gaps, see Appendix~\ref{app:s1-gaps}. \textsuperscript{$\dagger$}\,Gaofen-1/6 is openly distributed at Level 1 only: published ARD covers only China and its surrounding areas. \textsuperscript{$\ddagger$}\,The Dove constellation provides near-daily coverage of the land surface; sub-daily revisit of a specific site requires tasking a different constellation.
  \end{minipage}

\end{table*}

The EO data access landscape has expanded significantly in recent years, but it is also fast-moving and not cleanly delineated: providers and platforms regularly restructure their offerings, pricing, and access policies, and the boundaries between ``canonical provider'', ``cloud re-distributor'', and ``compute platform'' are increasingly blurred. A unifying constraint across this landscape is \emph{data gravity}: modern EO archives have grown to a scale where even freely available data is often impractical to move, and the position of compute relative to data has become a primary design factor. This is the reason some large-scale EO workflows today rely on platforms that co-locate data and compute instead of directly downloading from canonical providers.

\textbf{Canonical EO data providers} maintain authoritative archives of specific satellite missions and datasets. The Copernicus Data Space Ecosystem (CDSE) serves as the primary distribution point for Sentinel missions and related European products, while also providing access to Copernicus Contributing Missions (CCM) to complement core open collections with commercial and national high-resolution imagery. The United States Geological Survey (USGS) provides access to Landsat observations. NASA Earthdata hosts MODIS, GEDI, ICESat-2, and other NASA Earth science products. The European Organisation for the Exploitation of Meteorological Satellites (EUMETSAT) distributes data from MSG, MTG, Sentinel-3, Sentinel-6, and related systems. The Alaska Satellite Facility (ASF) offers NISAR, the ERS, JERS-1 and RADARSAT-1 legacy missions, and terrain-corrected Sentinel-1 products. Additional canonical providers include the JAXA Earth Observation Center for ALOS PALSAR, DLR Geoservice for TanDEM-X, and various other agencies maintaining specialized datasets on dedicated platforms.

\textbf{Re-distribution platforms} have emerged to simplify this access, though their roles vary considerably. Google Earth Engine (GEE) \citep{gorelick2017google} is best understood as a compute ecosystem with co-located data rather than a pure data provider. Access is primarily through server-side operations and the platform is highly effective for tasks that exploit this paradigm, including extracting moderate-scale ML training datasets, a common use in the community. However, large-scale data extraction is not what GEE is built for, and users quickly run into export quotas and rate limits on the free and non-commercial tiers. While commercial tiers relax these limits, bulk exports require routing through Google Cloud Storage and incur significant egress fees. This presents a key structural barrier for academic and open-science teams, who produce the vast majority of public EO benchmark datasets, as few research budgets can accommodate high-throughput cloud infrastructure and egress costs. Microsoft Planetary Computer (MPC), by contrast, is essentially a data access layer: a STAC catalog of cloud-hosted Cloud-Optimized GeoTIFFs (COGs) in Azure Blob Storage, with no built-in server-side processing since the retirement of its free compute environment in mid-2024.\footnote{\url{https://github.com/microsoft/PlanetaryComputer/discussions/347}} Amazon Web Services (AWS) Simple Storage Service (S3) hosts EO data on both free buckets (e.g., Cloud-Optimized GeoTIFF collections for Sentinel-2 L2A and Landsat 8/9 via the AWS Open Data Program) and requester-pays buckets (e.g., the primary USGS Landsat Collection 2 archive or complete Sentinel-2 SAFE zips, imposing transfer costs directly on the user). Additional platforms include Sentinel Hub, Google Cloud Storage public datasets, and CloudFerro\footnote{CloudFerro currently acts as the canonical Copernicus data provider by providing infrastructure supporting the CDSE platform.}. Relying on re-distribution platforms introduces several trade-offs. Since these platforms ingest data from canonical EO providers, there may be latency between acquisition and availability, and their processing pipelines may differ from those of the original providers. As an example, we provide an overview of the processing differences between the native Sentinel-1/2 data products and their GEE equivalents in Appendix~\ref{app:processing-pathways}. Data gaps can also occur, as regularly reported by users on the MPC \href{https://github.com/microsoft/PlanetaryComputer/issues}{GitHub repository}.

\begin{table*}[!t]
\centering
\caption{Overview of major Earth observation data providers.}

\label{tab:eo-providers}
\renewcommand{\arraystretch}{1.2}
\footnotesize

\begin{tabularx}{\textwidth}{@{} p{4.2cm} X c c c @{}}
\toprule
\textbf{Provider} & \textbf{Host} & \textbf{Online Ops} & \textbf{Bulk Egress} & \textbf{Cloud Compute} \\
\midrule
\addlinespace
\multicolumn{5}{@{}l}{\textit{Canonical providers}} \\[2pt]

\href{https://dataspace.copernicus.eu/data-collections}{Copernicus Data Space Ecosystem} & CloudFerro (PL, DE), T Cloud (DE, NL, CH) & \statusYes & \statusYes & \statusPartial \\
\href{https://www.earthdata.nasa.gov/data/catalog}{NASA Earthdata} & Amazon S3 (\texttt{us-west-2}) & \statusNo & \statusYes & \statusYes \\
\href{https://www.usgs.gov/centers/eros/data-access}{USGS} & Amazon S3 (\texttt{us-west-2}) & \statusNo & \statusYes & \statusYes \\
\href{https://user.eumetsat.int/catalogue}{EUMETSAT} & Own data centers, Amazon S3 & \statusNo & \statusYes & \statusPartial \\
\addlinespace
\multicolumn{5}{@{}l}{\textit{Redistribution platforms}} \\[2pt]
\href{https://planetarycomputer.microsoft.com/catalog}{Microsoft Planetary Computer} & Azure Blob Storage (NL) & \statusNo & \statusYes & \statusYes \\
\href{https://docs.cloud.google.com/storage/docs/public-datasets}{GCS public datasets} & Google Cloud Storage (various) & \statusNo & \statusYes & \statusYes \\
\href{https://cloudferro.com/data-tools/eo-data-catalogue/}{CloudFerro} & Own data centers & \statusPartial & \statusYes & \statusYes \\
\href{https://source.coop/products}{Source Cooperative} & Amazon S3 (\texttt{us-west-2}) & \statusNo & \statusYes & \statusYes \\
\href{https://registry.opendata.aws/}{AWS Open Data} & Amazon S3 (mostly \texttt{us-west-2}) & \statusNo & \statusYes & \statusYes \\
\href{https://developers.google.com/earth-engine/datasets/catalog}{Google Earth Engine} & Google Cloud Storage* & \statusYes & \statusNo & \statusPartial \\
\href{https://collections.sentinel-hub.com/}{Sentinel Hub} & Amazon S3 (\texttt{eu-central-1}) & \statusYes & \statusNo & \statusPartial \\
\bottomrule
\end{tabularx}

\vspace{1ex}
  \begin{minipage}{\textwidth} 
    \footnotesize
    \textbf{Host}: cloud infrastructure where the data resides, with geographic region when documented. \textbf{Online Ops}: whether server-side processing such as multi-scene compositing or cloud-masking is exposed through an API. \textbf{Bulk Egress}: feasibility of bulk archive download, \statusYes~= supported on the standard tier modulo cloud egress costs; \statusPartial~= requires commercial or institutional arrangements beyond standard egress; \statusNo~= the access API is not architecturally suited to bulk download regardless of budget. \textbf{Cloud Compute}: feasibility of running very large workloads such as deep learning inference at planetary resolution on infrastructure adjacent to the data, \statusYes~= the data resides on a hyperscaler (AWS, Azure, or Google Cloud Platform) offering large GPU/TPU pools in the same region; \statusPartial~= compute is available on the underlying infrastructure but does not reach hyperscaler scale, or the path between data and compute is interrupted by mandatory exports. *GEE data is not directly user-accessible at the storage layer; access requires the GEE platform's own constrained compute environment.
  \end{minipage}

\end{table*}

Accessing EO data therefore involves practical choices across multiple dimensions. Each platform offers distinct trade-offs in where the data is hosted, support for server-side operations, suitability for large-scale download, and feasibility of running large-scale co-located compute; we summarize them in Table~\ref{tab:eo-providers}. A practical pattern that follows from data gravity is to treat EO data access as a two-stage process. \emph{Model development} typically involves modest data volumes that can be downloaded comfortably from almost any provider; here, convenience and catalog coverage dominate the choice. \emph{Inference at scale}, on the other hand, is dominated by data movement: pulling petabyte-scale imagery from one provider into a different compute environment is rarely tractable. As a result, practitioners typically either (i) co-locate inference compute with the data, or (ii) accept the time and storage cost of moving the EO data into their compute environment. Co-location only helps if the compute you need is available where the data is: large GPU pools are a hyperscaler feature, so GPU-bound inference is restricted to data hosted on AWS, Azure or GCP; CPU-bound pipelines have more options. The platform that minimizes friction for training is therefore often not the one that minimizes cost for inference, and moving between them introduces a risk of distribution shift that should not be taken lightly. No universally optimal solution exists: the most appropriate choice depends on the scale of analysis, computational requirements, institutional constraints, the practical cost of download and storage, and the acceptable trade-off between immediate convenience and the risk of lock-in to proprietary platforms.

\begin{decision}{Provider}
\textbf{Alternatives.} A canonical EO data provider vs. a re-distribution platform.\\[2pt]
\textbf{Trade-off.} Canonical providers give provenance certainty; re-distributors give convenience, but introduce silent processing divergence. Co-located archives and compute remove the transfer bottleneck but restrict access to what they host.\\[2pt]
\textbf{Constrains.} The processing levels available for each provider; egress cost and lock-in. Because switching provider between training and inference introduces a distribution shift, this decision belongs at the start of the project rather than at the inference stage.
\end{decision}

\begin{runex}{Provider and compute}
\textbf{WorldCereal} has been produced twice. For the first phase (the 2021 global products), Sentinel-1, Sentinel-2 and thermal Landsat~8 time series, the Copernicus Digital elevation model (DEM) and Ecoregions2017 all came from AWS, and the pipeline ran there entirely, on CPU. Keeping compute beside the archive made iteration fast and inference cheap; most of the compute budget ($\approx$100k EUR) went into preprocessing the hundreds of terabytes of imagery. The second phase moved to CDSE and openEO.\\[3pt]
\textbf{CHM} used Sentinel-1, Sentinel-2, ALOS PALSAR-2, and the TanDEM-X DEM and Forest/Non-Forest maps. The $\approx$ 50 TB of data was downloaded from AWS and Google Earth Engine, to a university server (with 2\,PiB of local storage available). Training required GPUs, from the university HPC cluster. Inference then ran for five months. There were no extra costs. \\[3pt]
Both drew mainly on the Sentinel-1 and Sentinel-2 archives; they differed in where the computation ran, and in what it cost.
\end{runex}

\bigskip
\noindent

\section{Data selection and preprocessing}
\label{sec:all-things-data}

The quality of any ML-derived map is bounded by the quality of the data that feeds it. This section addresses the two foundational stages of the data pipeline: selecting which observations to include, and processing them into a form suitable for analysis. Data selection (Section \ref{subsec:data-selection}) involves navigating trade-offs between spatial coverage, temporal density, and data quality. These decisions are particularly consequential when working with global-scale EO archives where gaps, overlaps, and redundancies must all be managed explicitly. Data preprocessing (Section \ref{subsec:data-processing}) covers the transformation of sensor measurements into analysis-ready data, a step whose apparent simplicity conceals a number of subtleties. Throughout, we emphasize that these choices are not neutral: they encode assumptions that propagate into the training signal and ultimately shape the reliability of the final product.

\subsection{Data selection}
\label{subsec:data-selection}

Selecting which observations to include requires navigating two interrelated concerns: the temporal distribution of acquisitions (Section~\ref{subsubsec:temporal-distribution}) and the quality criteria for both optical (Section~\ref{subsubsec:optical}) and SAR data (Section~\ref{subsubsec:sar}). They involve trade-offs that are straightforward to state but difficult to resolve in practice, particularly at global scale.

\subsubsection{Temporal distribution}
\label{subsubsec:temporal-distribution}

We frame the present discussion around the task of producing a yearly map, a common context, which also requires addressing the challenge of temporal data selection. Under this setting, we discuss how to select EO data in the temporal dimension within a period of approximately 12 months. The conclusions can be extended to shorter or longer windows of time, depending on the nature of the underlying feature being mapped. Furthermore, this framing assumes that multiple temporal acquisitions are available; this might not be the case for high- and very-high-resolution sensors: the Maxar mosaic used by \citet{Tolan2024} and the SPOT mosaic underlying \citet{schwartz2025formspo} each rest on a single acquisition per location. Two questions emerge: how observations are represented in time, and what time range they are drawn from.

 Regarding representation, the first decision is whether the model sees observations at their original acquisition timestamps, or aggregated onto a fixed temporal grid. Three broad strategies emerge: single time-steps, time-series, or composites, each with its own trade-offs. \textbf{Time-series} are arguably the richest option, capturing phenomena like seasonal variation across a defined period. \citet{pauls2025capturing} showed that a one-year time-series made of the least cloudy Sentinel-2 observation per month preserved the seasonal nuances and geolocation shifts that median composites tend to wash away, and \citet{dixonCanopyHeightEstimation2025} found benefits in using a time-series of composited PlanetScope imagery over using a single image. The trade-off is data size and model capacity: they are memory-intensive to store by nature, and require a model handling temporal patterns. Furthermore, for each timestep in the time series comes the question of which observation to choose. In the case of optical data, most researchers opt for a time-series with a monthly temporal resolution, choosing the least cloudy product per month \citep{pauls2026echosat,herzog2025olmoearthstable,pauls2025capturing, schwartzRetrievingYearlyForest2025}. However, a pixel in the tropics and a pixel at high latitude will have different numbers of usable observations, depending on local weather patterns, so the model should be resilient to variable-length, irregularly spaced inputs. In some regions of the world, land observations of certain areas could be consistently underrepresented due to consistent high cloud coverage spanning across the same season over multiple years. \textbf{Composites} offer a pragmatic alternative, distilling an entire period of observations into the memory footprint of a single time-step, as in \citep{pauls2024estimating, schwartzFORMSForestMultiple2023a, suCanopyHeightBiomass2025, wanSatellitebasedMappingAnnual2025, schwartzHighresolutionCanopyHeight2024, favrichonMonitoringChangesForest2025}. Yet, compositing is not without its own costs: it requires deliberate design choices and non-trivial computations. Typical approaches consist of performing mean or median compositing, the latter involving computing percentiles, and the former being more sensitive to outliers. Moreover, if compositing cannot be performed on-the-fly before downloading the data, practitioners must store (at least temporarily) the time-series the composite will be computed from, undermining much of the storage advantage. While the per-pixel processing cost may appear modest in isolation, it scales dramatically: at a global scale, every additional step must be applied to each pixel on Earth, which can quickly become too costly in many settings. One can also produce a \textbf{time series of composites}, e.g.,\ monthly composites of Sentinel-2 \citep{herzog2025olmoearthstable}, seasonal composites \citep{Neumann_2025,glazer2025tempo,microsoft2024renewables,Van_Tricht_2023}, or using a predetermined moving window \citep{esa2021worldcover,potapov2021mapping,Mosig2026,dixonCanopyHeightEstimation2025}. Using \textbf{single time-steps}, by contrast, is the most cost-efficient route but comes at the expense of the temporal dimension, which often carries rich, discriminative information. Indeed, \citet{Mosig2026} found that performance steadily improved with access to longer input timespans when mapping tree mortality, and \citet{boudrasSERAHNativeSentinel2025} found similar results for canopy height mapping. The consequences of this choice will be felt at every subsequent stage: dataset construction, where samples may or may not share a common shape; model design, where the architecture must be able to
represent whatever sampling was chosen; and inference, where every pixel on Earth has to be served the same input structure.

Beyond the choice of temporal representation, the time range itself must be defined. Here, we focus on vegetation-related tasks, for which the trade-offs are particularly well documented, but the same considerations apply across domains. A common default is to restrict the analysis to the leaf-on season \citep{Mosig2026,pauls2024estimating,kacicForestStructureCharacterization2023,Liu_2023,schwartzRetrievingYearlyForest2025,langCountrywideHighresolutionVegetation2019}, but identifying that period for every pixel on Earth is in itself non-trivial. Operational phenology products offer one solution: \citet{Mosig2026} use the VIIRS Global Land Surface Phenology product \citep{Zhang2018}, while \citet{Liu_2023} use the MODIS equivalent \citep{Zhang2003}. Both products are released at annual intervals, so inter-annual shifts in phenological dates are in principle captured. However, the underlying algorithms assume a vegetation cycle with at most two growing cycles per year, an assumption that breaks down for multi-cropping systems with more than two cycles. \citet{Van_Tricht_2023} bypass this limitation by relying on global crop calendars instead \citep{Franch2022,MolettoLobos2026}. More fundamentally, the annual leaf-on/leaf-off framing is itself a temperate-zone assumption that does not generalize to tropical systems \citep{Davis2022}. The leaf-on framing can also be limiting in its own right: \citet{pauls2025capturing} show that incorporating both summer leaf-on and winter leaf-off observations improves performance, and \citet{Karaman2025} find that for biomass estimation in Finland, early-spring Sentinel-2 acquisitions are more informative than summer ones, which they attribute to ``the higher contrast between vegetation and snow, which enhances the ability to distinguish vegetation properties''. This logic does not extend everywhere, but regions with strong seasonal contrast can benefit from year-round data. A more task-specific variant samples contrasting images from different times within the same year to expose intra-annual variation, as applied by \citet{kerner2025fields} for crop field segmentation. The calendar year itself is also a poor natural boundary: for example, in the Southern Hemisphere, summer spans the end of one year and the beginning of the next, so a global-scale pipeline must accommodate such cross-year shifts rather than assume that a single calendar year is a sufficient window. Practitioners therefore need to decide how to balance a simple, globally consistent temporal definition against locally appropriate temporal characterization of the modeled phenomenon.

\begin{decision}{How to represent time}
\textbf{Alternatives (when available).} Single acquisitions, time-series of single acquisitions, time-series of composites, composites. \\[2pt]
\textbf{Trade-off.} Compositing over a long enough time-span substantially increases the fraction of valid pixels, at the cost of some temporal information. Fixed time series deliver a fixed tensor for every pixel on Earth, not all of them necessarily valid.\\[2pt]
\textbf{Constrains.} Storage, download budget, whether compositing can run server-side prior to download, inference cost, model architecture.
\end{decision}

\begin{runex}{Temporal strategy}
\textbf{WorldCereal} uses crop calendars to define appropriate time windows across the world. For each region and window, time series of composites are produced.\\[3pt]
\textbf{CHM} leverages a yearly (defined over the calendar year) time-series of the least cloudy Sentinel-2 acquisition per month.
\end{runex}

\subsubsection{Optical data selection}
\label{subsubsec:optical}

Even after temporal selection, individual observations vary in usability. Two factors dominate optical imagery: cloud contamination and sensor-specific artifacts that introduce gaps or duplications. Rather than discussing the per-pixel quality flags shipped with each product (e.g.,\ the Sentinel-2 SCL band or Landsat QA band), we focus on higher-level quality criteria that govern whether entire products should be included or discarded.

Upon selecting which data product to work with, a common filter in optical imagery is the \textbf{cloud cover}. In the deep learning literature, whether clouds require explicit treatment at all is debated \citep{Ruwurm2020,Turkoglu2021,Gawlikowski2022}. We argue that this debate, however, does not extend to \emph{scene-level} selection: a largely obscured acquisition contributes few valid pixels while incurring the full storage cost; filtering acquisitions based on cloud-cover is therefore primarily a question of dataset economy rather than of model robustness. We elaborate on the various methods to perform cloud detection in Section~\ref{subsec:data-processing} and henceforth assume access to a reliable cloud mask. A common approach is to discard data products with a cloud cover above a certain threshold (e.g.,\ 10\% in \citep{pauls2024estimating}, 35\% in \citep{Brown_2022}, 60\% in \citep{Neumann_2025}, 75\% in \citep{muhawenayo2026pruepracticalrecipefield}, 90\% in \citep{kerner2025fields}). However, for regions with persistent cloud cover, naively filtering out entire data products based on their cloud cover percentage induces the risk of throwing away valuable non-cloudy pixels. For example, when compositing, products with high cloud cover may contain cloud-free areas that fill critical gaps in the composite. Alternatives to thresholding differ mainly in the granularity at which they operate. Per-pixel approaches such as Best Available Pixel (BAP) \citep{Francini2023,White2014} score every observation independently and retain the highest-scoring one at each pixel, so a clear pixel is kept regardless of its parent product's overall cloud fraction. Product-level approaches instead select a subset of acquisitions: for example, \citet{muhawenayo2026pruepracticalrecipefield} designed a Sentinel-2 product selection algorithm where products are chosen based on their contribution of valid pixels to under-represented spatial regions within a tile. Which granularity is practical depends in part on where compositing happens. When it can be pushed to the data provider, per-pixel selection is cheap, since rejected observations are never transferred. When data must be retrieved locally, per-pixel scoring requires downloading every candidate acquisition and discarding most of it; selecting a small subset of products becomes the only tractable option. A similar constraint applies to the masks themselves: computing or accessing a cloud mask at the stage of data product selection can be computationally expensive, since it must be done for each candidate product. Most optical data providers report cloud cover statistics in their products metadata, which explains the persistence of scene-level thresholding, despite its known failure mode.

\begin{decision}{Selecting optical products based on clouds}
\textbf{Alternatives.} Discard whole products above a cloud-cover threshold; compute a mask per candidate and select at pixel level; rank products by their contribution of valid pixels to the area still uncovered.\\[2pt]
\textbf{Trade-off.} Product-level thresholds are nearly free, since the statistic ships in the metadata, but they discard cloudy products that may hold the only clear pixels over part of their footprint. Pixel-level selection keeps those pixels and costs a mask computation for every candidate.\\[2pt]
\textbf{Constrains.} How many observations survive in persistently cloudy regions; compute available for cloud masks; number of products one can afford to download and store.
\end{decision}

\begin{runex}{Cloud screening}
\textbf{WorldCereal} did not perform any optical image selection, but masked cloud and shadow (using a variant of the binarized Sen2Cor scene classification mask) prior to compositing.\\[2pt]
\textbf{CHM} selected one image per calendar month based on the highest percentage of valid pixels (excluding cloudy and black pixels as identified by the Sen2Cor algorithm). In persistently cloudy regions, the selected image may still be substantially contaminated.
\end{runex}

Another important criterion for selecting optical data products is the occurrence of \textbf{artifacts}. For example, Landsat 7 imagery acquired after 2003 exhibits wedge-shaped gaps from the Scan Line Corrector failure, affecting roughly 22\% of each scene \citep{arvidson2006landsat,ju2008availability}. More generally, a subtler problem arises when the \textit{distribution unit} of a product is not aligned with its \textit{acquisition unit}. Sentinel-2 is the clearest example: the granule is a fixed tile, while the sampling unit is the orbit. Consequently, a tile may be covered partially by one orbit and fully by another, may require mosaicking several orbits acquired on different days, or be affected by other subtle failure cases. Appendix~\ref{app:s2-artifacts} illustrates such cases. Addressing these issues is important to avoid duplication of pixel values, which would bias any statistics computed over the data — whether during compositing or when calculating overall dataset statistics from, for example, all Sentinel-2 products acquired within a given time frame, as in \citep{brown2025alphaearthfoundations,feng2025tessera}.

\subsubsection{SAR data selection}
\label{subsubsec:sar}

\begin{figure}[!t]
\centering
\includegraphics[width=\columnwidth]{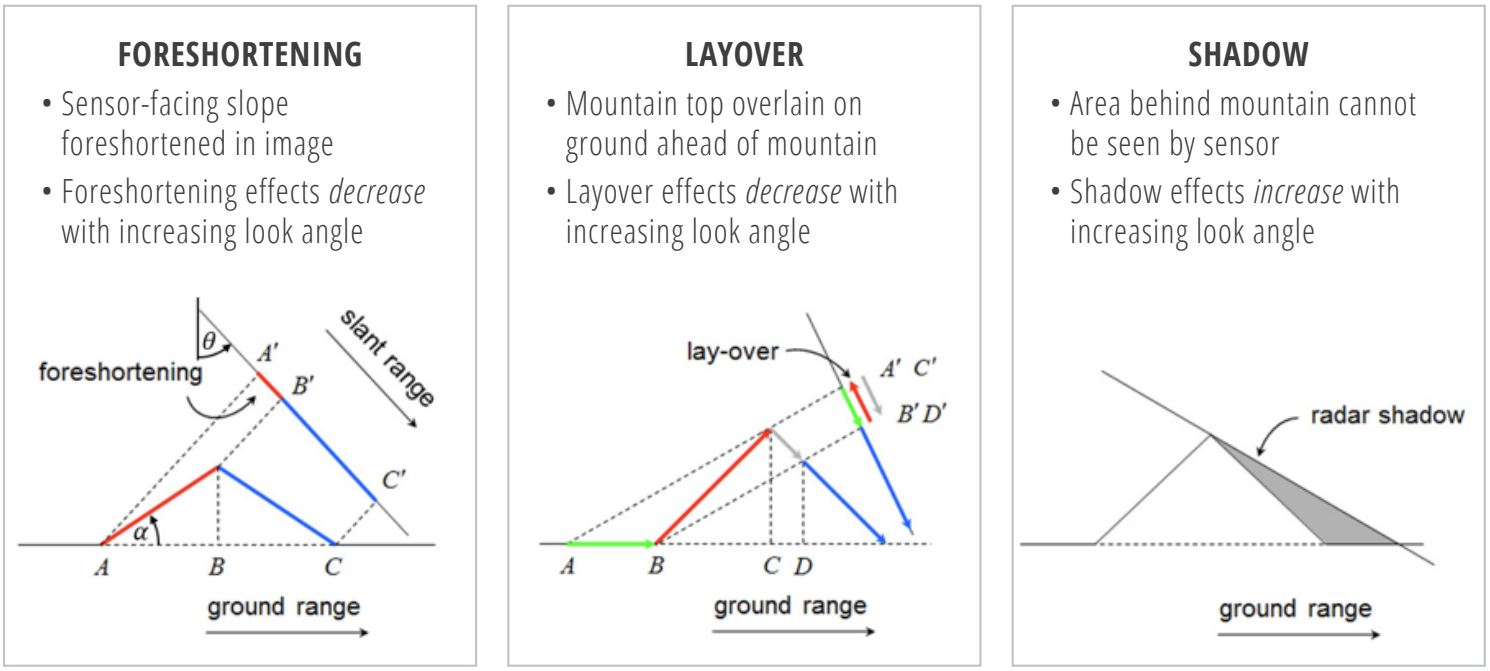}
\caption{Main geometric distortions on SAR images with their dependence on acquisition geometry. Figure from \citep{sarhandbook}.}
\label{fig:s1-distortions}
\end{figure}

SAR imagery is not affected by clouds, but its side-looking acquisition geometry introduces its own artifacts: geometric distortion due to foreshortening, layover, and shadow (see Figure \ref{fig:s1-distortions}). These deterministic perturbations arise from the interaction between terrain relief and the radar's look vector: given a digital elevation model and the acquisition geometry, the distortion pattern can be predicted. This predictability is useful at the selection stage: the sheer data volume and the need for geometric consistency over time argue for working with a subset of the data. Since the distortion associated with each candidate geometry can be computed in advance, that subset can be chosen analytically, before any data is downloaded. Appendix~\ref{app:sar-geometry} gives an example of this procedure for Sentinel-1.

Another common choice is that of the pass direction. While some regions (among them large portions of Africa, Australia, Russia, Canada and the eastern Americas) are systematically acquired in a single pass direction \citep{BauerMarschallinger2021}, when both ascending and descending acquisitions of the same area are available, they can be complementary. They should not, however, be treated interchangeably: unless corrected, the azimuthal dependency of $\sigma^0$ can cause major problems when comparing images acquired from ascending and descending passes \citep{Schaufler2018}. This has a direct impact on model design: global coverage favors models that accept missing modalities (Section~\ref{subsec:model-design}) over ones that assume a complete and fixed input stack.

\noindent

\subsection{Data preprocessing}
\label{subsec:data-processing}

Processing steps for Earth observation data vary considerably across the major data sources, but share common goals of converting sensor measurements into analysis-ready data formats. While specific steps differ between optical and SAR, and between missions within each, one constraint governs them all: choices made here shape the distribution of the training data. If any preprocessing choice differs when the model is run at scale, the resulting distribution shift might cause silent failure, with a plausible-looking map. Model evaluation cannot anticipate this; only validating the map against independent reference data can (Section~\ref{subsec:validation}). Another related constraint is availability: all processing steps are not available on all platforms. The practical question is therefore not only which processing is theoretically preferable, but which is appropriate enough for the application and available consistently at the scale required.

\subsubsection{SAR preprocessing}
\label{subsubsec:sar-processing}

Synthetic aperture radar (SAR) data requires a distinct and often more involved preprocessing chain than optical imagery. SAR sensors measure the coherent backscatter of microwave pulses, and the resulting images are affected by multiplicative speckle noise, radiometric variations driven by local terrain geometry, and geometric distortions inherent to the side-looking acquisition (as described in Section~\ref{subsubsec:sar}). Unlike optical sensors, typically delivered as atmospherically corrected products, SAR products such as Sentinel-1 Ground Range Detected (GRD) products are distributed at a relatively early processing stage and require several user-side corrections before they can be meaningfully compared across space and time.

\begin{runex}{Two routes through the speckle decision}
\textbf{WorldCereal} filters spatially: a Gamma-MAP filter with a kernel of 7 and three equivalent looks, applied per acquisition and chosen to preserve the original spatial resolution. The series is then composited to 12-day means, with missing values linearly interpolated. \\[3pt]
\textbf{CHM} filters temporally: no per-acquisition filter, with quarterly compositing doing the work instead. \\[3pt]
What separates both approaches is what each was willing to give up: a little spatial resolution in the first case, the individual acquisitions in the second.
\end{runex}

\paragraph{Processing steps} The corrections needed to convert GRD products into analysis-ready backscatter can be grouped into four stages. First, \textbf{thermal noise removal} suppresses additive noise artifacts that are particularly visible near swath edges and in cross-polarization channels. Second, \textbf{radiometric calibration} converts digital numbers to a physically meaningful backscatter coefficient by normalizing by a reference area; the conventions in use ($\sigma^0$, $\beta^0$, $\gamma^0$) differ in which area they assume, and in whether that area comes from an ellipsoid or from a digital elevation model (see Appendix~\ref{app:sar-rtc}). Third, \textbf{speckle reduction} addresses the multiplicative speckle noise produced by the nature of SAR imaging. Spatial filtering \citep{LeeFiltering,FrostFiltering,RefinedLee} gives up spatial resolution for radiometric smoothness; multi-temporal filtering \citep{Quegan2001} denoises an image using a stack of acquisitions at higher computational cost, preserving resolution; and temporal averaging, the strategy common in the SatML community \citep{pauls2025capturing, schwartzRetrievingYearlyForest2025}, preserves spatial detail but discards individual observations. Fourth, \textbf{geometric and radiometric terrain correction} resolves the distortions topography introduces. GRD products are projected onto an Earth ellipsoid and do not account for local relief, so Range-Doppler correction against a DEM is required to recover accurate geolocation, indispensable whenever SAR data is co-located with optical imagery over anything but flat ground. Radiometric terrain correction separately normalizes against the true illuminated area per pixel, without which intensity remains partly a function of slope rather than of ground properties (Section~\ref{subsubsec:sar}). Established toolboxes including pyroSAR \citep{pyrosar} and the ESA SNAP toolbox \citep{esasnap} provide implementations of these workflows.

\paragraph{Platform-specific processing} The absence of a standardized preprocessing pipeline for Sentinel-1 GRD data means that the choice of data provider is, in practice, inseparable from the preprocessing approach. Different platforms deliver Sentinel-1 data at different stages of processing, using different DEMs, backscatter conventions, and speckle treatments — differences that can propagate silently into downstream models. Details are provided in Appendix~\ref{app:s1-pathways}. Within the Copernicus Data Space Ecosystem (\textbf{CDSE}) \citep{cdses1}, data are available from GRD products to the more recent Sentinel-1 RTC product. \textbf{Google Earth Engine} applies its own processing to native GRD data; the reference implementation \citep{Mullissa2021} and its associated package \citep{gees1ard} provide guidance on proper analysis-ready data preparation on GEE. \textbf{AWS S3} hosts the GRD archive from CDSE in cloud-optimized GeoTIFF format \citep{aws3grd} as well as a Sentinel-1 RTC collection over the contiguous United States \citep{aws3rtc}. \textbf{Microsoft Planetary Computer} \citep{mpc} provides both a GRD collection from ESA and an RTC collection that they process themselves; for the former, they encourage users to apply geometric or radiometric terrain correction on the fly using any DEM available on the platform. For practitioners that would rather not maintain a SAR processing chain, two products deliver already processed SAR data: the Sentinel-1 Level-3 ARD monthly mosaics (Copernicus Data Space Ecosystem), global at 20 m, and the PALSAR/PALSAR-2 yearly mosaics (JAXA), global at 25 m. Both are free and both are delivered as terrain-corrected $\gamma^0$, requiring little user-side preprocessing. We compare them stage by stage in Appendix~\ref{app:sar-ard}.

\begin{decision}{How much of the SAR chain to perform}
\textbf{Alternatives.} Start from GRD and run the four corrections in-house; take a provider's partially processed product, where thermal noise removal, calibration and terrain correction are applied but speckle is not; or take a finished Level-3 mosaic, where compositing has handled speckle too. \\[2pt]
\textbf{Trade-off.} Owning the chain buys control over every convention and keeps individual acquisitions addressable, at the cost of building and maintaining it at scale. Inheriting it buys global consistency produced by people who know the sensor, at the cost of accepting their DEM, their normalization and their compositing rule. \\[2pt]
\textbf{Constrains.} Taking a composited product fixes the cadence, so the temporal decision of Section~\ref{subsubsec:temporal-distribution} is made in advance, and orbit direction and incidence angle are no longer selectable, foreclosing the choices of Section~\ref{subsubsec:sar}.
\end{decision}

\subsubsection{Optical preprocessing}
\label{subsubsec:optical-processing}

Optical data is typically distributed at two processing levels: top-of-atmosphere (TOA) radiance or reflectance, and bottom-of-atmosphere (BOA) reflectance after atmospheric correction. Most ML workflows operate on the latter (Sentinel-2 Level-2A, Landsat Collection 2 Level-2) but several steps remain before the data is ready for model ingestion.

\paragraph{Cloud masking} Cloud and shadow contamination affects all passive optical sensors. Because Sentinel-2 dominates the recent ML literature, most dedicated detection methods have been developed and benchmarked on it. Many such methods transfer to other optical sensors but may require retraining for different bands and resolutions. The available methods range from the Scene Classification Layer shipped with every Sentinel-2 L2A product to dedicated deep learning models, compared in Appendix~\ref{app:s2-cloud-masking}. The choice is largely governed by compute: SCL and, for GEE users, Cloud Score+ are already computed, while other methods must be run per product, and the deep learning methods that outperform traditional ones can be substantially more expensive. Furthermore, whether to apply cloud and cloud shadow masking at all depends on the intended use of the data. For workflows that temporally aggregate observations into composites, median aggregation already suppresses most cloud contamination, though masking cloudy and shadowed pixels beforehand generally yields cleaner composites and, for that purpose, the SCL layer is typically sufficient. For workflows that operate on individual scenes, the decision involves a trade-off between input quality and model robustness: aggressive masking produces cleaner inputs but discards data, while retaining cloudy pixels forces the model to learn to identify and handle these artifacts itself, consuming part of its representational capacity.

\paragraph{Radiometric conventions} Reflectance is distributed as scaled integers, and the scaling is a convention rather than a constant. Sentinel-2 is a cautionary example: since Processing Baseline 04.00, a radiometric offset is applied to all bands, and because the entire archive was reprocessed in the Collection-1 campaign, the offset must be determined from the \texttt{PROCESSING\_BASELINE} metadata field rather than the acquisition date. Some platforms apply the offset upstream, in which case it must not be applied again. The conversion equations and the divisors used in practice are given in Appendix~\ref{app:s2-radiometry}. The general point is that a scaling convention can change mid-archive, can differ between providers of the same nominal product, and produces no error when applied incorrectly, only a shifted input distribution.

\paragraph{Illumination corrections} Level-2 products still contain systematic reflectance variation unrelated to surface properties. The largest source is viewing and illumination geometry: pixels at swath edge are observed at a different angle than those at nadir, and sun position changes across orbits and dates, so the same ground pixel yields different values purely because of geometry. Bidirectional Reflectance Distribution Function (BRDF) normalization adjusts reflectance to a standard viewing and illumination configuration, producing smoother time series and reducing visible seam lines at tile boundaries, at the cost of surface structure assumptions and an additional preprocessing step. Implementations are available through Sentinel Hub \citep{sentinelhub}, the sen2nbar package \citep{Montero2024} and the FORCE framework \citep{force}. Landsat, and its harmonization with Sentinel-2, is covered in Appendix~\ref{app:combining-missions}.


\begin{runex}{Who performed the preprocessing}
\textbf{WorldCereal} ran its own chain. In the first phase, it started from Sentinel-2 L1C on AWS and applied atmospheric correction and cloud masking in-house, because the L2A archive was not yet consistent across the globe, and computed backscatter from Sentinel-1 GRD in the same pass. In the second phase, openEO on CDSE computes backscatter on the fly, but the chain remains theirs to specify. \\[3pt]
\textbf{CHM} inherited most of it. Sentinel-1 was taken from Google Earth Engine, which fixes the convention at $\sigma^0$ and the DEM at SRTM; ALOS PALSAR-2 was taken already terrain-corrected and composited. Only Sentinel-2 was processed from the archive directly. \\[3pt]
Neither choice was free. WorldCereal owned the chain because no available product met its consistency requirement; the canopy height group inherited it because running that chain over seven years of global data was not feasible for a university group. Circumstance settled it in both cases, not preference.
\end{runex}

\section{Machine learning dataset construction and model training}
\label{sec:ml}

With preprocessed data in hand, the focus shifts to the machine learning core of the pipeline. Everything up to this point has concerned analysis-ready data, whereas from here the concern is ML-ready data, in the sense fixed in Section~\ref{subsec:eo-data-landscape}. This section covers two stages: constructing an ML-ready dataset (Section \ref{subsec:ml-ready-dataset}) and designing and training the model (Section \ref{subsec:model-design}). Each stage raises challenges that are specific to the EO context, and the decisions made at these stages directly determine the quality of the pixel-level predictions.

\subsection{Constructing a machine-learning-ready dataset}
\label{subsec:ml-ready-dataset}

Constructing an ML-ready dataset from EO imagery introduces challenges that have no direct analog in standard computer vision: spatial autocorrelation in the labels inflates apparent performance \citep{ploton2020spatialvalidation,Kattenborn2022} if not handled at the splitting stage, geographic bias in the sampling distorts what the model learns, and the choice of tiling and gridding scheme can silently introduce data leakage or duplication. We address these issues in turn, starting with the data formats that underpin the entire dataset design.

\subsubsection{Data formats}
\label{subsec:data-formats}

The data format used to store an ML-ready dataset is a \textit{creation time} decision that propagates through the entire pipeline. It defines how array payloads are physically laid out, how metadata are encoded, how chunks or tiles are addressed, and which codec is required to decode the data. These choices are not strictly irreversible, but changing them requires repacking, rechunking, or rewriting the dataset, and they should therefore be treated as part of the dataset design rather than as a late implementation detail. The most important consequence of a format choice is the efficiency of the access pattern. The optimal layout minimizes \emph{over-read}, understood as the gap between bytes fetched from storage and bytes actually used by the model. In practice, this depends on how well the physical layout matches the logical access pattern of training, including the shape of the chunks, the band interleaving strategy, the compression block size, the placement of metadata, and the latency of the storage backend. Most formats used for ML on EO data can be understood through three broad storage patterns, summarized in Table~\ref{tab:data-formats}.

\paragraph{Container formats with an internal index}

The first pattern stores data inside \textit{binary containers}, each with its own internal metadata and offset structures. HDF5~\citep{folk1999hdf5}, netCDF-4~\citep{rew1990netcdf}, GeoTIFF~\citep{ogc_geotiff}, and COG~\citep{ogc_cog} belong to this family. HDF5 (Hierarchical Data Format 5) organizes data as hierarchical groups and arrays, with chunked storage the only layout supporting compression. NetCDF-4~\citep{rew2006netcdf4} builds on HDF5 while imposing the NetCDF (Network Common Data Form) data model of named variables, dimensions, and attributes, and is usually paired with the Climate and Forecast conventions~\citep{hassell2017cf}, which enables the comparison of datasets from different sources without ad hoc translation. GeoTIFF (georeferenced TIFF) extends TIFF (Tag Image File Format) with geospatial keys that tie raster coordinates to a coordinate reference system. Unlike HDF5 and netCDF-4, it is fundamentally a two-dimensional raster format, so spectral bands are carried as samples per pixel or as separate images, rather than as axes of an N-dimensional array. COG is a profile of GeoTIFF that constrains this layout for efficient partial reads over HTTP. Two properties govern performance across the family. The first is the order in which array values are serialized, called the \emph{interleave} (or chunk shape), which fixes which neighbors end up physically adjacent in the file and therefore which reads are cheap. The second is the cost of opening the file. The internal index of offsets and byte lengths must be fetched before any data can be addressed, but once that index sits in memory, reads to individual chunks are independent and parallel by design.

\paragraph{Key-value chunk stores}

The second pattern stores arrays as independently addressable chunks in a key-value store, of which Zarr~\citep{miles2020zarr} is the canonical example. Each encoded chunk sits under a key in an abstract store, which may be a directory on a filesystem or any backend exposing equivalent read and write operations. This removes the need to open a single monolithic container before locating chunk data, since the storage keys themselves provide the addressing mechanism, which aligns naturally with distributed readers and parallel workloads. The trade-off is that a naive Zarr layout can create very large numbers of small files or objects, which stresses POSIX metadata operations and filesystem inode limits, or inflates object storage request overhead. The Zarr v3 sharding extension (ZEP 2) addresses this by packing multiple logical chunks into a larger physical storage object while preserving chunk level access. Zarr defines array storage, chunking, codecs, and metadata containers, but not the geospatial meaning of coordinates, CRS, or affine transforms, which GeoZarr and related conventions are emerging to supply.

\paragraph{Columnar tables}

The third pattern stores samples as rows rather than arrays as chunks. Not every model consumes patches: for pixel-based algorithms, a table is the natural container. GeoParquet stores samples as rows, features and labels as columns, and the row group rather than the chunk as the unit of access; the geometry column keeps each sample georeferenced. Reading a subset of columns does not require touching the others, which suits feature tables that are wide but sparsely used.

\begin{table*}[!t]
\centering
\caption{Storage layout and practical properties of array and raster data formats relevant for AI4EO datasets.}
\label{tab:data-formats}
\footnotesize
\renewcommand{\arraystretch}{1.2}
\begin{tabularx}{\textwidth}{@{} L L L L L @{}}
\toprule
\textbf{Format} &
\textbf{Storage model} &
\textbf{Typical payload} &
\textbf{Random access mechanism} &
\textbf{Geospatial metadata} \\
\midrule
HDF5 &
Single binary container with groups, datasets, metadata, and internal chunk indices &
Heterogeneous N-dimensional scientific arrays &
Contiguous or compact offsets, or chunk index; performance depends on chunk shape &
Optional; requires domain-specific conventions \\
netCDF-4 &
netCDF data model on an HDF5 backend &
N-dimensional variables with named dimensions, attributes, and groups &
HDF5 chunking through the netCDF API &
Usually via CF conventions \\
GeoTIFF &
TIFF image container extended with GeoTIFF tags and keys &
Georeferenced raster imagery, usually 2-D with bands &
Strips or tiles; efficient spatial access requires tiling &
Built-in through GeoTIFF tags and keys \\
COG &
Tiled GeoTIFF with all metadata first and internal overviews &
Cloud-distributed geospatial rasters and image scenes &
Tile offsets and HTTP range requests; overviews avoid full-resolution reads &
Built-in through GeoTIFF tags and keys \\
Zarr &
Key--value hierarchy of metadata documents and encoded chunks &
Chunked N-dimensional arrays for cloud or distributed processing &
Chunk keys computed from grid indices; v3 sharding groups chunks into larger objects &
Usually CF conventions via xarray; GeoZarr emerging \\
(Geo)Parquet &
Columnar table of row groups with per-column statistics &
Tabular samples: one row per pixel or object, features and labels as columns &
Row-group offsets with column selection; no spatial chunking &
Geometry column and CRS via the GeoParquet specification \\
\bottomrule
\end{tabularx}
\end{table*}

\begin{runex}{Which axis has to be contiguous}
\textbf{WorldCereal} trains a per-pixel classifier on summary features computed from each location's time series. The series must be read time-contiguously to compute them, but what the training loop reads is one row per pixel.  \\[3pt]
\textbf{CHM} trains on spatial patches, so space is the axis that
must be contiguous and the data stay as chunked arrays. \\[3pt]
Neither layout is better than the other, and neither is a property of the data:
the access pattern is fixed by the model. This is why the format decision
cannot be deferred until the dataset has been built.
\end{runex}

\begin{decision}{Which access pattern to optimize for}
\textbf{Alternatives.} Store patch-contiguous, with spatial chunks covering one
or few dates; store time-contiguous, with all dates for a small spatial extent
in one chunk; or store tabular, one row per sample, for pixel-based
models. \\[2pt]
\textbf{Trade-off.} Each layout serves one read shape at a single read per
sample and the others at many. Mixed workloads either accept the penalty,
maintain two copies, or repack between stages. \\[2pt]
\textbf{Constrains.} The choice is fixed by the model's input shape, so it
cannot be settled before the architecture is; and repacking a built dataset is
a second full pass over the archive, at a cost comparable to generating it.
\end{decision}

\subsubsection{Dataset splits}

When evaluating the performance of an ML model, we want to avoid over-confident results. A conventional uniformly at random assignment of data samples to training and test sets can bias the evaluation if labels are spatially auto-correlated and clustered \citep{rolf2024mission}. A widely spread approach to tackle spatial autocorrelation is to perform block cross-validation, which \citet{Roberts2017} recommends using ``wherever dependence structures exist in a dataset, even if no correlation structure is visible in the fitted model residuals, or if the fitted models account for such correlations''. Spatial $K$-fold block cross-validation first divides the area into non-overlapping spatial blocks of a certain size, before assigning each block to one of $K$ folds. Model training and evaluation are then performed using the leave-one-fold-out method, where samples from a single fold are set aside for model evaluation, while samples from the remaining $K-1$ folds are used for model training, and the process is repeated for each fold, as in \citep{Mosig2026,Lusk_2026}. Alternatively, one fold can be assigned for model evaluation, one fold for validation, and the $K-2$ others for training, as in \citep{garnot2021panoptic}. Results are then reported across the concatenated test folds. In order to avoid cross-fold contamination, a good practice consists in enforcing spatial buffers between blocks, as in \citep{garnot2021panoptic}. Buffers and block size address the same underlying issue and are partially substitutable: sufficiently large blocks reduce edge leakage to negligible levels without explicit buffers \citep{Roberts2017}, while buffers are most useful when block size is constrained by study extent or sample distribution. We illustrate those different approaches in Figure \ref{fig:splits}. While spatial block cross-validation provides a reliable assessment of model generalizability, most works perform a more limited spatial block validation equivalent to setting $K=1$, in which blocks are partitioned once and randomly assigned to train, validation, and test sets \citep{Neumann_2025, herzog2025olmoearthstable, kerner2025fields, pauls2024estimating}. The definition of a block varies across works. Some impose a regular grid (\citet{Mosig2026} grid the Earth into non-overlapping blocks of 50 km in an equal area projection) while others adopt the geographical unit at which the data is delivered, such as Sentinel-2 tiles \citep{lang2023canopyheight} or Planet quads \citep{herzog2025olmoearthstable, glazer2025tempo}. \citet{Lusk_2026} take a more principled approach, determining the spatial autocorrelation via semivariograms at a $1\text{km}^2$ resolution and defining hexagonal blocks accordingly. The appropriate block size depends on the spatial scale of residual autocorrelation: blocks should be substantially larger than the distance at which model residuals become uncorrelated, since smaller blocks leave train and test samples near boundaries effectively dependent \citep{Roberts2017}. We elaborate on various gridding choices and their implications in the next section.

\begin{figure*}[!t]
    \centering
    \resizebox{\textwidth}{!}{\input{splits.tex}}
    \caption{Random splits leave train and test points autocorrelated, producing optimistic error estimates (a). Spatial blocks separate folds geographically, but when blocks are not much larger than the range of residual autocorrelation, points near block boundaries remain correlated across folds (b). Adding a buffer zone that excludes near-boundary samples removes this leakage and yields realistic evaluation (c). Equivalently, blocks sized substantially larger than the residual autocorrelation range can achieve realistic evaluation without explicit buffers. The buffered design is most useful when block size is constrained by data extent or sample distribution.}
    \label{fig:splits}
\end{figure*}

While we have focused on spatial blocking, the underlying principle generalizes to any source of dependence in the data: folds must be independent along whatever structure the model is expected to extrapolate across. Extensions include phylogenetic blocking for taxonomic relatedness \citep{Revell2010}, temporal blocking for serial autocorrelation \citep{Bergmeir2012}, and blocking in environmental or feature space, which tests generalization to novel conditions rather than novel locations \citep{meyer2021areaofapplicability}. The appropriate choice depends on what form of generalization the evaluation is meant to characterize.

Furthermore, in the case of geographically distributed labeled data, it might be of interest to evaluate the performance of the model for different regions, by defining individual splits for each region, as in \citep{kerner2025fields}. This also enables the assessment of a model's geographical generalization abilities in a comprehensive manner: any region can be held out from the training set and be evaluated on, as in \citep{butsko2025deploying}.

\begin{runex}{How the two products drew their splits}
\textbf{WorldCereal} divides its data randomly into 70\% train, 20\% validation and 10\% test samples, a split that takes no account of spatial structure at all.\\[3pt]
\textbf{CHM} holds out one test area of $960\,\text{m}\times960\, \text{m}$, placed at random inside each Sentinel-2 tile. Every tile therefore contributes both training and test data, and the separation between them rests on a minimum distance of 360\,m between training and testing patches. 
\end{runex}

\subsubsection{Spatial gridding}
\label{subsub:grid}

A map-making pipeline involves not one, but four different grids. The \emph{source grid} is how the EO data archive is organized; it is inherited, and its seams and overlap are to be dealt with (Section~\ref{subsubsec:optical}). The \emph{sampling grid} indexes the dataset: it determines which locations become samples, how they are deduplicated, and where the boundaries between training and test data fall. The \emph{inference grid} partitions the Earth for prediction, and the \emph{product grid} determines how the finished map is stored and distributed; both are discussed in Section~\ref{sec:making-a-map}. We hereby discuss the second. The three properties compared in Table~\ref{tab:grid_comparison} are the ones that matter for sampling: a grid is equal-area if every cell represents the same ground area; conformal if it preserves local angles and shapes, so that a square pixel corresponds to a square patch on the ground; and globally continuous if it covers the Earth without seams or inter-zone overlaps. We group candidate grids into four families, which we now describe.

\begin{table}[!t]
\centering
\caption{Comparison of candidate grids for global EO dataset construction.}
\footnotesize
\renewcommand{\arraystretch}{1.2}
\label{tab:grid_comparison}
\begin{tabular*}{\columnwidth}{@{\extracolsep{\fill}} l c c c @{}}
\toprule
\textbf{Grid} & \textbf{Equal-area} & \textbf{Conformal} & \textbf{\shortstack{Globally\\continuous}} \\
\midrule
\multicolumn{4}{@{}l}{\textit{Geographic}} \\
\addlinespace
Plate carrée & \texttimes & \texttimes & \checkmark \\
\addlinespace
\multicolumn{4}{@{}l}{\textit{Projection-based}} \\
\addlinespace
Web Mercator & \texttimes & \checkmark & \checkmark$^a$ \\
UTM / MGRS & \texttimes, \checkmark\textsuperscript{\faMapPin} & \texttimes, \checkmark\textsuperscript{\faMapPin} & \texttimes \\
Equi7 Grid & $\approx$ & \texttimes & \texttimes$^b$ \\
Equal Earth grid & \checkmark & \texttimes & \texttimes$^c$ \\
\addlinespace
\multicolumn{4}{@{}l}{\textit{DGGS}} \\
\addlinespace
S2 Geometry & $\approx$ & \texttimes & \checkmark \\
H3 & $\approx$ & \texttimes & \checkmark \\
HEALPix & \checkmark & \texttimes & \checkmark \\
\addlinespace
\multicolumn{4}{@{}l}{\textit{Indexing}} \\
\addlinespace
Major TOM & $\approx^d$ & n/a$^e$ & \checkmark \\
\bottomrule
\end{tabular*}

\vspace{1ex}
  \begin{minipage}{\columnwidth} 
    \footnotesize
    \textit{Globally continuous} indicates that the grid tiles the entire sphere with each point assigned to a uniquely defined cell and well-defined adjacency across all cell boundaries. $\approx$ \ indicates that the property holds approximately but not exactly. \textsuperscript{\faMapPin} The property holds locally within a single UTM zone, not globally.\quad $^a$\,Clips at $\pm85\degree$ latitude.\quad $^b$\,Continental boundaries have 50\,km overlap.\quad $^{c}$\,Grid cells do not align cleanly across the antimeridian.\quad $^d$\,Via implicit Voronoi tessellation of sample points.\quad $^e$\,Point-based grid; projection not prescribed.
  \end{minipage}

\end{table}

\paragraph{Geographic grids} partition the Earth by lines of constant latitude and longitude, indexing cells in angular rather than metric units. The simplest instance is plate carrée: cells of fixed angular extent that are trivial to index but whose ground area scales with the cosine of the latitude. A $1\degree \times 1\degree$ cell covers approximately 12{,}300 km$^2$ at the equator and about 2000 km$^2$ at 80$\degree$ latitude, such that uniform sampling on this grid systematically oversamples high latitudes.

\paragraph{Projection-based tilings} lay a regular metric Cartesian grid over a planar coordinate system produced by a general-purpose cartographic projection. The grid's geometric properties (conformality, area preservation, global continuity) are inherited from the projection. Web Mercator (EPSG:3857) is the industry standard of $256\times256$ pixel tiles used by web mapping platforms. It is conformal but not equal-area, with ground resolution varying by roughly a factor of six between the equator and $\pm80\degree$ latitude. UTM partitions the globe into 60 UTM zones of $6\degree$ width, each with a conformal transverse Mercator projection. MGRS \citep{NGA_SIG_0012_2014} subdivides each UTM zone by $\approx8\degree$ latitude bands into \textit{grid zones}, each further divided into 100 km × 100 km squares. The Sentinel-2 tiling scheme, which inherits the UTM/MGRS framework, delivers products as 109.8 km $\times$ 109.8 km tiles, such that locations within the resulting $\approx$10 km border strips appear in two (or more) tiles \citep{BauerMarschallinger2023}. The Equi7 Grid \citep{BauerMarschallinger2014} uses seven continental sub-grids in Equidistant Azimuthal projections with a multi-level tiling structure (100\,km $\times$ 100\,km at level T1), offering a principled alternative to MGRS. A single worldwide plane is simpler still: partitioning an equal-area projection such as Equal Earth (EPSG:8857) into regular cells gives exact equal-area stratification, at the cost of shape distortion toward the poles and a discontinuity at the antimeridian.

\paragraph{Discrete Global Grid Systems (DGGS)} are tessellations of the sphere designed to provide approximately or exactly equal-area cells with global continuity, avoiding the zone boundaries and overlaps of projection-based tilings. S2 Geometry \citep{veach2017s2} projects the sphere onto a cube and subdivides each face via a quadtree, yielding approximately equal-area quadrilateral cells (area ratio $\leq 2$ at any level). H3 \citep{h3} tiles the sphere in nested hexagons with similar distortion bounds. HEALPix \citep{Gorski2005} achieves exact equal-area pixelation of the sphere on iso-latitude rings; its rHEALPix variant \citep{Gibb2016} unfolds the tessellation onto a planar rectangle.

\paragraph{Indexing grids} The three families above all define cells, committing to a geometry before any imagery is read. A fourth approach defines only an index and leaves the geometry to the source. Major TOM \citep{Francis2024MajorTOM} defines a set of approximately equidistant sampling \emph{points} on the WGS84 ellipsoid, adapting longitudinal spacing to latitude so that high latitudes are not oversampled, while prescribing neither projection nor cell extent; patches are extracted in the source product’s native CRS.

\begin{decision}{Which grid to sample and split over}
\textbf{Alternatives.} A geographic grid, trivial to index but with cell area varying by latitude; a projection-based tiling such as UTM/MGRS, which matches how optical archives are delivered but carries zone seams and tile overlaps; a discrete global grid system, which buys equal-area cells and global continuity at the cost of reprojecting every patch; or an indexing grid that defines no cell at all. \\[2pt]
\textbf{Trade-off.} Equal-area cells make sampling density interpretable as labels per unit ground area and make area-based statistics meaningful; conformal, archive-native tilings avoid resampling the imagery. No grid offers both, and the choice is between distorting the sampling or distorting the pixels. \\[2pt]
\textbf{Constrains.} A grid whose cells overlap cannot guarantee disjoint splits however the folds are assigned, deduplication has to operate in a common geographic reference rather than on cell identifiers, and uniform sampling over its cells is biased toward the overlaps.
\end{decision}

\subsubsection{Data samples}
\label{subsubsec:gridtopatches}

With the data format, splitting strategy, and grid in place, the final step is to extract the actual samples used to train the model. We define a data sample as a $W\times H\times T \times C$ patch, where $W$ is the width, $H$ is the height, $T$ is the number of time steps, and $C$ is the number of channels (dimensionality of the EO data at hand). This process involves three considerations: the choice of projection, the sampling strategy, and the optional alignment of the various data modalities.

\paragraph{Sampling strategy} ML researchers typically need to extract EO data samples corresponding to a set of labeled points (or polygons) distributed non-uniformly. In label-scarce regimes, the classic approach is to extract a sample per label, a sampling design that inherits the geographic bias the labels might have. An alternative approach is to sample a certain number of samples per grid cell, keeping all labels in sparsely populated cells and subsampling in densely populated ones, preventing dense regions from dominating. The area that each cell represents impacts the sampling: if the cells have area variation (like S2/H3) or are latitude-distorted (plate carrée), this sampling implicitly gives smaller cells more samples per unit area. For the cap to correspond to a true ``labels per $\text{km}^2$'' ceiling, one can use equal-area grids (HEALPix) or weight by cell area.

\paragraph{Projection choice} The projection in which training samples are processed is a separate decision from the projection in which the finished map is delivered, and the two need not agree. This paragraph concerns the first, Section~\ref{sec:making-a-map} the second. Most works keep the training data in the native CRS at which it is distributed \citep{Mosig2026,feng2025tessera,brown2025alphaearthfoundations,Van_Tricht_2023,lang2023canopyheight,pauls2024estimating,pauls2026echosat}, which for Sentinel-2 is UTM/WGS84. UTM is conformal and holds ground sampling distance near-constant within a zone, so a pixel represents approximately $10\text{\,m}\times 10\text{\,m}$ on the ground everywhere within a zone \citep{Snyder1987} and a patch of fixed pixel dimensions covers a near-constant ground area. Others reproject to a single global CRS such as EPSG:3857 (Web Mercator) \citep{glazer2025tempo} or EPSG:4326 \citep{kerner2025fields}. The cost is that the same patch then represents different ground areas at different latitudes, so a model with a fixed receptive field sees a different spatial scale depending on where its training sample came from. If a single global CRS is required, an equal-area projection (e.g.,\ Equal Earth, EPSG:8857) holds that area constant, at the cost of shape distortion.

\paragraph{Multimodal alignment} When training patches combine data from multiple sensors (e.g., Sentinel-2 and ALOS-2 PALSAR-2), the inputs will generally differ in native projection, pixel spacing, and temporal sampling. The most common solution is to reproject and resample all modalities onto a shared grid and resolution before patch extraction, so that each training sample is aligned. This simplifies the data loader and model architecture but forces a choice of target resolution. The choice of resampling kernel also interacts with the downstream task: nearest-neighbor preserves categorical or integer-valued layers but introduces blocky patterns, whereas smooth kernels suit continuous fields but blur edges. An alternative is to preserve each modality at its native resolution and delegate alignment to the model architecture (Section \ref{subsec:model-design}), avoiding information loss at the cost of greater architectural complexity.

Taken together, these considerations argue for decoupling two concepts that are often conflated: the indexing strategy (how to select samples) and the patch projection (the coordinate system each sample lives in). Indexing should be done on a global grid that minimizes over-/under-sampling and overlap, independently of the patch projection and of how the EO data is tiled. Operations that rely on geography, such as train/test splitting, should be done on this grid. The patch projection, by contrast, only needs to minimize distortion at the scale of a single patch. UTM/WGS84, for instance, is a natural choice when the source archive is already UTM-native.

\subsection{Model design and training}
\label{subsec:model-design}

Model design must balance predictive accuracy against inference throughput at continental or global scale, and accommodate the heterogeneity of EO data. Effective model training must examine the specificities of large EO datasets to guide hyperparameter tuning and data augmentation.

\paragraph{Trade-off} Developing ML models for large-scale EO applications requires balancing model complexity, predictive performance, and inference throughput. This trade-off becomes critical in operational settings, where maps must be produced on fixed schedules (quarterly to yearly monitoring) or on demand (real-time alerts, post-disaster response) over wide areas. It is largely shaped by two choices: the model family, and the use of spatial context. Model families range widely in cost. At the simple end, tree ensembles such as gradient boosting and random forests remain common in operational products \citep{Van_Tricht_2023,esa2021worldcover,potapov2021mapping}, being cheap to train and deploy without GPUs. At the complex end, Vision Transformer architectures have attracted growing interest, including for large-area mapping \citep{Mosig2026,Neumann_2025,herzog2025olmoearthstable,pauls2026echosat}. Simpler convolutional encoder-decoder architectures remain widely adopted in operational pipelines, where the latency and memory constraints often favor their lower computational footprint \citep{Liu_2023,glazer2025tempo,kerner2025fields}. In particular, UNet architectures \citep{Ronneberger2015} with EfficientNet backbones \citep{efficientnet} offer a well-documented trade-off between accuracy and throughput for such settings. Most of these architectures operate on image patches, exploiting the spatial context around each pixel. Many operational products instead predict each pixel independently, from its own time series \citep{Van_Tricht_2023,feng2025tessera,Mosig2026}, giving up spatial information in exchange for smaller inputs and simpler inference. Knowledge distillation \citep{HintonKD} offers a further avenue to improve inference efficiency without notable performance degradation \citep{HIMEUR2025102742}.

\begin{decision}{Model family and spatial context}
\textbf{Alternatives.} Tree ensembles, convolutional encoder-decoders or vision transformers; each pixel predicted from its own time series, or within its spatial neighborhood. \\[2pt]
\textbf{Trade-off.} Larger models can improve predictions but raise the cost of inference, which recurs with every update of a global map; per-pixel prediction keeps inputs small and inference simple, at the price of the spatial information that patches provide. \\[2pt]
\textbf{Constrains.} The data layout (Section~\ref{subsec:data-formats}), the inference tiling and its artifacts (Section~\ref{subsec:generation}), and, in the other direction, the compute budget available for deployment.
\end{decision}

\paragraph{Handling input heterogeneity} EO applications frequently combine inputs that differ in spatial resolution, number of channels, or temporal cadence. Rather than resampling all modalities to a common grid before ingestion (as described in Section \ref{subsubsec:gridtopatches}), an alternative approach consists in delegating this alignment to the model itself. Typical designs employ modality-specific encoder branches that operate at each source's native resolution and fuse representations at a shared spatial scale, either through learned upsampling, cross-attention, or feature concatenation after spatial alignment layers \citep{Li2022multimodal}. This strategy avoids the interpolation artifacts and information loss inherent in a priori resampling. Recent geospatial foundation models have pushed this flexibility further, using dynamic patch embeddings and scale-adaptive encoders to train a single model across sensors with arbitrary channel counts, resolutions, and tile extents, while also gracefully handling missing or variable modalities at inference time \citep{Xiong2024dofa,Astruc2024anysat,Fuller2023croma}. The added architectural complexity must, however, be weighed against the throughput and simplicity advantages of a single pre-aligned input tensor, particularly in operational pipelines where inference latency is a binding constraint.

\paragraph{Hyperparameter tuning} Hyperparameter optimization (HPO) is an important step in the ML pipeline, and we refer the reader to \citep{HPO} for a general review of HPO good practices. We focus on two key aspects that are relevant for large EO datasets: \textit{which} parameters to optimize, and \textit{how much} data to use for it. Which parameters matter depends on the model family. Hyperparameter tuning for deep learning models is inherently challenging: the number of configurable hyperparameters is large, and their relative influence on model performance varies across tasks, architectures, and datasets, making it difficult to establish universally valid tuning priorities. Nevertheless, a broad consensus has emerged in the literature that the learning rate and its scheduler, batch size, and regularization mechanisms such as dropout and weight decay tend to be the most influential hyperparameters for deep neural networks \citep{Smith2018}. Classical ML models have fewer hyperparameters: for gradient boosting trees, for example, the learning rate is the main one to tune \citep{Probst2019tunability}. Regarding how much data should be used for HPO, the common practice is to use the entire validation set, as in \citep{Mosig2026,pauls2024estimating}. However, given the computational cost of training on large datasets, a strategy encountered in the literature (e.g., \citet{pauls2025capturing}) is to perform hyperparameter optimization on a subset of the data, and subsequently train the final model on the full dataset using the selected configuration. This approach is a special case of multi-fidelity hyperparameter optimization, and many methods have been developed for it \citep{AML,FABOLAS,AUTOMATA}, ranging from simple random subsampling to gradient-based subset selection. When complex subset selection methods are not warranted, stratified random sampling should be preferred over naive uniform sampling, particularly in the presence of class imbalance or long-tailed distributions, to ensure that the subset adequately reflects the class distribution of the original dataset. When using the full dataset is prohibitively expensive, we suggest adopting a two-stage strategy: first using the subset for efficient hyperparameter search to narrow the configuration space, then validation of the top-performing configurations on the full dataset before final model training. We encourage researchers to document their hyperparameter optimization or to report when none was performed, as in \citep{kerner2025fields}.

\paragraph{Data augmentation} In the SatML field, less attention is brought to data augmentation than in the general ML field. This is partly due to the fact that ”augmentation techniques designed for natural images should not be applied to satellite imagery without careful consideration” \citep{Hopkins2025}. Their findings suggest that while standard natural image techniques (particularly color-based augmentations) do not translate effectively to the satellite domain, geometric operations remain a beneficial approach. Furthermore, \citet{muhawenayo2026pruepracticalrecipefield} have identified that without specific data augmentation techniques to make them resilient, ML models often lack the necessary robustness to handle domain shifts (e.g., in terms of brightness) which can lead to decreased performance and visual artifacts. Various data augmentation techniques have been established in the literature to improve model generalization. Geometric transformations (such as rotations, flips, and random cropping) are being applied to EO data \citep{Brown_2022, glazer2025tempo, microsoft2024renewables, Liu_2023}), although particular care must be taken for SAR, since the look direction fixes which side of a slope shows layover or shadow. Recent research has introduced domain-specific augmentations tailored to EO data. To enhance spectral robustness, noise is frequently added to Sentinel-2 surface reflectance values \citep{Mosig2026, Brown_2022}, while \citet{muhawenayo2026pruepracticalrecipefield} advocate for varying Sentinel-2 processing strategies to mitigate sensitivity to shifting processing conventions. For models incorporating spatial metadata, \citet{Van_Tricht_2023} apply random perturbations to latitude and longitude coordinates. Furthermore, temporal augmentations, such as dropping timesteps, are utilized to simulate cloud-induced data gaps \citep{Mosig2026} or missing modalities. Finally, several approaches specifically target sensor-induced artifacts to ensure resilience during inference; these include simulating narrow no-data seams in Sentinel-1 GRD mosaics \citep{brown2025alphaearthfoundations}, as well as modeling swath boundaries, misregistration, and detector failures \citep{Pasquarella2023}.

\bigskip
\noindent A model trained with the practices described above will produce a prediction for every pixel on Earth. However, a prediction should come with a measure of its confidence. The model cannot perform equally well everywhere: it will be more reliable in regions well-represented in the training data than in under-sampled biomes, more confident under clear-sky conditions than under persistent cloud cover, and more accurate for common classes than for rare ones. Characterizing where and how much to trust the predictions is the subject of the next section, which surveys ML uncertainty quantification methods and argues that uncertainty is not a supplementary layer to be added as an afterthought, but a necessary component of any credible map product.

\section{Uncertainty quantification}
\label{sec:uq}

International frameworks for climate reporting, including the UNFCCC REDD+ mechanism \citep{UNFCCC2013_Warsaw_REDD} and IPCC national greenhouse gas inventory guidelines \citep{ipcc2006guidelines}, require that participating countries report estimates of forest carbon stock changes with quantified uncertainty. Carbon credit markets depend on this, as overconfident uncertainty bounds can overstate the precision of issued credits, while excessively wide bounds waste verification resources \citep{haya2020managing}. Beyond carbon accounting, uncertainty is what transforms a map from a static product into a tool that can guide action by identifying where additional field measurements would be most valuable, where policy conclusions are robust, and where they are not.

The challenge is that uncertainty quantification (UQ) for maps is different from UQ for individual predictions. Maps are spatial products, so users do not consume pixel values in isolation but aggregate them into regional statistics, compare them across years, and derive change estimates. A model may be well-calibrated on average across a global test set while being systematically overconfident in specific biomes or regions. If the uncertainty layer does not capture this spatial variation in reliability, it fails where it is most needed.

This section surveys the methods available for quantifying prediction uncertainty in large-scale EO maps, the techniques for assessing whether that uncertainty is trustworthy, the underappreciated role of spatial error correlation in aggregated estimates, and the practical question of what to do with uncertainty once it has been estimated.

Throughout, we distinguish two branches of large-scale mapping that differ in what an uncertainty estimate is. For continuous targets (biomass, canopy height, soil properties) uncertainty is a predictive distribution or interval around a real-valued prediction, and calibration is assessed by whether intervals achieve their nominal coverage. For categorical targets (land cover, crop type, change detection) uncertainty is a probability vector over classes, and calibration is assessed by whether those probabilities match observed class frequencies. The two branches share most of the method families below but not all of them, and they have different validation methods. Categorical mapping has a design-based tradition for aggregate accuracy and area estimation \citep{olofsson2014good, stehman2019key} but comparatively weak per-pixel uncertainty products, whereas continuous mapping has invested in per-pixel uncertainty but has only recently begun to address the aggregate problem (Section~\ref{sec:uq-spatial}). We note where method applications diverge as we go.

\subsection{Sources of uncertainty}
\label{sec:uq-sources}

The total uncertainty in a map prediction arises from multiple, largely independent sources that are usefully organized into the classical distinction between aleatoric uncertainty (irreducible noise inherent in the data) and epistemic uncertainty (reducible uncertainty arising from limited knowledge) \citep{derkiureghian2009aleatory}. In the context of EO-derived maps, however, this abstract distinction maps onto pipeline stages.

\paragraph{Upstream (input) uncertainty} Every observation that enters the model carries errors introduced before any machine learning is applied. Sensor noise, imperfect atmospheric correction, cloud mask failures, co-registration errors between sensors, compositing artifacts, and BRDF effects all contribute. These sources are predominantly aleatoric at the pixel level though the magnitude of the noise varies spatially (e.g., atmospheric correction is harder over bright surfaces) and can therefore appear epistemic at the map level. In principle, one would propagate input uncertainty through the model via Monte Carlo simulation over perturbed inputs \citep{heuvelink1998error}. In practice, this is rarely done since the computational cost is substantial, and the input uncertainty distributions are themselves poorly characterized. Nevertheless, it is important to acknowledge that model-derived uncertainty is only one component of the total error.

\paragraph{Label (reference data) uncertainty} The training targets are themselves uncertain. For example, GEDI footprint-level AGBD estimates carry uncertainty from allometric models and instrument noise \citep{Duncanson2022GEDI_AGB}, national forest inventory plots carry measurement error and representativeness uncertainty, and crowdsourced labels carry annotator disagreement. Label noise is a form of aleatoric uncertainty that sets a floor on achievable prediction accuracy. Methods that assume noise-free labels, which are most standard loss functions, will conflate label noise with model error in their uncertainty estimates.

\paragraph{Model (epistemic) uncertainty} This encompasses everything the model does not know such as gaps in the training data distribution, misspecification of the functional form, and sensitivity to hyperparameter choices. Model uncertainty is highest where the inference domain differs from the training domain, which is exactly the regime that matters for global maps, which inevitably extrapolate beyond the spatial and temporal extent of their training data.

Many UQ methods address only model uncertainty. The gap between what is quantified and what contributes to the total error budget should be made explicit.

\subsection{Uncertainty quantification methods}
\label{sec:uq-methods}

We organize UQ methods along two axes by whether uncertainty is intrinsic to the model (produced during training) or applied post hoc, and what type of uncertainty each method captures. Table~\ref{tab:uq-methods} provides a summary and we discuss each family below. The methods below represent different points on the trade-off between statistical rigor and computational feasibility, and no existing method occupies the ideal corner of that space.

\begin{table*}[!tb]
\centering
\caption{Overview of uncertainty quantification methods for large-scale EO maps.}
\label{tab:uq-methods}
\footnotesize
\renewcommand{\arraystretch}{1.2}
\begin{tabularx}{\textwidth}{@{} P{4cm} c c Y c @{}}
\toprule
\textbf{Method} & \textbf{Captures} & \textbf{Overhead} & \textbf{EO failure modes} & \textbf{Calibrated?} \\
\midrule
Gaussian Processes & Both & $O(n^3)$ & Kernel misspecification or non-stationarity & \checkmark$^*$ \\
Heteroskedastic NN & Aleatoric & $1\times$ & Blind to epistemic; weak variance signal & $\times$ \\
Distributional regression & Aleatoric & $1\times$ & Distribution family misspecification & $\times$ \\
Neural Processes & Both & $1\times$ & VI approximation error & $\sim$ \\
Deep Ensembles & Epistemic & $N\times$ & Correlated members from data & $\times$ \\
MC Dropout & Epistemic & $K\times$ & Systematic underestimation & $\times$ \\
Cheap ensembles$^\dagger$ & Epistemic & $1.2$--$2\times$ & Same as ensembles, reduced & $\times$ \\
Quantile Regression & Aleatoric & $1$--$Q\times$ & No calibration guarantee; crossing & $\times$ \\
Conformal Prediction & Marginal & $1\times$ & Only marginal coverage; exchangeability & \checkmark$^\ddagger$ \\
CQR & Both$^\S$ & $1\times$ & Marginal not conditional coverage & \checkmark$^\ddagger$ \\
\bottomrule
\end{tabularx}

\vspace{1ex}
  \begin{minipage}{\textwidth}
    \footnotesize
    Inference overhead is relative to a single forward pass. ``Calibrated'' indicates whether the method produces well-calibrated intervals by theoretical construction (\checkmark), requires post-hoc calibration ($\sim$), or provides no calibration guarantee ($\times$).
$^*$If kernel correctly specified.
$^\dagger$Snapshot ensembles, SWAG, BatchEnsemble.
$^\ddagger$Marginal coverage guaranteed; conditional coverage not guaranteed for spatial data.
$^\S$Inherits aleatoric adaptivity from base quantile model and the conformal wrapper provides marginal epistemic correction.

  \end{minipage}

\end{table*}

\subsubsection{Intrinsic methods}
\label{sec:uq-intrinsic}

\paragraph{Gaussian processes}

Gaussian processes (GPs) provide the theoretically principled baseline for spatial prediction with uncertainty \citep{rasmussen2005gaussianprocesses}. Under a well-specified kernel, the GP posterior yields exact, calibrated predictive distributions including at unobserved locations, which is the inference setting for map generation. The classical limitation is computational as exact GP inference scales as $O(n^3)$ in the number of observations (pixels usually), which is prohibitive for the millions-to-billions of data points typical of global EO datasets.

Sparse approximations based on inducing points \citep{titsias2009variational, hensman2013gaussian} reduce the cost to $O(nm^2)$ where $m \ll n$, making GPs feasible for moderately large datasets. However, the choice of kernel function is significant and the placement of inducing points is itself also an optimization problem that interacts with kernel specification. A fixed inducing set may adequately represent covariance structure in data-dense regions while underrepresenting data-sparse ones, precisely where uncertainty estimates matter most. For non-stationary EO data, this means the user must jointly navigate kernel selection, inducing point placement, and potential domain partitioning, which often require domain expertise.

EO data are non-stationary (tropical forests and savannas have fundamentally different covariance structures), and misspecifying the kernel undermines the calibration that is the GP's principal advantage. The geostatistical literature, where GPs appear under the name \textit{kriging}, has a long history of addressing non-stationarity and change-of-support problems \citep{cressie1993statistics, goovaerts2001geostatistical} and these insights remain relevant even as the ML community increasingly favors neural approaches.

\paragraph{Heteroskedastic and distributional regression}

The simplest neural approach to uncertainty is to add a second output head predicting the variance alongside the mean, trained via a Gaussian negative log-likelihood loss \citep{nix1994estimating}. This captures aleatoric uncertainty cheaply (because the overhead is a single additional output neuron) but is blind to epistemic uncertainty. If the model is systematically wrong in some region (e.g., due to training data gaps), the variance head will not reflect this. The training signal for the variance head is also weak since the model must learn the noise structure from the same data it uses to learn the function, and the loss landscape for the variance parameter is poorly behaved when the model is misspecified.

More expressive approaches predict the full parameters of a flexible distribution rather than just a Gaussian mean and variance. For skewed variables such as biomass, where a Gaussian assumption is often inappropriate even after log-transformation, mixture density networks \citep{bishop1994mixture} or distributions with explicit skewness and kurtosis parameters (e.g., sinh-arcsinh normal) \citep{jonespewsey2009} offer a richer representation of predictive uncertainty.

\paragraph{Variational inference and neural processes}

Neural processes (NPs) \citep{garnelo2018conditional, kim2018attentive} can be understood as amortized variational approximations to Gaussian processes. They learn to map a set of context observations to a predictive distribution, much as kriging maps observations to a conditional Gaussian, but with the kernel learned implicitly rather than specified parametrically. This connection to the geostatistical framework that underpins decades of spatial prediction is underappreciated, in part because the NP literature has developed largely within the meta-learning community using different vocabulary and training conventions. From the perspective of an EO researcher already familiar with kriging, the core proposition is not a new uncertainty framework but a scalable approximation to an existing one, with $O(n)$ inference cost and the capacity to learn non-stationary covariance structure directly from data.

The standard caveats of variational inference apply. The approximate posterior can underestimate uncertainty if the variational family is too restrictive, and the ELBO objective can trade off data fit against KL divergence in ways that are not transparent. In practice, design choices specific to the NP framework can produce well-calibrated prediction intervals. As an example, \citet{young2026interpolation, young2026neural} demonstrated that attentive neural processes produce residual standard deviations near the ideal value for biomass mapping across ecologically distinct regions, including under active deforestation. However, these remain among the few validated applications of NPs in large-scale EO mapping, and broader adoption has been limited by the absence of mature tooling and the unfamiliarity of the meta-learning paradigm.

\subsubsection{Post-training methods}
\label{sec:uq-posthoc}

\paragraph{Ensembles}

Deep ensembles \citep{lakshminarayanan2017deep} train $N$ models with different random seeds or bootstrap samples and take the spread as an uncertainty estimate. Their dominance in practice is largely explained by their simplicity. Ensembles require no changes to the base model architecture, loss function, or training pipeline. One simply trains the same model multiple times and measures disagreement. Random forests offer an even lower barrier, since every tree is already an ensemble member and quantile regression forests \citep{meinshausen2006quantile} formalize this by retaining the full distribution of training samples at leaf nodes. For those already using tree-based methods for mapping, uncertainty estimates are essentially a byproduct.

As an example, \citet{lang2022canopy} estimated uncertainty in global canopy height predictions using an ensemble of heteroskedastic CNNs, where each member predicts both a mean and a variance via Gaussian negative log-likelihood. Aleatoric uncertainty is estimated by averaging the individual variance predictions, and epistemic uncertainty by the variance of the individual means with a decomposition that follows the law of total variance. The approach illustrates both the appeal and the cost, as the heteroskedastic heads required no post-hoc calibration step, but inference at global scale meant running ten forward passes over the full domain.

However, ensemble spread measures model disagreement, which is related to but distinct from prediction interval width. For bootstrap based ensembles, including random forests, samples drawn from spatially autocorrelated data overlap substantially, and \citet{wager2014randomforestsci} showed that the resulting confidence intervals are only valid under conditions that such data systematically violate. Deep ensembles face a more severe form of the same problem class as members typically share the entire training set and differ only in random initialization, so the sole source of disagreement is optimization stochasticity. When the training data contain spatially structured biases as EO data invariably do, all members learn similar biases in the same regions and agree with each other where they are collectively wrong. In both cases, the mechanism parallels the well documented effect by which spatial autocorrelation inflates apparent model accuracy \citep{ploton2020spatialvalidation, Roberts2017}: the same data structure that makes test-set $R^2$ overly optimistic makes ensemble spread overly narrow.

Several approaches reduce the computational cost of ensembles without addressing the spread-versus-interval issue. Monte Carlo dropout \citep{gal2016dropout} approximates an ensemble by applying dropout at test time, but systematically underestimates uncertainty because it explores a narrow region of weight space. Snapshot ensembles \citep{huang2017snapshot} collect checkpoints along a cyclic learning rate schedule. Stochastic Weight Averaging (SWA/SWAG; \citealt{izmailov2018averaging}) averages weights or samples from a Gaussian approximation to the weight posterior. BatchEnsemble \citep{wen2020batchensemble} shares most parameters across members, reducing cost to approximately $1.2\times$ a single model. FiLM-Ensemble \citep{turkoglu2022film} modulates network activations via feature-wise linear modulation to produce diverse ensemble members from a single base network, achieving calibration comparable to explicit ensembles at a fraction of the memory cost. These techniques change the cost--calibration trade-off but do not resolve the correlation-induced miscalibration when applied to spatially structured data.

\paragraph{Quantile regression}

Rather than deriving prediction intervals from a point-estimate model, quantile regression directly targets conditional quantiles by optimizing a pinball (check) loss \citep{koenker1978regression}. One can train separate models for the $\tau$-th and $(1-\tau)$-th quantiles, or use a single model with multiple quantile heads, to obtain prediction intervals at any nominal level. Quantile Random Forests \citep{meinshausen2006quantile} retain the full distribution of training samples at leaf nodes to estimate conditional quantiles non-parametrically.

The appeal is directness as the model is optimized for the quantity of interest (interval bounds) rather than deriving intervals from variance estimates. The limitation is that there is no guarantee the resulting intervals are calibrated. A model trained for the 5th and 95th percentiles may achieve empirical coverage far from 90\%. Quantile crossing (the upper bound falling below the lower bound) can also occur, though isotonic regression or architectural constraints can mitigate this.

\paragraph{Conformal prediction}

Conformal prediction \citep{vovk2005algorithmiclearning} provides distribution-free coverage guarantees. Given exchangeable calibration data, a conformal procedure produces prediction sets that contain the true value with at least the nominal probability. The accessible tutorial by \citet{angelopoulos2023conformal} has driven recent adoption in the EO community \citep{singh2024uncertaintyconformal, valle2023luluconformal}.

The appeal is theoretical cleanliness as coverage is guaranteed without distributional assumptions, and practical cheapness since conformal methods operate as a post-hoc wrapper around any base predictor. The limitation for spatial mapping is that the guarantee is marginal. The coverage holds on average over the full test distribution, not conditionally for specific regions or subpopulations. For a map user who needs to know how uncertain a given pixel is in the Amazon, a guarantee that averages over the entire globe is of limited value.

More fundamentally, standard conformal prediction assumes exchangeability of calibration and test points, which spatial autocorrelation violates. Spatially correlated data are not exchangeable: nearby points carry redundant information, and the effective calibration sample size is smaller than the nominal one \citep{young2026samples}. This can lead to coverage that is correct marginally but arbitrarily poor conditionally \citep{barber2023conformal}. To address this limitation, \citet{guan2023localized} proposed localized conformal prediction, which emphasizes a local region around each test point when computing prediction sets, retaining the finite-sample marginal coverage guarantee while offering additional local coverage guarantees under suitable assumptions. Adapting such methods to the data volumes and non-stationarities of global EO maps remains an open problem, but they represent a promising direction beyond purely marginal guarantees.

Conformalized quantile regression (CQR; \citealt{romano2019conformalized}) offers a middle ground. CQR uses conformal calibration to adjust model-produced quantile intervals, inheriting the adaptivity of the base quantile model while gaining the marginal coverage guarantee. \citet{kuronen2025conformal} applied conformal prediction with a regression model to forest attribute mapping, using split conformal inference to construct prediction intervals and evaluating empirical coverage on withheld test data. While CQR does not solve the conditional coverage problem for spatial data, it is a practical post-hoc calibration approach that is currently available.

\paragraph{Classification}

For categorical maps the natural uncertainty output is the class probability vector, typically from a softmax layer, and several of the methods above transfer directly. Ensembles and MC dropout provide epistemic uncertainty as disagreement between members' probability vectors, and the aleatoric/epistemic decomposition follows the same law-of-total-variance logic as in the continuous case, expressed as total predictive entropy minus expected entropy (the mutual information between prediction and model parameters \citealt{houlsby2011bayesian, kendall2017uncertainties}). Conformal prediction yields prediction sets of classes rather than intervals, with the same marginal guarantee and the same exchangeability caveat \citep{angelopoulos2023conformal}. Quantile regression has no categorical counterpart.

However, the difference from regression lies in calibration. Raw softmax probabilities are usually miscalibrated, and modern networks are systematically overconfident \citep{guo2017calibration}. Temperature scaling, a single scalar fitted on held-out data, is the standard post hoc fix and is often sufficient. Platt scaling and isotonic regression handle non-uniform miscalibration \citep{niculescu2005predicting}, and class-wise or Dirichlet calibration is needed when miscalibration differs by class \citep{kull2019beyond}. Focal loss is a training-time alternative that implicitly regularizes toward higher-entropy outputs \citep{mukhoti2020calibrating}. A further distinction is that area estimation from a categorical map depends on the full probability vector, not only the top-label probability, so calibration of the entire vector matters even when only the argmax is used for producing outputs (Section~\ref{sec:uq-calibration}).

\subsubsection{Methods summary}

Unfortunately no method is simultaneously cheap, scalable, and well-calibrated. Gaussian Processes are calibrated (if the kernel is correct) but do not scale. Ensembles scale but are expensive and can be prone to overconfidence for spatial data. Conformal methods are cheap but provide only marginal coverage. Neural processes offer a favorable trade-off but require bespoke architectures. Heteroskedastic regression is cheap and scalable but captures only aleatoric uncertainty.

The appropriate choice depends on what will be done with the uncertainty. If the goal is to mask out unreliable regions, even a rough uncertainty proxy may suffice. If the goal is to produce calibrated prediction intervals for carbon accounting, the requirements are stringent and the cost is correspondingly higher. If the goal is to report aggregate uncertainty over administrative units, the dominant concern is spatial error correlation (Section~\ref{sec:uq-spatial}), and no pixel-level UQ method alone will suffice. Users should articulate the intended use of uncertainty before choosing a method, not after. Figure~\ref{fig:uq-flowchart} summarizes the decision logic discussed in the sections that follow.

\begin{figure*}[!t]
  \centering
  \resizebox{\textwidth}{!}{\input{uq_flowchart.tikz}}
  \caption{Decision flowchart for selecting an uncertainty quantification method based on the intended downstream use. Dashed arrow indicates that spatial aggregate uncertainty requires both a pixel-level UQ method and explicit modeling of spatial error correlation.}
  \label{fig:uq-flowchart}
\end{figure*}

\subsection{Assessing calibration}
\label{sec:uq-calibration}

The preceding methods differ not only in cost but in whether they produce calibrated uncertainty, which raises the question of how to verify calibration in the first place. Reporting uncertainty without verifying its calibration is akin to reporting accuracy without a test set. We recommend a minimal set of diagnostics that should accompany maps with per-pixel uncertainty:

\paragraph{Coverage and reliability} Do the $1\sigma$ ($\approx$68\%) prediction intervals contain 68\% of held-out observations, and the $2\sigma$ ($\approx$95\%) intervals contain 95\%? Coverage above the nominal level indicates underconfident (conservative) intervals and coverage below indicates overconfidence. Plot observed coverage against nominal coverage for a range of probability levels (e.g., 10\% through 90\%). A well-calibrated model lies on the diagonal. This generalizes the coverage check to all levels simultaneously and is the regression analog of the classification reliability diagram \citep{kuleshov2018accurate}.

\paragraph{Standardized residuals} For regression, compute $z = (y_\text{true} - y_\text{pred}) / \sigma_\text{pred}$. If the predicted uncertainties are well-calibrated, $z$ should follow $\mathcal{N}(0, 1)$. The Z-score mean (ideal: 0.0) detects systematic bias; the Z-score standard deviation (ideal: 1.0) detects miscalibrated interval width. Values above 1 indicate overconfidence; below 1 indicate underconfidence.

\paragraph{Scoring rules} The interval score (Winkler score; \citealt{gneiting2007probabilistic}) simultaneously penalizes miscoverage and interval width, rewarding intervals that are both sharp and calibrated. The continuous ranked probability score (CRPS) evaluates the full predictive distribution. These are more informative than coverage alone, which can be achieved by making intervals arbitrarily wide.

\paragraph{Additional diagnostics} Probability integral transform (PIT) histograms and Q-Q plots provide visual checks of distributional calibration. The uncertainty--error correlation assesses whether predicted uncertainty is informative about where the model makes large errors, within the domain where errors can be measured.

The diagnostics above require reference observations, and so can only be computed where reference observations exist. This is an inconvenient constraint, because epistemic uncertainty is most consequential where the inference domain departs from the training and reference domain, and for a global map that departure is the norm rather than the exception. It is therefore worth thinking about what can and cannot be established. Three regimes may be distinguished. In the first, reference data are available at sufficient density that calibration can be verified directly, stratified by biome or land cover, and reported as such. In the second, no reference data exist in the target region, but the model's behavior under extrapolation can be tested by proxy. We may hold out an entire region or biome from training (not merely pixels within it) and check whether predicted uncertainty rises there and whether it remains calibrated against that region's held-out observations.

This is the only direct test of epistemic uncertainty under the conditions that matter, and it is distinct from spatial cross-validation at the pixel or block scale, which tests interpolation rather than extrapolation. Weaker checks, such as confirming that predicted uncertainty increases with feature-space distance from the training data \citep{meyer2021areaofapplicability}, establish that the uncertainty is responsive but not that it is calibrated, and should be labeled accordingly. In the third regime, neither direct verification nor a credible proxy is available, and the map documentation may state that the uncertainty layer is unverified there rather than implying that the band carries the same meaning everywhere. A map's uncertainty product could therefore come with its own map that delineates where calibration was verified, where it was tested under proxy extrapolation, and where it is inferred only.

When calibration diagnostics reveal problems, the response depends on the severity and pattern of the failure. Mild, smooth miscalibration where the reliability diagram shows a consistent curve away from the diagonal can often be corrected post hoc. Isotonic regression applied to the predicted CDF values, or Platt scaling for parametric models, can remap predicted intervals to achieve empirical coverage without retraining \citep{kuleshov2018accurate}. These are cheap fixes and should be attempted first. If miscalibration is more structured, for instance, systematic overconfidence in one biome but reasonable calibration elsewhere, post hoc recalibration on a stratified held out set can help, but the stratification should reflect the structure of the miscalibration, which requires enough validation data in each stratum to fit the recalibration reliably. When miscalibration is severe or regionally patterned in ways that post hoc correction cannot resolve, the model itself may need revisiting: retraining with a proper scoring rule as the loss function (e.g., CRPS for distributional regression, or Gaussian NLL rather than MSE for heteroskedastic models) can produce better-calibrated uncertainty at source.

Switching to a method that provides calibration by construction such as conformal prediction as a post hoc wrapper is another option, with the caveat that the resulting guarantees are marginal rather than conditional. At a minimum, if calibration cannot be achieved, this should be documented explicitly by reporting the empirical coverage alongside the nominal coverage, stratified by region or land cover class, so that downstream users can judge for themselves whether the uncertainty layer is trustworthy for their application.

\paragraph{Classification diagnostics}

The categorical analogs of the above are the reliability diagram (observed frequency of correctness binned by predicted confidence), expected calibration error (ECE) as its scalar summary, and the Brier score as a proper scoring rule that, like the interval score, rewards both calibration and sharpness. There are, however, two caveats specific to maps, which is the context we care about. First, ECE and the standard reliability diagram assess only the top-label probability. For area estimation, where all class probabilities are summed over pixels, class-wise reliability diagrams or a class-wise ECE are the relevant diagnostics \citep{kull2019beyond}. Second, per-class diagnostics on rare classes (change, in particular) are dominated by the few reference observations available, and should be reported with their sample sizes.

\subsubsection{Qualifying a model before inference}
\label{sec:uq-qualification}

The recalibration and retraining remedies above presuppose that miscalibration can be, and is, discovered while retraining is still an option. For global maps, training is typically cheap relative to inference over the full domain, so the cost of discovering a regionally patterned failure late is not the retraining itself, but rather the inference run that must now be repeated. Due to the compute use differential, calibration assessment should therefore be treated as a sanity check that precedes wall-to-wall inference. To this end, we suggest a qualification step with a few example components. (i) calibration diagnostics of Section~\ref{sec:uq-calibration} computed on spatially and environmentally blocked holdouts, reported per stratum, since pooled diagnostics can hide regionally patterned miscalibration. (ii) feature space coverage analysis over the full inference domain, such as the area of applicability or a dissimilarity index \citep{meyer2021areaofapplicability}, which requires predictor data only and can be computed before prediction is made. This identifies where the model will be extrapolating and therefore where the proxy-extrapolation test above should be targeted. (iii) end-to-end inference on a small number of tiles chosen to span ecologically distinct regions, including regions that the coverage analysis flags as poorly supported, evaluated against whatever reference data exist there. This should also surface tiling, compositing, and edge artifacts (Section~\ref{sec:making-a-map}) that do not appear in pixel-level cross-validation.

Of course, this does not guarantee calibration where reference data are absent, since such regions are unverifiable by modeling alone. However, testing models before inference makes sure that failures that could, in principle, have been detected are detected before expensive inference compute.

\subsection{Spatial structure of uncertainty}
\label{sec:uq-spatial}

At what resolution does uncertainty operate? Per-pixel uncertainty is the default assumption, but whether it is meaningful depends on the model's effective receptive field and the spatial scale of the prediction errors. A model with a $256 \times 256$ pixel input patch does not make independent predictions at each pixel as neighboring pixels share the same input context and will have correlated uncertainties. Reporting per-pixel uncertainty in this setting is not wrong exactly but it may overstate the spatial resolution at which the uncertainty is informative. In reality, map errors are spatially correlated. If the model overpredicts biomass at one pixel, it likely overpredicts at neighboring pixels as well, because the same model biases, input errors, and training data gaps affect nearby locations.

This has a consequence for any user who aggregates pixel-level predictions into regional estimates such as total biomass in a country, total deforested area in a watershed, average canopy height in a management unit. If pixel errors were independent, the uncertainty of the aggregate would shrink as $1/\sqrt{n}$ with the number of pixels. But when errors are positively correlated, they do not cancel upon averaging and the uncertainty of the aggregate is much larger than the naive $1/\sqrt{n}$ calculation suggests.

\citet{wadoux2023spatialaverages} formalized this problem and demonstrated that a number of recent high-profile studies have reported uncertainty for spatial aggregates without accounting for spatial autocorrelation of map errors, leading to substantially underestimated confidence intervals. They propose block kriging as the geostatistically rigorous solution, which requires fitting a variogram to the spatial structure of the model residuals and using it to compute the variance of the spatial average.

The practical importance of this effect has been studied empirically. \citet{johnson2025flexible} produced model based uncertainty estimates for biomass aggregated to ownership parcels in New York State, incorporating all four recognized uncertainty components (reference data, sample variability, residual variability, and auxiliary data) as well as spatial autocorrelation of model residuals. They found that residual variance, largely resulting from spatial correlation of residuals, dominated all other sources of uncertainty for most parcels. This means that even if one perfectly characterized label noise, model variance, and input errors, ignoring spatial correlation would still produce dramatically underestimated aggregate uncertainty. \citet{kangas2023uncertainty} reached a similar conclusion in the forest mapping context. When errors are spatially correlated, they introduce domain-level bias in aggregated estimates that must be explicitly modeled, for instance via random area effects or distance-dependent correlation structures.

The same argument applies to categorical maps, where aggregates such as total cropland area or annual deforested area are sums of per-pixel class indicators, and the errors in those indicators are spatially clustered for the same reasons that regression residuals are. The consequence, however, manifests differently. In the continuous case, the dominant effect of correlated errors is variance that does not shrink with $n$ but in the categorical case it is bias. Whenever commission and omission of a class are unequal which is commonly the case, pixel counting yields an area estimate that is systematically too large or too small, and averaging over more pixels does not reduce this error but fixes it in place. When the confusion is itself regionally patterned (a crop confused with a neighboring class in one agro-ecological zone but not another), the bias is spatially structured as well. Hence the land cover community has deprecated pixel counting in favor of design-based area estimation, in which the map serves as a stratification and the area estimate and its variance are derived from a probability reference sample and the resulting confusion matrix \citep{olofsson2013making, stehman2013estimating, olofsson2014good}.

The design-based area estimation approach sidesteps the correlation of map errors altogether, because the variance comes from the sampling design rather than from the map, at the cost of requiring a probability sample of reference observations in the reporting unit. Spatial correlation re-enters only when the reference sample is clustered, where it reduces the effective sample size in the usual way \citep{stehman1997estimating}. Per-pixel outputs, whether variances or class probabilities, are not sufficient for reporting aggregate uncertainty, and something additional (a residual variogram, or a probability sample) is required. Change detection is the hardest case in the categorical branch, since the change class is rare, its errors are the most clustered, and the reference sample has to be sized for the rare class rather than for the map as a whole.

This remains one of the largest gaps between current practice in ML-based mapping and what is needed for rigorous area-based reporting. The methodology exists: block kriging on model residuals provides a rigorous estimate of aggregate uncertainty \citep{wadoux2023spatialaverages, johnson2025flexible}. However, fitting a variogram to residuals requires spatially distributed validation data dense enough to characterize the correlation structure, and that structure is unlikely to be stationary. Most mapping efforts lack the validation density to estimate this, let alone estimate it regionally. Until validation protocols are designed with spatial error correlation in mind by sampling densely enough in representative regions to fit at least regional variograms, aggregate uncertainty will remain underestimated by default. As a practical minimum, we recommend that mapping projects characterize residual spatial correlation in a small number of ecologically distinct validation regions, even if global estimation is infeasible. This provides at least an approximate bound on the aggregation effect and prevents the false precision that arises from treating pixel errors as independent.

\subsection{Operationalizing uncertainty}
\label{sec:uq-downstream}

Producing an uncertainty layer is only useful if it informs downstream decisions. We identify several modes of use, each with different requirements on the UQ method:

\paragraph{Masking} Threshold the uncertainty and mask out high-uncertainty pixels before release or before aggregation. This is conservative but wasteful because it discards information rather than using it. For this use case, any method that produces a reasonable ranking of pixel reliability suffices and calibration is desirable but not strictly necessary, since the uncertainty values are used only to threshold, not to interpret quantitatively.

\paragraph{Reporting as a separate band} Ship the uncertainty layer alongside the map and let users decide how to use it. For categorical products the choice of what to ship is itself a design decision. The maximum class probability, the predictive entropy, the full probability vector, and a conformal prediction set carry different information and different storage costs, and only the full vector supports downstream area estimation without loss. This is the most flexible option but places the burden of interpretation on the user, who may not have the statistical expertise to use it correctly. Calibration is essential here since if the reported uncertainty is systematically too narrow, users will make overconfident decisions.

\begin{runex}{Uncertainty in the two products}
\textbf{WorldCereal} ships a confidence layer rescaled from the winning class probability, and states that it ``simply reflects how certain the model is of its prediction'' and ``does not reflect actual accuracy'' \citep{Van_Tricht_2023}. \\[3pt]
\textbf{CHM} provides no per-pixel uncertainty layer \citep{pauls2026echosat}. \\[3pt]
Neither product therefore offers calibrated uncertainty in the sense of Section~\ref{sec:uq-calibration}.
\end{runex}

\paragraph{Weighting in post-processing} Use uncertainty to weight pixels during temporal smoothing, spatial aggregation, or multi-model combination. Pixels with high uncertainty receive less weight, stabilizing the aggregate. This requires well calibrated uncertainty at the pixel level.

\paragraph{Change estimation} Many downstream applications require not just single epoch estimates but differences between maps at two time points, such as annual deforestation area or biomass change. Propagating uncertainty through differencing introduces additional subtleties. If the same model and input pipeline are used for both epochs, errors may be positively correlated in time, causing them to partially cancel in the difference and yielding change estimates that are more precise than either single-epoch map alone but if the sensor, compositing strategy, or model version changes between epochs, the errors may be weakly correlated or independent, and the change uncertainty will be larger than users accustomed to the correlated case might expect.

\paragraph{Guiding data collection} Calibrated uncertainty identifies where additional observations would most reduce decision-relevant uncertainty, closing the loop between prediction and data collection. An overconfident model forecloses this feedback loop by masking the locations where additional data is most needed \citep{young2026interpolation}. Active learning strategies that target high-uncertainty regions can direct field campaigns, future LiDAR acquisitions, or validation sampling toward the locations with the highest expected information gain \citep{settles2009activelearning}. \citet{meyer2021areaofapplicability} proposed the related concept of an ``area of applicability'' as a spatial delineation of where the model's training data support reliable predictions, which can be viewed as a binary uncertainty mask derived from the distance between inference and training data in feature space.

\bigskip
\noindent Uncertainty quantification completes the modeling phase of the pipeline: the practitioner now has not only a trained model but also a principled framework for assessing how much to trust its outputs across space. What remains is to operationalize these components — to run inference at scale, mitigate the artifacts that inevitably arise when processing millions of patches, apply post-processing corrections that ensure temporal and spatial coherence, validate the result against independent reference data, and disseminate it in formats that are accessible and useful to downstream users. These operational stages are the subject of the next section, which covers map generation, post-processing, validation, and sharing. Although often treated as engineering details rather than research contributions, the choices made here can easily undermine the quality of an otherwise sound model, and they deserve the same level of methodological care.

\section{Map production and distribution}
\label{sec:making-a-map}

Producing a geospatial map product from a trained model is a non-trivial engineering task that extends well beyond inference. Generating predictions at scale requires choosing a compute architecture suited to the volume of EO data involved, and the tiling strategies that make large-area inference tractable can introduce spatial artifacts that must be addressed. Downstream applications typically require additional corrections, and some use cases call for a conversion from raster predictions to discrete vector geometries. Furthermore, making a map practically accessible to end users raises its own set of considerations: from how the data is compressed and spatially partitioned, to how it is projected and through which platforms it is served. This section describes the full pipeline from trained model to published map product, detailing the challenges to be expected at each stage.

\subsection{Map generation}
\label{subsec:generation}

Generating wall-to-wall predictions over large geographic extents requires processing thousands of satellite scenes through the trained model, a task that is both computationally intensive and logistically complex. Two aspects of this process warrant particular attention: the choice of compute infrastructure, which determines how the workload is distributed and scaled, and the treatment of spatial artifacts that emerge as a byproduct of tiled inference.

\subsubsection{Compute environments}

The computational strategy for large-scale EO/ML mapping is primarily dictated by the relationship between data volume and processing power. We distinguish between two primary regimes: local high-performance computing (HPC) and cloud-native environments. The decision of where to execute inference therefore involves a fundamental trade-off between bandwidth, cost, and architectural flexibility.

\paragraph{Data-to-Compute} This regime is most attractive to researchers with access to subsidized or institutional clusters, where the compute itself comes at little to no direct cost. Most major EO platforms (MPC, AWS Open Data, CDSE) do not charge egress fees, but the transfer is bounded by network bandwidth and provider-side rate limits, which can make the ingestion phase a significant bottleneck for global-scale mapping campaigns involving tens of thousands of satellite scenes. A compounding factor is the physical distance between the cluster and the data provider. Most EO platforms serve data from a single cloud region (e.g., \texttt{us-west-2} for NASA Earthdata, \texttt{eu-central-1} for the Sentinel archives on AWS, West Europe for Microsoft Planetary Computer, Warsaw for CDSE), so a researcher downloading from a different continent pays a steep latency penalty on every network request compared to an in-region client. Content Delivery Networks (CDNs) solve this by caching data on geographically distributed edge nodes, but none of the major EO data providers currently distribute data through a CDN. Hugging Face, which hosts a growing number of EO/ML datasets (which are relatively small fractions of the full archive data), is a notable exception owing to its use of CloudFront.

Once the data reaches the local infrastructure, further challenges arise. Storing the full dataset locally can be prohibitive at petabyte scale, and institutional clusters typically separate storage nodes from the high-throughput compute nodes where GPU or CPU inference is executed. This means that data sitting on a storage node must be read across an internal network at inference time, introducing an additional I/O bottleneck that can stall the pipeline. One way to sidestep the storage layer entirely is to stream the data directly to the compute nodes during inference, which avoids both the storage capacity constraint and the inter-node transfer overhead. Streaming comes with its own drawback, however: if multiple passes over the same scenes are required, each pass triggers a fresh download, effectively multiplying the external transfer burden and the geographic penalty described above.

\paragraph{Compute-to-Data} When data volumes, rate limits, or transfer times make the previous regime impractical, the alternative is to provision compute next to the data. The model is deployed within the same cloud ecosystem where the EO data already resides (e.g., AWS, Azure, CloudFerro, Google Compute Engine). This neatly eliminates the data transfer penalty that plagues institutional clusters, but only if compute and data are genuinely co-located in the same cloud region. Cloud providers offer compute resources across dozens of regions worldwide, and it is the researcher's responsibility to provision instances in the region where the EO data is actually stored. Launching compute in the wrong region silently reintroduces the same cross-continental latencies discussed above, compounded by inter-region egress fees. When colocation is correctly configured, however, read latencies drop to those of an internal network call, making this regime ideal for I/O-bound inference workloads. An inherent limitation is that the usable data pool is restricted to what the chosen platform hosts, both in terms of data sources and processing baselines: researchers cannot easily mix data from different providers and must accept the specific preprocessing version available on the platform. As a result, the sources and processing baselines of the platform intended for inference should already be considered when the training dataset is built (Section~\ref{sec:all-things-data}). Discovered only at inference, they can introduce a train–inference shift.

The dominant concern here is the price of compute itself. Costs can escalate quickly, particularly when heavy preprocessing must be performed at inference time rather than being precomputed. To limit these costs, some pipelines are built around CPU execution, using models that run efficiently without GPUs \citep{Van_Tricht_2023,feng2025tessera}. Spot or other preemptible instances offer substantial discounts but are subject to reclamation by the provider at any time, which means that inference jobs can be interrupted mid-execution. Running reliably on spot instances therefore requires a checkpoint-and-resume pipeline capable of recovering gracefully from preemption, along with orchestration tooling such as Kubernetes, Ray, or Flyte to manage job scheduling and restarts. Cost pressures also extend well beyond inference itself: downloading the resulting predictions incurs egress fees. Even their long-term storage on the cloud incurs recurring fees for both storage and egress (either on the user side or creator side), costs that persist well after the project ends and can become difficult to sustain. More economical alternatives exist and are discussed in Section~\ref{subsec:sharing}.

\subsubsection{Mitigation of artifacts}

Spatial artifacts are patterns in a map that result from the generation pipeline. Many originate in the input data (Section~\ref{sec:all-things-data}) and affect per-pixel and patch-based models alike: boundaries and overlaps between source tiles and orbits, mosaics of acquisitions from different dates, swath edges and viewing geometry, and regional differences in the number of valid observations. Their mitigation belongs mostly upstream, in preprocessing (e.g., BRDF normalization, see Section~\ref{subsubsec:optical-processing}) and in training augmentations that simulate such artifacts (Section~\ref{subsec:model-design}). Patch-based models add a source of artifacts of their own. Due to the significant memory constraints of modern ML architectures, EO scenes cannot be processed in their entirety. Instead, they are partitioned into a grid of smaller patches for inference. However, many models exhibit a spatial bias: predictive confidence tends to be highest near the center of a patch and degrades toward the borders \citep{huang2018tiling}. When these individually processed patches are reassembled into a final output, this bias manifests as patch artifacts: unnatural discontinuities or ``seams'' along the patch boundaries.

This spatial bias is a consequence of translational variance in neural networks, arising in convolutional architectures primarily from zero-padding and non-unary strides \citep{huang2018tiling,zhang2019making,kayhan2020translation}, and in vision transformers from patch embedding, positional encodings, and windowed attention \citep{rojasgomez2023making}. Architectural remedies exist \citep{zhang2019making,kayhan2020translation,olmoearthv12} but are rarely employed. Instead, researchers typically employ one of two strategies, illustrated in Figure~\ref{fig:inference-tiling}.

\begin{figure*}
    \centering
    \resizebox{\textwidth}{!}{\input{patches.tex}}
    \caption{Two strategies for mitigating patch artifacts.}
    \label{fig:inference-tiling}
\end{figure*}

\paragraph{Overlapping patches and blending} The first strategy defines an inference grid with overlapping patches, ensuring that boundary pixels receive multiple predictions. These redundant outputs are then merged using a weighted average based on each pixel's spatial distance from its respective patch center, as in \citep{Neumann_2025,herzog2025olmoearthstable}. Common weighting functions (e.g., 2D Gaussian kernels, cosine tapers, or linear ramps) prioritize the more reliable central predictions. In regression tasks, this aggregation is performed directly in the continuous output space; for classification, it is typically applied to the raw class probabilities. While effective at smoothing transitions between patches, this blending can subtly shift individual pixel predictions and demands careful treatment of predictive uncertainty.

A naive approach to propagating uncertainty through the averaging step is to compute the weighted average of the predicted variances alongside the weighted average of the means. This ignores the disagreement between overlapping predictions, which is itself a source of uncertainty. The correct combined variance follows the law of total variance:
\begin{equation}
    \sigma^2_{\text{combined}} = \sum_i w_i \sigma_i^2 + \sum_i w_i(\mu_i - \bar{\mu})^2,
\end{equation}
where the first term is the weighted mean of the individual variances and the second captures the spread between overlapping point estimates. Omitting the second term underestimates uncertainty at locations where overlapping patches disagree, which would typically be near land cover boundaries and abrupt transitions where reliable uncertainty matters most. This correction requires no additional model calls and is straightforward to implement. It does not fully account for the correlation between overlapping predictions, which share most of their input context, but is strictly preferable to naive variance averaging and captures the dominant source of error.

\paragraph{Padding and cropping} The second strategy uses padding to provide the model with broader spatial context for every pixel. Each patch is expanded with neighboring pixel data (or a specialized padding scheme at scene boundaries) during inference. After the forward pass, predictions are cropped back to the original patch dimensions, discarding the less reliable border outputs. This is the approach followed by \citep{brown2025alphaearthfoundations,Pasquarella2023,pauls2024estimating}. Because it retains only those predictions made with a full spatial context, this method is considered more faithful to the model's internal learned representations and avoids algebraically blending distinct predictions. If the margin covers the model's receptive field, every retained prediction is computed from the same context whichever patch it falls in, so adjacent patches agree at their boundary and no blending is needed. Convolutional networks have a finite receptive field; in a transformer with global attention, by contrast, every output depends on the whole input, so no margin fully removes the dependence on where patch boundaries fall, and tiled inference leaves seams even with large overlaps. Searchlight inference restores a finite receptive field by restricting attention to a fixed window centered on each pixel, so that the model applies the same operation everywhere \citep{olmoearthv13}.

The choice between the two strategies therefore follows from the model: where the margin can cover the receptive field, cropping suffices; where it cannot, blending smooths the residual seams rather than removing them. Beyond patch-level inference, these techniques are equally critical for mitigating tiling artifacts, larger-scale discontinuities that arise when inference is partitioned into standard geospatial tiles (e.g., UTM zones or Sentinel-2 tiles), each processed without context from its neighbors.

\begin{runex}{Pixels or patches at inference}
\textbf{WorldCereal} classifies each pixel independently, so the model's output cannot depend on where an inference tile ends, although a majority filter in post-processing does look at neighboring pixels. \\[3pt]
\textbf{CHM}: only the first of the three maps documents its inference \citep{pauls2024estimating}. It predicts on $512\times512$-pixel patches and keeps only the central $312\times312$, discarding a 100-pixel border on each side \citep{pauls2024estimating}. The kept centers tile the Earth without overlap, so nothing needs blending. The cost is that about 2.7 times as many pixels pass through the network as end up in the map.
\end{runex}

\subsection{Post-processing}
\label{subsec:post-processing}

Raw model outputs rarely constitute publication-ready data products. Depending on the application, additional post-processing may be required to enforce physical plausibility, ensure temporal and thematic consistency, or convert raster predictions into vector geometries suitable for downstream use. Notably, most correction strategies are contingent on the availability of redundant information (whether from multiple time steps or from jointly predicted variables) that can be leveraged to identify and suppress inconsistencies. Single-epoch, single-task predictions offer comparatively little basis for correction beyond external validation.

\subsubsection{Corrections}

When generating multi-temporal maps with consistent methodologies, the resulting temporal redundancy provides a critical opportunity to detect and suppress observational noise. Existing correction approaches fall into two broad families. The first relies on rule-based constraints rooted in domain knowledge, such as imposing biological growth limits on tree canopy height, smoothing sensor noise via moving-window trends \citep{turubanova2023tree,glazer2025tempo}, or stabilizing boundary pixels that would otherwise appear to jitter across time steps \citep{turubanova2023tree}. The second employs statistical or optimization-based trajectory modeling to identify the most plausible temporal path for individual pixels; examples include BFAST, Hidden Markov Models, and multi-model comparison methods \citep{copernicus2021globalland,anderson2026forest}, as well as Total Variation regularization to enforce piecewise-constant trajectories that preserve disturbance events while filtering high-frequency noise \citep{schwartz2025formspo}.

In multi-task settings, corrections must also enforce consistency across predicted variables. Physical dependencies can be maintained by nullifying secondary structural metrics (such as canopy height or biomass) whenever the primary variable falls below a minimum ecological threshold \citep{anderson2026forest}, while rule-based logic can resolve classification conflicts and align related outputs \citep{Van_Tricht_2023}.

\subsubsection{Polygonization}

The transition from pixel-wise semantic segmentation masks to vector-based geometries presents a significant bottleneck in modern geospatial ML pipelines. While deep learning models have reached high levels of semantic maturity, many tasks require going beyond pixel-level predictions to produce actionable, topologically sound vector data. This requirement is pervasive across diverse domains, including field or forest delineation \citep{kerner2025fields,schwartz2025formspo}, building mapping \citep{glazer2025tempo}, and the mapping of photovoltaic installations \citep{microsoft2024renewables}. Beyond geometric precision, this conversion is critical for data portability; vector geometries are significantly easier to distribute to stakeholders than dense rasters.

The primary difficulty stems from the inherent discretization artifacts of the raster grid, which often fail to align with the continuous or highly regularized boundaries of geographic entities. To address these inconsistencies, a multi-stage pipeline is typically employed. Initial noise attenuation can be achieved through mathematical morphology; for example, \citet{schwartz2025formspo} utilized an opening filter \citep{serra1982} to regularize mask contours and remove small-scale artifacts. Subsequent vectorization generally relies on boundary-following algorithms with 8-connectivity logic \citep{Suzuki1985}, as in \citep{glazer2025tempo}. To ensure computational efficiency and geometric fidelity, simplification algorithms \citep{POLY} can reduce vertex density. Despite these refinements, the gap between pixel-level probabilities and GIS-ready polygons remains a subject of active research, driving the development of geometry-aware loss functions and constrained regularization techniques \citep{9412866}.

\bigskip
\noindent Once post-processing yields a final raster or vector product, the remaining challenge is to make it accessible. The choices of compression, tiling, hosting, and projection collectively determine whether the map will be widely adopted or remain difficult to discover and reuse.

\subsection{Sharing}
\label{subsec:sharing}

To maximize the impact and reproducibility of the results, it is essential to provide a robust infrastructure for data sharing that addresses two distinct user requirements: \emph{large-scale batch downloads} for researchers who need to integrate spatial predictions into their own analytical pipelines, and \emph{interactive visualization interfaces} for stakeholders who want to explore results and perform targeted downloads of specific areas.

\subsubsection{Compression}

Data accessibility is inextricably linked to storage efficiency: large datasets can be a deterrent to researchers with limited storage resources. To facilitate the dissemination of high-resolution maps, compression provides a practical framework for reducing storage overhead. In scenarios where high numerical precision is not the primary requirement, lossy compression can be employed by rounding $32$-bit floating-point values to the nearest integer and storing them as $8$-bit unsigned integers. Where finer detail is necessary, fixed-point scaling offers a viable alternative: for instance, biomass values in the range of $[0, 500]$ can be multiplied by a factor of $100$ to preserve two decimal places within a $[0, 50000]$ range, enabling the use of $16$-bit unsigned integers. Furthermore, for applications demanding bit-for-bit accuracy, modern lossless compression methods (e.g.,\ \citep{pcodec}) ensure data integrity while minimizing the digital footprint. By adopting these multi-tiered compression strategies, data providers can significantly lower the barrier to entry for geospatial analysis.

\subsubsection{Product grid}

The product grid determines how the finished map is stored and distributed. At a minimum, the map should be distributed in the CRS in which it was predicted, typically the UTM zone of each tile, so that no pixel is resampled and its values are exactly those the model produced. It can additionally be offered in a single global CRS, which is convenient for visualization and for analyses spanning several UTM zones, but which requires reprojecting every pixel. Resampling introduces artifacts whose severity depends on the interpolation method and on the geometric distortion between the source and target projections.

Native-CRS products can be distributed as a collection of tiles, one per CRS involved. Tiles can be stored as COGs, which support partial reads via HTTP range requests, allowing downstream users to retrieve only the spatial extent or resolution level they need. Discovery and access can be further streamlined by publishing the tile collection as a SpatioTemporal Asset Catalog (STAC), which provides a standardized, queryable index of assets along with their spatial extents, timestamps, and associated metadata. The tiles can also be stored as Zarr arrays, which do not need STAC to be navigated.

A global layer in a single CRS is most easily distributed as one chunked array, such as a Zarr store. Some platforms, however, require other formats: Google Earth Engine, for example, ingests GeoTIFFs, and large areas are in practice uploaded as several tiles. Many global products are distributed as such tiles: the ESA WorldCover \citep{esa2021worldcover} and global canopy height \citep{lang2023canopyheight} products are distributed in 3\degree$\times$3\degree\ tiles, and the Tropical Moist Forests dataset \citep{Vancutsem_2021} in 10\degree$\times$10\degree\ tiles.

For products that include vector outputs (e.g.,\ from polygonization), cloud-native formats such as GeoParquet for efficient columnar storage, PMTiles for web-based visualization, and FlatGeobuf for streaming and spatial filtering over HTTP offer lightweight, interoperable alternatives to traditional shapefiles.

\begin{decision}{Which grid to distribute on}
\textbf{Alternatives.} The CRS in which the map was predicted, as tiles or one array per CRS; or a single global CRS, as tiles, one chunked mosaic, or a platform that reprojects on request. \\[2pt]
\textbf{Trade-off.} The native grid keeps every pixel as the model produced it; a global CRS is seamless and convenient but resamples every pixel. \\[2pt]
\textbf{Constrains.} Accuracy assessment and any analysis that depends on exact pixel values must use the native version; platforms that ingest only GeoTIFFs force tiling; area statistics on a grid that is not equal-area require per-pixel area weighting (Section~\ref{subsec:validation}).
\end{decision}

\subsubsection{Platforms}

Several platforms are available for hosting global products, each with different trade-offs in accessibility, infrastructure requirements, and long-term sustainability. Research data repositories such as Zenodo are built for archiving rather than access: their per-record limits (currently 50 GB and 100 files, extendable to 200 GB on request) mean that large products are typically split into regional files or archives, which users must download before use rather than stream for a specific area of interest. In exchange, they offer long-term preservation, versioning, and a DOI. Object storage provides a reliable backbone for distributing large datasets in cloud-native formats, and community-oriented repositories such as Source Cooperative lower the barrier further by offering (to this date) free hosting for open geospatial data. Self-hosting on an institutional server gives full control over formats and access at little direct cost, but ties the product's availability to the group's resources and staff. Google Earth Engine offers high visibility and built-in interactive tooling, but requires data to be ingested into its own infrastructure, introducing platform dependency. Public infrastructures such as the French Theia or the Belgian Terrascope keep data under public control with long-term commitments, often exposing it through STAC catalogs, although they are typically not yet as convenient to use as commercial platforms. Domain-specific platforms such as Global Forest Watch and EOForestSTAC provide established dissemination channels with dedicated user bases; however, contributing new datasets to these platforms typically requires coordination with their maintainers rather than direct self-service upload, making them more suitable as complementary high-visibility channels than as primary distribution mechanisms. Furthermore, no single platform is universally optimal, and the long-term availability of any given service cannot be guaranteed.

\bigskip
\noindent A map product is only as credible as its validation. The preceding sections addressed how to build a map; the next section addresses how to determine whether that map is accurate, and under what conditions its accuracy claims can be trusted. We argue that validation is not a final box-checking exercise but a design problem in its own right, one that should be planned alongside the map itself.

\section{Validation}
\label{subsec:validation}

Validation is between map production and map credibility. Yet, it is not a one-size-fits-all procedure and the validation protocol must be designed in relation to the intended use of the map and integrated into project planning from the outset. The choice of validation approach, metrics, and reporting standards should therefore be defined before any map is produced. This is increasingly relevant with the growing accessibility of large-scale computing infrastructure and ML methods, which have dramatically lowered the barrier to generating new maps.

This section focuses on the principles and good practices of \textbf{design-based map validation}, which forms the foundation of current international guidelines for geospatial accuracy assessment \citep{olofsson2014good, tyukavina2025practical}. Under this framework, validation relies on reference data collected through probability sampling designs with known, non-zero inclusion probabilities, enabling statistically rigorous and unbiased estimation of map accuracy. While many of these concepts originated in land-cover validation, they remain broadly applicable across a wide range of geospatial products.
The discussion applies to both categorical and continuous map products. Categorical products include multi-class land-cover maps as well as thematic single-class products such as forest extent or forest change maps. Continuous products include variables such as biomass and forest height. Although the general validation principles are shared, specific aspects of response design and accuracy assessment may differ across product types.

We begin by clarifying several common misconceptions surrounding map validation, including the distinction between model evaluation, map uncertainty, map comparison, and map validation. We then present the practical workflow of design-based accuracy assessment, covering sampling design, response design and statistical analysis.
Design-based accuracy assessment relies on reference data collected through probability sampling, which enables statistically rigorous and unbiased estimation of map accuracy. However, probability sampling is not feasible for all mapping applications. In cases where validation must rely on existing non-probability reference data (e.g., opportunistic field campaigns or legacy datasets), alternative validation approaches are required. These are discussed in Section~\ref{subsec:beyond_probability}, together with recommended good practices for situations in which probability-based, design-based accuracy assessment cannot be implemented.

\subsection{Common misconceptions in map validation}

Despite the central role of validation in establishing the scientific credibility of geospatial products, accuracy assessment methodology is frequently under-reported, and reference datasets are rarely made publicly available \citep{stehman2019key, foody2002status}. Four concepts are frequently conflated: \textbf{model evaluation}, \textbf{algorithm-derived uncertainty}, \textbf{map-to-map comparison}, and \textbf{map validation}. While the first three provide valuable information about model performance or product consistency, they do not constitute rigorous validation of the final map product. In this section, we focus specifically on map validation, namely the independent assessment of map quality using reference data, and clarify how it differs from these related but distinct concepts.

\paragraph{Model evaluation} Model evaluation is often confused with map validation. In many studies, standard ML practices (such as evaluating performance on a held-out test set) are presented as evidence of map quality.
Model evaluation is indeed an important component of map development, as the final map is generated through spatial application of the model across a large geographic domain. Consequently, model performance and map quality are inherently related. However, they are not equivalent. Model evaluation primarily measures how well the model generalizes to a held-out test dataset, whereas map validation concerns the quality of the final spatial product across the full mapped domain.
A held-out test set is typically not a probability sample of the map and may not represent the full range of landscapes, class distributions, and edge cases encountered in the final product. Even when the training sample is itself a probability sample and a portion is set aside for evaluation, this is generally not recommended: biases and errors present in the training data may transfer to the reference data \citep{tyukavina2025practical}. More broadly, when map and validation data originate from the same source or interpretation process, their errors tend to be correlated, and correlated errors inflate quality estimates, while independent errors deflate them \citep{foody2010assessing}. For these reasons, strong test-set performance of the model does not guarantee, and cannot substitute for, an independent assessment of map quality.

\paragraph{Algorithm-derived uncertainty} Metrics produced by ensemble variance, posterior probabilities, or model confidence scores quantify internal model uncertainty, not agreement with ground truth. These measures reflect what the model ``thinks'' about its own predictions and should not be interpreted as accuracy estimates \citep{lcgoodpractices}.

\paragraph{Map comparison} Comparison against an existing map product is another practice that is frequently mistaken for validation. In principle, validation should rely on reference data of higher quality than the product being evaluated, such as field observations or carefully interpreted high-resolution imagery. Existing map products, however, are often the result of inference procedures, including machine learning models or rule-based gridded approaches, and therefore contain their own uncertainties and biases. While map-to-map comparisons can provide valuable qualitative insights and help identify systematic differences between products \citep{strahler2006global}, they should not be interpreted as ground truth validation. What these comparisons measure is inter-product consistency, not map quality. They are more accurately described as consistency checks and should be reported as such.

\subsection{Design-based accuracy assessment}
\label{subsec:design_based}

A design-based accuracy assessment is more than the calculation of accuracy metrics from a set of reference samples. Rather, it is a statistical framework in which the validity of the resulting accuracy estimates depends jointly on how reference samples are selected, how their reference labels are determined, and how the results are analyzed. Accordingly, three components form the foundation of a statistically rigorous accuracy assessment \citep{Stehman1998}:
\begin{itemize}
    \item the sampling design used to select the reference sample,
    \item the response design used to obtain the reference for each sampling unit,
    \item the estimation and analysis procedures.
\end{itemize}
These components must be considered jointly \citep{stehman2001statistical, congalton2019assessing, tyukavina2025practical}. Across all stages, a clear and consistent problem definition is critical and should precede the validation process. In particular, the definitions used in sampling and labeling (such as what constitutes ``forest'' or the boundaries between land-cover classes) must be explicitly defined and aligned with the broader conceptual framing of the map product.

\subsubsection{Sampling design}

Sampling design determines how locations are selected for validation across the mapped domain. Design-based accuracy assessment derives its validity from the randomness of the sample selection process rather than from model assumptions \citep{stehman2001statistical}. Under this framework, every location within the study area must have a known, non-zero probability of being selected. This requirement is known as the inclusion probability criterion \citep{pengra2015global, tyukavina2025practical}.

A wide range of probability sampling strategies can be applied, including simple random, systematic, and stratified approaches. The choice of sampling design, as well as the definition of strata, directly influences the statistical properties of the resulting accuracy estimates \citep{olofsson2014good}. While simple random sampling provides an unbiased representation of the mapped domain, it is often inefficient because common classes dominate the sample, leaving insufficient observations for rare but important classes. Consequently, stratified random sampling is widely recommended as an adequate general-purpose design \citep{olofsson2014good, stehman2009sampling}. It allows better control over sample allocation across strata (e.g., land cover classes) while maintaining a probabilistic framework for unbiased inference.
Once a sampling strategy has been selected, an appropriate sample size must be determined. Sample size should balance statistical precision against the practical cost of reference labeling. Minimum sample sizes per class are generally recommended to ensure reliable class-specific accuracy estimates, and formal procedures for determining sample size under stratified designs are well established \citep{congalton2019assessing, olofsson2014good}.

A further consideration arises when validating maps containing rare classes, such as deforestation or urban expansion. Different sampling designs affect the balance between the precision of user's and producer's accuracy, particularly under severe class imbalance \citep{olofsson2020mitigating}. In such cases, buffer-based stratification methods can substantially improve estimation. \citet{olofsson2020mitigating} proposed introducing an additional stratum around mapped change areas, thereby increasing the likelihood of sampling near class boundaries where omission errors are most likely to occur and improving the estimation of rare-class areas.

Several operational implementations demonstrate how these sampling principles can be applied at large scales. A widely adopted approach is the stratified one-stage cluster approach \citep{olofsson2012making,pengra2015global,tsendbazar2021towards,xu2024comparative}, which combines stratification and clustering to balance statistical efficiency with the logistical constraints of human interpretation.

The study area is first divided into strata based on attributes such as biomes, after which clusters, referred to as primary sampling units (PSUs), are randomly selected within each stratum. In a one-stage design, every pixel within a selected cluster is interpreted, and each sample is assigned a weight based on its inclusion probability, enabling unbiased estimation of overall accuracy and class-specific areas \citep{tsendbazar2021towards}.
A representative example of this approach is the multi-purpose Global Land Cover Validation dataset developed for the Copernicus Global Land Service \citep{tsendbazar2018developing, tsendbazar2021towards}.
The dataset comprises more than 21,000 PSUs globally and has been used to validate products including CGLS-LC100m and ESA WorldCover, demonstrating that CEOS WGCV Validation Stage 4, the highest validation stage currently achieved for satellite-derived land cover products and characterized by sustained validation across successive product updates, is practically feasible at the global scale, at least for classes that can be labeled by visual interpretation of high-resolution imagery. Targets that cannot, such as crop types or canopy height, generally rely on the approaches of Section~\ref{subsec:beyond_probability}.

\subsubsection{Response design}

Response design specifies how reference labels are assigned at the sampled locations. It encompasses the choice of spatial assessment unit (pixel, spatial window, or object-based), the source of reference information (field observations, high-resolution imagery, or crowdsourced data), the labeling protocol, and the definition of agreement between map and reference labels. These decisions should be explicitly documented, as they directly affect the interpretation and reproducibility of the validation results \citep{tyukavina2025practical}.

Reference information may be obtained from field observations, visual interpretation of high-resolution imagery, crowdsourced platforms, or existing reference datasets. Field observations generally provide the highest-quality reference data but are often unavailable at the spatial and temporal scales required for large-area validation. Visual interpretation of high-resolution imagery by trained experts is the most commonly adopted approach. Although widely regarded as the best practical alternative, it is not fully reproducible: interpreter disagreement rates of 30\% or more have been reported \citep{powell2004sources}, with agreement varying substantially across land-cover classes \citep{pengra2020quality}. Crowdsourced platforms, such as Geo-Wiki, provide broader geographic coverage but generally at the expense of more variable labeling quality \citep{fritz2012geo, see2015crowdsourcing, fonte2015vgi}. More generally, the suitability of any reference data source depends on the mapping objective, spatial scale, and thematic classes being evaluated \citep{tsendbazar2015assessing}.

Several practices help improve the reliability of reference labels and reduce uncertainty. Reference data should be produced by interpreters who are independent from the map makers and unaware of map labels at sample locations, to avoid confirmation bias. Using multiple interpreters with consensus-based labeling further improves reliability \citep{tyukavina2025practical}. Spatial uncertainty arising from imperfect co-registration between the map and reference data can be mitigated by considering alternative labels within the spatial neighborhood of each sample unit, for instance by assigning the majority class within a $3\times3$ kernel around each reference pixel \citep{xu2024comparative,tsendbazar2016comparative}. Finally, reference observations should be temporally consistent with the map epoch to avoid discrepancies caused by land cover change between the acquisition dates of the map and the reference data \citep{tsendbazar2021towards}.

\subsubsection{Analysis}

Analysis refers to the statistical procedures used to derive accuracy metrics from the validation sample. For categorical maps, this typically involves constructing an error (confusion) matrix, which should be expressed in terms of area proportions rather than sample counts \citep{tyukavina2025practical}. When sampling is stratified or otherwise non-uniform, unequal inclusion probabilities between strata must be accounted for, since sample sites are typically not allocated proportionally to strata areas \citep{olofsson2012making, wickham2010effect}.
Sample weights, calculated as the inverse of inclusion probabilities, are therefore used to construct an area-weighted confusion matrix. This allows unbiased estimates of overall accuracy, class-specific accuracies, and area proportions to be derived. Inclusion probabilities and area weights refer to ground area, not to pixel counts. On a grid that is not equal-area, each pixel’s weight must be scaled by its ground area.

For categorical maps, user’s and producer’s accuracy (or equivalently, precision and recall), along with other metrics quantifying omission and commission errors, should be reported for individual classes in addition to overall accuracy. This is particularly important when a target class occupies a small proportion of the map, as high overall accuracy can mask poor performance on the class of interest \citep{tyukavina2025practical}. In addition to traditional accuracy metrics, alternative decompositions of classification error, such as quantity disagreement and allocation disagreement \citep{pontius2011death}, can provide complementary insight into the nature of map errors and are commonly adopted in the literature. For continuous-field products (e.g., tree cover fraction), adaptations of the standard confusion matrix framework, along with continuous accuracy metrics such as mean absolute error (MAE) and root mean square error (RMSE), are available \citep{pengra2015global, tyukavina2025practical}.

Accuracy metrics should be accompanied by measures of uncertainty, such as standard errors or confidence intervals, to reflect the variability inherent in the sampling process \citep{stehman2019key}. In practice, standard errors of accuracy metrics are often not reported, making it difficult for users to assess whether the provided estimates are reliable. Bootstrapping with replacement is commonly used to derive confidence intervals because it does not rely on parametric assumptions about the error distribution, although it does not correct systematic sampling bias in the validation design. 

\subsubsection{Reporting and transparency}

Transparent reporting of validation procedures is essential for ensuring credibility, reproducibility, and appropriate interpretation of map products \citep{stehman2019key}. At a minimum, studies should document the sampling design (including strategy, sample size, and allocation), response design (reference data sources, labeling protocol, and uncertainty considerations), and analysis methods (accuracy metrics, area estimation procedures, and uncertainty measures such as confidence intervals). Any deviations from probability-based sampling or limitations in reference data should be clearly acknowledged.

Where licensing allows, a held-out probability sample of the mapped domain, typically the validation sample, should also be released \citep{Lu2025regression}: its reference labels, the map's values at those locations, and what is needed to reconstruct inclusion probabilities. This lets users correct their own analyses for map error. Mapped values are increasingly used as outcomes or covariates in downstream regressions, where map errors can bias coefficient estimates and invalidate their confidence intervals.

For products that are operationally updated, accuracy should be monitored for temporal stability, not only assessed at a single point in time. \citet{tsendbazar2021towards} propose to quantify stability by calculating an index from class accuracies at different moments in time, following the requirement that omission and commission error should each remain below 15\% with comparable stability over time \citep{gcos2011systematic}.

\subsection{Validation beyond probability sampling}
\label{subsec:beyond_probability}

The preceding sections describe the recommended framework for map validation based on probability sampling and design-based accuracy assessment. In practice, however, this framework is not always feasible. Large-area mapping projects may lack purpose-designed probability samples, existing reference data may have been collected for different objectives, or certain products (particularly continuous-field variables) may require specialized validation strategies. In these situations, validation typically relies on one or more of three approaches: \textbf{alternative inferential frameworks}, \textbf{existing reference datasets}, and \textbf{product-specific validation protocols}. Although these approaches provide valuable evidence of map quality, they generally do not provide the same statistical guarantees as design-based accuracy assessment.

\begin{runex}{How the two products were validated}
\textbf{WorldCereal} validated its cropland extent on a purpose-built probability sample: 50{,}000 sites drawn at random in an equal-area projection before the maps were produced, each interpreted by several experts from very high-resolution imagery and Sentinel-2 and MODIS time series. Its user's and producer's accuracies are therefore estimates of map accuracy. Crop types could not be labeled this way; they were checked against an opportunistic sample of street-level photographs, and the results reported as agreement rather than accuracy. \\[3pt]
\textbf{CHM} was evaluated on held-out GEDI footprints, one random area per Sentinel-2 tile, with airborne LiDAR used as a local check over one French forest. This is model evaluation rather than map validation, and since GEDI does not sample north of $51.6^{\circ}$N, the map's forests there are not evaluated at all; the authors note that independent validation remains to be done. \\[3pt]
The difference follows from the target: cropland can be labeled from imagery, canopy height cannot.
\end{runex}

\paragraph{Alternative inferential framework} One alternative is to replace the design-based framework with model-based inference. Whereas design-based inference derives its validity from the randomness of the sampling process, model-based inference relies on a statistical model that describes how map errors relate to predictors such as landscape characteristics or data availability. This enables the use of non-probability reference data, such as opportunistic field plots or spatially sparse inventories, by extrapolating accuracy estimates to unsampled locations through the model \citep{stehman2019key}. The trade-off is that validity depends entirely on whether the assumed model is correct, which may be difficult to verify. Consequently, model-based inference has seen only limited adoption in operational map accuracy assessment \citep{tyukavina2025practical}.

\paragraph{Existing reference datasets} In practice, validation more commonly relies on existing reference datasets than on model-based inference. These include, for example, national forest inventories (NFIs), agricultural parcel registers, and airborne LiDAR campaigns. Whether such datasets can validate a map depends, among others, on its sampling design and the population it covers, its reference period and its independence from the training data. When the design is probabilistic and its population and definitions align with the product (for example an NFI used to assess a national forest map with a compatible forest definition), design-based estimation remains possible. However, in practice, NFIs are used to assess products with different thematic definitions, spatial extents, or target classes, such as global land cover products. In these situations, the original sampling design may no longer correspond to the validation objective, and the resulting assessment should be interpreted with appropriate caution. The results describe agreement at the reference locations rather than accuracy over the map, that is, verification rather than validation.

\paragraph{Product-specific validation protocols} Some geospatial products require dedicated validation protocols that extend beyond the general recommendations described above. Biomass mapping provides a representative example. The CEOS WGCV Land Product Validation subgroup has developed dedicated protocols that address the particular challenges of validating continuous-field products, including the use of plot-level field measurements, allometric uncertainty, and spatial scaling from plot to pixel \citep{ceos_biomass_protocol}. These protocols recommend combining field inventory data with airborne LiDAR transects to bridge the scale gap between ground plots and satellite-derived estimates. Even under such protocols, however, spatial coverage remains fundamentally limited, and the resulting accuracy characterization applies only to the sampled regions.

Regardless of the validation strategy adopted, the principles of transparency and rigorous reporting remain unchanged. Reference data sources and their limitations should be clearly documented, the distinction between verification and validation should be made explicit, and results should be interpreted as indicative rather than as unbiased estimates of map accuracy.

\bigskip
\noindent The preceding sections have traced the full arc of the EO mapping pipeline, from the processing of satellite observations through to a validated, disseminated product. At each stage, we have highlighted choices that are easy to get wrong and hard to diagnose after the fact — from the silent differences between data providers, through the spatial structure of training splits, to the gap between algorithm-derived uncertainty and genuine accuracy assessment. We conclude by summarizing the key recommendations that emerge from this analysis, examining the structural and institutional challenges that the community must confront as ML-based mapping matures from a research activity into an operational capability, and acknowledging the limitations of the present work.

\section{Conclusion}
\label{sec:conclusion}

\subsection{Summary}
\label{subsec:summary}

This paper has traced the full pipeline from satellite observations to validated, disseminated map products, exposing the interdependencies between stages that are easy to overlook when each is treated in isolation. We distill the central lessons into actionable recommendations, organized by pipeline stage.

\paragraph{Data landscape and access (Section~\ref{sec:eo-data-landscape})} The choice of data provider is not neutral. Different platforms apply different processing chains to nominally identical products, and these silent differences propagate into downstream models. Practitioners should document the exact provenance of their data (including the platform, processing baseline, and access date) and be aware that re-distribution platforms may introduce latency, processing divergence, or coverage gaps relative to canonical providers. When feasible, co-locating compute with data storage eliminates the transfer bottleneck that dominates global-scale workflows, but limits the pipeline to the sources and processing baseline that the platform hosts, which should therefore be considered before the training dataset is built.

\paragraph{Data selection and preprocessing (Section~\ref{sec:all-things-data})} Every preprocessing decision encodes assumptions that shape the training signal. Temporal strategy (single scenes, composites, or time series) should be guided by the task rather than by convention, and researchers should account for cross-hemisphere seasonal shifts when defining temporal windows for global products. Cloud masking thresholds involve a direct trade-off between input quality and data availability; overly aggressive filtering discards valuable pixels in persistently cloudy regions. For Sentinel-2, practitioners must handle partial orbits, split products, and tile-boundary overlaps explicitly to avoid pixel duplication and biased statistics. Sentinel-1 preprocessing is particularly sensitive to platform choice, as thermal noise removal, radiometric calibration conventions, and terrain correction differ substantially across providers; analysis-ready SAR mosaics, such as the Sentinel-1 monthly mosaics or the PALSAR yearly mosaics, offer an alternative to running the chain oneself. Radiometric normalization of optical data (scaling, clipping, BRDF correction) should be chosen deliberately rather than inherited uncritically from prior work.

\paragraph{ML dataset construction and model training (Section~\ref{sec:ml})} The data format chosen at dataset creation time should be treated as a design decision rather than a late implementation detail, chosen for the access pattern training requires (spatial patches, per-pixel time series, or tabular samples), since that pattern governs chunking efficiency and I/O overhead throughout the pipeline. Spatial autocorrelation demands spatially blocked dataset splits with buffers to prevent train--test leakage; random pixel-level splitting yields overconfident performance estimates \citep{Roberts2017,ploton2020spatialvalidation}. A pipeline involves four grids (source, sampling, inference, and product), chosen at different stages. The choice of sampling grid has underappreciated consequences: grids that are not equal-area introduce geographic bias, and grids that inherit sensor-tiling overlaps (such as MGRS) can silently duplicate training samples. We advocate a two-layer design that separates the sampling frame from the raster storage projection. On the modeling side, the trade-off between architectural complexity and inference throughput remains decisive for operational pipelines, and depends on the use of spatial context as much as on the model family. Data augmentation should be domain-specific, as standard natural-image augmentations do not necessarily transfer well to satellite data \citep{Hopkins2025}. Hyperparameter optimization should be documented systematically; when full-scale tuning is prohibitive, a two-stage strategy using a representative subset for search followed by validation on the full dataset offers a practical compromise.

\paragraph{Uncertainty quantification (Section~\ref{sec:uq})} Uncertainty is not an optional add-on but a requirement for any map that informs policy or scientific analysis. No currently existing method is simultaneously cheap, scalable, and well-calibrated; the appropriate choice depends on the downstream use of the uncertainty layer. A critical and widely underappreciated finding is that spatial autocorrelation of prediction errors causes aggregate uncertainty to be severely underestimated when pixel errors are treated as independent \citep{wadoux2023spatialaverages,johnson2025flexible}. At a minimum, practitioners should characterize residual spatial correlation in representative validation regions and report calibration diagnostics (coverage plots, standardized residuals, scoring rules) alongside the uncertainty layer.

\paragraph{Map generation, post-processing, and sharing (Section~\ref{sec:making-a-map})} Spatial artifacts arise from the input data as well as from tiled inference: the former are best handled upstream, while patch-boundary artifacts are removed by padding and cropping when the margin covers the model's receptive field, and otherwise require blending, with careful propagation of the uncertainty, or attention restricted to a local window. Post-processing corrections (temporal smoothing, cross-variable consistency checks, polygonization) should be documented as part of the product specification rather than treated as ad hoc fixes. For dissemination, cloud-native formats (COG, Zarr) and open hosting platforms reduce barriers to reuse. The authoritative product should stay in the CRS in which it was predicted, and be archived with a DOI; reprojected global versions serve as derived access layers.

\paragraph{Validation (Section~\ref{subsec:validation})} Model evaluation on a held-out test set is not the same as map validation. Rigorous accuracy assessment requires an independent probability sample with known inclusion probabilities, a transparent response design, and area-weighted estimation that accounts for the sampling strategy \citep{olofsson2014good,tyukavina2025practical}. When probability sampling is infeasible (for instance because suitable reference data do not exist, are prohibitively expensive to collect, or cannot cover the full mapped domain) alternative approaches such as existing reference datasets matched to the map's design and definitions, or opportunistic ground-truth datasets can still provide valuable verification, provided their limitations are stated explicitly: they measure consistency at sampled locations rather than accuracy over the full map extent. Map-to-map comparisons likewise measure inter-product consistency, not accuracy, and should be reported as such. Validation protocols should be defined before map production begins, and accuracy metrics should be accompanied by confidence intervals and stratified by region or class to expose structured errors. The reference sample itself, with the map's values at its locations and its design information, should be released where possible, so that users can correct downstream analyses for map error \citep{Lu2025regression}.

\bigskip\noindent A unifying theme across all stages is that choices made early in the pipeline (which provider to use, how to composite observations, how to split the data) propagate through every subsequent step and are difficult to diagnose or correct after the fact. Conversely, decisions that seem to belong at the end (the platform on which inference will run, the compute budget for deployment, the design of the validation sample) constrain choices at the beginning. Awareness of these dependencies, and explicit documentation of the decisions made at each stage, is what separates a technically sound and scientifically credible map product from one whose apparent quality cannot be verified.

\subsection{Looking forward}
\label{subsec:looking-forward}

The rapid maturation of ML-based EO mapping has exposed several structural challenges that individual research groups cannot resolve alone and that will shape the field's trajectory over the coming years. We highlight six directions that we consider both urgent and tractable.

\paragraph{Foundation models and transfer learning for EO} Geospatial foundation models pre-trained on large satellite archives \citep{Jakubik2023,satmae2022,brown2025alphaearthfoundations,feng2025tessera} promise to reduce the labeled-data requirements of downstream tasks and to learn representations that generalize across sensors, resolutions, and geographies. However, several open questions remain. The optimal pre-training data mix (in terms of sensor diversity, geographic coverage, temporal extent, and preprocessing level) is not yet established, and current models differ widely in these choices. Whether a single model can meaningfully span the spectral and geometric heterogeneity of the full EO data landscape, from submeter optical imagery to coarse-resolution SAR composites, or whether sensor-family-specific models will prove more effective, is an empirical question that the community is only beginning to address. Equally important is the development of systematic evaluation protocols: without standardized benchmarks that cover diverse geographies, tasks, and sensor configurations, it is difficult to compare foundation models on equal footing or to determine when fine-tuning outperforms training from scratch. A parallel development is the release of precomputed global embeddings \citep{brown2025alphaearthfoundations,feng2025tessera,olmoearthv13}, which pay the inference cost of a large model once and let users train lightweight models on top. These embeddings are map products in their own right, and every stage discussed in this paper applies to them, from their input data and tiling artifacts to their distribution grid.

\paragraph{Multi-modal and multi-scale learning} Many of the map products surveyed in this paper rely on a single sensor or a fixed combination of two. Yet the EO data landscape offers a far richer array of complementary modalities (optical, SAR, LiDAR, thermal, gravimetric) each with distinct spatial, temporal, and spectral characteristics. Architectures capable of ingesting arbitrary subsets of available modalities at their native resolutions, and of gracefully handling missing inputs at inference time \citep{Xiong2024dofa,Astruc2024anysat,Fuller2023croma}, represent a promising direction. The challenge extends beyond architecture design to data infrastructure: aligning heterogeneous sources in space and time at global scale, while preserving each modality's native information content, remains an engineering problem that lacks standardized solutions.

\paragraph{Standardization of preprocessing and provenance tracking} A recurring finding throughout this paper is that nominally identical data products differ across platforms in ways that are poorly documented and difficult to detect. The absence of standardized, reproducible preprocessing pipelines (particularly for Sentinel-1) means that two researchers starting from the same mission can arrive at materially different analysis-ready datasets without being aware of the divergence. Community-endorsed reference implementations for common preprocessing chains, accompanied by machine-readable provenance metadata that records every transformation applied to the data, would substantially improve the reproducibility and comparability of large-scale mapping efforts. Emerging specifications such as TACO \citep{taco} and initiatives around STAC extensions represent steps in this direction, as do analysis-ready products such as the Sentinel-1 monthly mosaics, whose processing chain is documented stage by stage, but broader adoption and institutional support are needed.

\paragraph{Validation as infrastructure} Current validation practice is overwhelmingly project-specific: each mapping effort designs its own accuracy assessment, collects its own reference data, and reports results using its own metrics and stratification. This fragmentation makes cross-product comparison difficult and wastes collective resources. Establishing and maintaining globally distributed, multi-purpose validation datasets (updated regularly, designed around probability sampling, and made openly available) would transform validation from a per-project cost into shared infrastructure \citep{tsendbazar2021towards}. Released together with the map values at their locations, such samples would also let users correct downstream analyses for map error \citep{Lu2025regression}. Such datasets should be designed with spatial error correlation in mind, sampling densely enough in representative regions to support variogram estimation and rigorous aggregate uncertainty reporting.

\paragraph{Uncertainty quantification as standard practice} Despite the methodological maturity of many UQ approaches (Section~\ref{sec:uq-methods}), uncertainty layers remain the exception rather than the norm in published map products. Making UQ a default component of every released map will require not only methodological advances (particularly in scalable methods that capture both aleatoric and epistemic uncertainty while remaining computationally tractable) but also community agreement on minimum reporting standards: which calibration diagnostics to include, how to communicate uncertainty to non-technical users, and how to handle the spatial aggregation problem that renders naive pixel-level uncertainty misleading at regional scales.

\paragraph{Open data infrastructure and long-term sustainability} The accessibility of EO data has improved enormously over the past decade, but the ecosystem remains fragile. Key re-distribution platforms have changed their pricing and access models with little advance notice, and the long-term availability of any commercial hosting service cannot be guaranteed. Building a durable infrastructure for large-scale EO mapping requires investment in platform-independent standards, community-governed data repositories, public infrastructures, and institutional commitments to long-term data stewardship. The convergence around COG and Zarr as open formats is encouraging, but formats alone do not ensure sustainability without the governance structures and funding models to maintain the repositories that host them.

\subsection{Limitations}
\label{subsec:limitations}

This article is subject to several limitations that should inform how its recommendations are interpreted.

First, the practical discussion of data selection and preprocessing focuses predominantly on Sentinel-1, Sentinel-2, and Landsat. While these are the most widely used open-access missions for large-scale ML-based mapping, the EO landscape encompasses many other sensors, including commercial very-high-resolution optical systems (Planet, Airbus and Vantor, formerly known as Maxar), hyperspectral missions (PRISMA, EnMAP, EMIT), and specialized instruments (GEDI, ICESat-2, GRACE), each with its own preprocessing challenges that we do not address in comparable depth. Researchers working with these data sources will encounter sensor-specific issues not covered here.

Second, we have framed the discussion primarily around the task of producing wall-to-wall raster maps at moderate resolution (10--30\,m). Many important EO applications operate at substantially different scales or produce fundamentally different output types: submeter mapping for urban or infrastructure monitoring, coarse-resolution products for climate variables, time-series-based change detection, and point-based estimation rather than spatially exhaustive prediction. The good practices articulated here may require adaptation for these settings.

Third, this is a rapidly evolving field. The foundation model landscape, the available data platforms and their access policies, and the state of the art in uncertainty quantification and validation methodology are all changing on timescales shorter than the publication cycle of a journal article. Some of the specific tools, platforms, and methods discussed here will inevitably be superseded; our intent has been to emphasize principles and design considerations that are more durable than any particular implementation.

Fourth, while we have aimed to cover the full pipeline from data acquisition to map dissemination, certain topics receive less attention than their importance warrants. In particular, we do not provide an in-depth treatment of active learning for label acquisition, the downstream use of map products in statistical analyses, the legal and ethical dimensions of large-scale mapping (including privacy, data sovereignty, and the socio-political implications of map products), or the organizational and institutional barriers to adopting the practices we recommend. Each of these topics merits dedicated treatment. Similarly, our treatment of validation draws mainly on land-cover practice, where reference labels can often be obtained by interpreting imagery; continuous variables and classes that require field data receive less detailed guidance.

Finally, the recommendations in this paper reflect the collective experience and perspective of the authors, shaped by our own research contexts. The EO mapping community spans a far broader range of applications, geographies, and institutional settings than any single group of authors can represent. We encourage readers to adapt these guidelines critically to their own contexts rather than to adopt them prescriptively.

\appendices

\input{divedeeper}

\section*{Acknowledgments}
The authors thank I. Tingzon, C. Ceccobello, and S. Hafner (RISE Research Institutes of Sweden), H. Herzog (Allen Institute for AI), C. Vancutsem (EU Joint Research Centre), R. N.\ Masolele and J. Reiche (Wageningen University), P. Potapov (World Resources Institute), L. de Vitry (Kanop), V. J.\ Pasquarella (Google), M. Santoro (Gamma Remote Sensing), D. Lusk (University of Freiburg), G. Belouze and M. Schwartz (LSCE), N. Lang (University of Copenhagen), J. Pauls (University of Munster), M. Kluczek (CloudFerro), C. Robinson (Microsoft AI for Good), H. Kristen (Joanneum Research), I. Corley (Taylor Geospatial), and Z. Feng (University of Cambridge) for generously sharing their experience and insights during the preparation of this work. Their input helped shape several of the practical recommendations presented here, though responsibility for any errors or omissions remains with the authors.

\clearpage
\bibliographystyle{IEEEtranN}
\bibliography{references}

\clearpage

\end{document}

%% file: boxes.tex
\definecolor{DecRed}{RGB}{198,32,40}
\definecolor{DefSlate}{RGB}{62,84,112}
\definecolor{RunTeal}{RGB}{0,105,110}
\definecolor{RevRed}{RGB}{198,32,40}   
\definecolor{NoteGrey}{RGB}{110,110,110}

\newtcolorbox{decision}[1]{%
  breakable, enhanced, sharp corners,
  colback=DecRed!4, colframe=DecRed, boxrule=0.4pt,
  left=5pt, right=5pt, top=4pt, bottom=4pt, toptitle=1pt, bottomtitle=1pt,
  fonttitle=\bfseries\scriptsize, coltitle=white,
  title={DECISION\ \ \textbullet\ \ #1},
  fontupper=\scriptsize}

\newtcolorbox{definition}[1]{%
  breakable, enhanced, sharp corners,
  colback=DefSlate!4, colframe=DefSlate, boxrule=0.4pt,
  borderline west={2pt}{0pt}{DefSlate},
  left=8pt, right=5pt, top=4pt, bottom=4pt, toptitle=1pt, bottomtitle=1pt,
  fonttitle=\bfseries\scriptsize, coltitle=white,
  title={DEFINITION\ \ \textbullet\ \ #1},
  fontupper=\scriptsize}

\newcommand{\term}[1]{\par\smallskip\noindent\textbf{#1}\quad\ignorespaces}

\newtcolorbox{runex}[1]{%
  breakable, enhanced, sharp corners,
  colback=RunTeal!4, colframe=RunTeal, boxrule=0.4pt,
  left=5pt, right=5pt, top=4pt, bottom=4pt, toptitle=1pt, bottomtitle=1pt,
  fonttitle=\bfseries\scriptsize, coltitle=white,
  title={RUNNING EXAMPLE\ \ \textbullet\ \ #1},
  fontupper=\scriptsize}

\newtcolorbox{sidebar}[1]{%
  breakable, enhanced, sharp corners,
  colback=black!2, colframe=black!55, boxrule=0.4pt,
  left=5pt, right=5pt, top=5pt, bottom=5pt, toptitle=1pt, bottomtitle=1pt,
  fonttitle=\bfseries\scriptsize, coltitle=white,
  title={DIVE DEEPER\ \ \textbullet\ \ #1},
  fontupper=\footnotesize}


%% file: overview.tex
\usetikzlibrary{arrows.meta}
\providecolor{RevRed}{RGB}{198,32,40}

\begin{tikzpicture}[
  >=Stealth, font=\sffamily,
  stage/.style={draw=black!45, fill=black!4, rounded corners=3pt, line width=0.8pt,
                minimum width=3.1cm, minimum height=1.5cm, align=center,
                inner sep=4pt, font=\sffamily\small},
  goal/.style={draw=RevRed!70!black, fill=RevRed!8, rounded corners=3pt, line width=1pt,
               minimum width=5.4cm, minimum height=1.0cm, align=center,
               font=\sffamily\small\linespread{1}\selectfont},
  fwd/.style={->, black!55, line width=1.1pt},
  spec/.style={->, RevRed!45, line width=0.7pt, dotted},
  back/.style={->, RevRed, dashed, line width=1pt},
  bl/.style={font=\sffamily\small\linespread{1}\selectfont, text=RevRed, align=center,
             fill=white, inner sep=2.5pt}
]
\def\dx{3.6}

\node[goal] (goal) at (2.5*\dx, 2.6) {\textbf{Large-scale EO/ML map}};

\node[stage] (s2) at (0,0)      {\textbf{\S\,II}\\EO data\\landscape};
\node[stage] (s3) at (\dx,0)    {\textbf{\S\,III}\\Selection \&\\preprocessing};
\node[stage] (s4) at (2*\dx,0)  {\textbf{\S\,IV}\\ML dataset\\\& training};
\node[stage] (s5) at (3*\dx,0)  {\textbf{\S\,V}\\Uncertainty\\quantification};
\node[stage] (s6) at (4*\dx,0)  {\textbf{\S\,VI}\\Production \&\\distribution};
\node[stage] (s7) at (5*\dx,0)  {\textbf{\S\,VII}\\Map\\validation};

\foreach \a/\b in {s2/s3, s3/s4, s4/s5, s5/s6, s6/s7}
  \draw[fwd] (\a.east) -- (\b.west);

\foreach \s in {s2,s3,s4,s5,s6,s7}
  \draw[spec] (goal.south) to[out=-90,in=90] (\s.north);

\def\ba{-1.85} \def\bb{-3.3} \def\bc{-4.6} \def\bd{-5.8} \def\be{-7.0}


\draw[back, rounded corners=6pt]
  ([xshift=1.1cm]s5.south) -- (11.9,\ba) -- (7.2,\ba) -- (s4.south);
\node[bl] at (9.55,\ba) {uncertainty needs\\constrain model choice};

\draw[back, rounded corners=6pt]
  ([xshift=-0.3cm]s6.south) -- (14.1,\bb) -- (6.7,\bb) -- ([xshift=-0.5cm]s4.south);
\node[bl] at (10.4,\bb) {deployment budget shapes\\model development};

\draw[back, rounded corners=6pt]
  ([xshift=0.3cm]s6.south) -- (14.7,\bc) -- (3.9,\bc) -- ([xshift=0.3cm]s3.south);
\node[bl] at (9.5,\bc) {training and inference data must match};

\draw[back, rounded corners=6pt]
  ([xshift=-0.3cm]s7.south) -- (17.7,\bd) -- (6.25,\bd) -- ([xshift=-0.95cm]s4.south);
\node[bl] at (11.98,\bd) {hold out the validation regions from training};

\draw[back, rounded corners=6pt]
  ([xshift=0.3cm]s7.south) -- (18.3,\be) -- (3.3,\be) -- ([xshift=-0.3cm]s3.south);
\node[bl] at (10.8,\be) {acquire data where you intend to validate};

\end{tikzpicture}

%% file: splits.tex
\begin{tikzpicture}[
    >=Stealth,
    every node/.style={font=\sffamily},
    heading/.style={font=\sffamily\normalsize\bfseries},
    sub/.style={font=\sffamily\scriptsize\itshape, text opacity=0.45, align=center},
    annot/.style={font=\sffamily\scriptsize, text opacity=0.7},
    legendlabel/.style={font=\sffamily\scriptsize, text opacity=0.65, anchor=west},
]

\definecolor{trainblue}{RGB}{74,144,217}
\definecolor{testred}{RGB}{226,85,85}
\definecolor{foldA}{RGB}{74,144,217}
\definecolor{foldB}{RGB}{232,168,56}
\definecolor{foldC}{RGB}{91,174,91}
\definecolor{foldD}{RGB}{199,90,155}
\definecolor{goodgreen}{RGB}{46,139,87}
\definecolor{midorange}{RGB}{255, 143, 0}

\def\R{0.09}

\newcommand{\landpath}{
    (0.1, 2.3) .. controls (0.3, 2.6) and (0.7, 2.5) ..
    (1.1, 2.55) .. controls (1.5, 2.7) and (2.0, 2.5) ..
    (2.4, 2.6) .. controls (2.8, 2.75) and (3.2, 2.55) ..
    (3.6, 2.65) .. controls (3.9, 2.7) and (4.1, 2.55) ..
    (4.2, 2.4)
    -- (4.2, 0.5)
    .. controls (4.0, 0.3) and (3.6, 0.35) ..
    (3.2, 0.3) .. controls (2.8, 0.2) and (2.4, 0.35) ..
    (2.0, 0.25) .. controls (1.6, 0.15) and (1.2, 0.3) ..
    (0.7, 0.25) .. controls (0.4, 0.2) and (0.2, 0.35) ..
    (0.1, 0.5) -- cycle
}

%

\def\panelshift{5.2}

\begin{scope}[shift={(0,0)}]
    \node[heading] at (2.1, 3.2) {(a) Random split};

    \fill[black, opacity=0.05] \landpath;
    \draw[black!20, line width=0.6pt] \landpath;

    \fill[trainblue, opacity=0.75] (0.5, 2.2) circle (\R);
    \fill[testred, opacity=0.75]   (1.0, 2.0) circle (\R);
    \fill[trainblue, opacity=0.75] (0.7, 1.85) circle (\R);
    \fill[testred, opacity=0.75]   (1.4, 2.3) circle (\R);
    \fill[trainblue, opacity=0.75] (1.6, 2.1) circle (\R);
    \fill[testred, opacity=0.75]   (0.4, 1.0) circle (\R);
    \fill[trainblue, opacity=0.75] (0.9, 0.7) circle (\R);
    \fill[trainblue, opacity=0.75] (1.3, 0.5) circle (\R);
    \fill[testred, opacity=0.75]   (0.65, 0.85) circle (\R);
    \fill[trainblue, opacity=0.75] (1.1, 0.4) circle (\R);
    \fill[testred, opacity=0.75]   (2.7, 2.2) circle (\R);
    \fill[trainblue, opacity=0.75] (3.2, 2.0) circle (\R);
    \fill[testred, opacity=0.75]   (2.9, 1.85) circle (\R);
    \fill[trainblue, opacity=0.75] (3.6, 2.35) circle (\R);
    \fill[trainblue, opacity=0.75] (2.6, 0.7) circle (\R);
    \fill[testred, opacity=0.75]   (3.0, 0.9) circle (\R);
    \fill[trainblue, opacity=0.75] (3.5, 0.65) circle (\R);
    \fill[testred, opacity=0.75]   (2.8, 0.4) circle (\R);
    \fill[trainblue, opacity=0.75] (3.3, 0.5) circle (\R);
    \fill[trainblue, opacity=0.75] (1.9, 1.6) circle (\R);
    \fill[testred, opacity=0.75]   (2.25, 2.05) circle (\R);
    \fill[testred, opacity=0.75]   (1.7, 1.3) circle (\R);
    \fill[trainblue, opacity=0.75] (2.3, 1.35) circle (\R);
    \fill[testred, opacity=0.75]   (3.1, 1.45) circle (\R);

    \node[sub] at (2.1, -0.15) {Train/test neighbors share\\autocorrelated signal};
    \node[annot, testred, font=\sffamily\scriptsize\bfseries] at (2.1, -0.65) {$\rightarrow$ Optimistic evaluation};
\end{scope}

\begin{scope}[shift={(\panelshift, 0)}]
    \node[heading] at (2.1, 3.2) {(b) Spatial blocks};

    \fill[black, opacity=0.05] \landpath;
    \draw[black!20, line width=0.6pt] \landpath;

    \begin{scope}
        \clip \landpath;
        \fill[foldA, opacity=0.12] (0, 1.5) rectangle (2.1, 2.8);
        \fill[foldB, opacity=0.12] (2.1, 1.5) rectangle (4.2, 2.8);
        \fill[foldC, opacity=0.12] (0, 0.1) rectangle (2.1, 1.5);
        \fill[foldD, opacity=0.12] (2.1, 0.1) rectangle (4.2, 1.5);
    \end{scope}

    \draw[black!30, line width=0.8pt, dashed] (2.1, 0.2) -- (2.1, 2.7);
    \draw[black!30, line width=0.8pt, dashed] (0.15, 1.5) -- (4.15, 1.5);

    \fill[foldA, opacity=0.75] (0.5, 2.2) circle (\R);
    \fill[foldA, opacity=0.75] (1.0, 2.0) circle (\R);
    \fill[foldA, opacity=0.75] (0.7, 1.85) circle (\R);
    \fill[foldA, opacity=0.75] (1.4, 2.3) circle (\R);
    \fill[foldA, opacity=0.75] (1.6, 2.1) circle (\R);
    \fill[foldC, opacity=0.75] (0.4, 1.0) circle (\R);
    \fill[foldC, opacity=0.75] (0.9, 0.7) circle (\R);
    \fill[foldC, opacity=0.75] (1.3, 0.5) circle (\R);
    \fill[foldC, opacity=0.75] (0.65, 0.85) circle (\R);
    \fill[foldC, opacity=0.75] (1.1, 0.4) circle (\R);
    \fill[foldB, opacity=0.75] (2.7, 2.2) circle (\R);
    \fill[foldB, opacity=0.75] (3.2, 2.0) circle (\R);
    \fill[foldB, opacity=0.75] (2.9, 1.85) circle (\R);
    \fill[foldB, opacity=0.75] (3.6, 2.35) circle (\R);
    \fill[foldD, opacity=0.75] (2.6, 0.7) circle (\R);
    \fill[foldD, opacity=0.75] (3.0, 0.9) circle (\R);
    \fill[foldD, opacity=0.75] (3.5, 0.65) circle (\R);
    \fill[foldD, opacity=0.75] (2.8, 0.4) circle (\R);
    \fill[foldD, opacity=0.75] (3.3, 0.5) circle (\R);
    \fill[foldA, opacity=0.75] (1.9, 1.6) circle (\R);     
    \fill[foldB, opacity=0.75] (2.25, 2.05) circle (\R);   
    \fill[foldC, opacity=0.75] (1.7, 1.3) circle (\R);     
    \fill[foldD, opacity=0.75] (2.3, 1.35) circle (\R);    
    \fill[foldD, opacity=0.75] (3.1, 1.45) circle (\R);    

    \node[sub] at (2.1, -0.3) {Folds separated, but cross-fold\\pairs near boundaries still correlated \\if blocks are too small};
    \node[annot, midorange, font=\sffamily\scriptsize\bfseries] at (2.1, -0.95) {$\rightarrow$ Better, but potential edge leakage};
\end{scope}

\begin{scope}[shift={(2*\panelshift, 0)}]
    \node[heading] at (2.1, 3.2) {(c) Blocks + buffer zone};

    \fill[black, opacity=0.05] \landpath;
    \draw[black!20, line width=0.6pt] \landpath;

    \begin{scope}
        \clip \landpath;
        \fill[foldA, opacity=0.12] (0, 1.75) rectangle (1.85, 2.8);
        \fill[foldB, opacity=0.12] (2.35, 1.75) rectangle (4.2, 2.8);
        \fill[foldC, opacity=0.12] (0, 0.1) rectangle (1.85, 1.25);
        \fill[foldD, opacity=0.12] (2.35, 0.1) rectangle (4.2, 1.25);
        \fill[pattern=north east lines, pattern color=black!15]
            (1.85, 0.1) rectangle (2.35, 2.8);   
        \fill[pattern=north east lines, pattern color=black!15]
            (0, 1.25) rectangle (4.2, 1.75);      
    \end{scope}

    \fill[foldA, opacity=0.75] (0.5, 2.2) circle (\R);
    \fill[foldA, opacity=0.75] (1.0, 2.0) circle (\R);
    \fill[foldA, opacity=0.75] (0.7, 1.85) circle (\R);
    \fill[foldA, opacity=0.75] (1.4, 2.3) circle (\R);
    \fill[foldA, opacity=0.75] (1.6, 2.1) circle (\R);
    \fill[foldC, opacity=0.75] (0.4, 1.0) circle (\R);
    \fill[foldC, opacity=0.75] (0.9, 0.7) circle (\R);
    \fill[foldC, opacity=0.75] (1.3, 0.5) circle (\R);
    \fill[foldC, opacity=0.75] (0.65, 0.85) circle (\R);
    \fill[foldC, opacity=0.75] (1.1, 0.4) circle (\R);
    \fill[foldB, opacity=0.75] (2.7, 2.2) circle (\R);
    \fill[foldB, opacity=0.75] (3.2, 2.0) circle (\R);
    \fill[foldB, opacity=0.75] (2.9, 1.85) circle (\R);
    \fill[foldB, opacity=0.75] (3.6, 2.35) circle (\R);
    \fill[foldD, opacity=0.75] (2.6, 0.7) circle (\R);
    \fill[foldD, opacity=0.75] (3.0, 0.9) circle (\R);
    \fill[foldD, opacity=0.75] (3.5, 0.65) circle (\R);
    \fill[foldD, opacity=0.75] (2.8, 0.4) circle (\R);
    \fill[foldD, opacity=0.75] (3.3, 0.5) circle (\R);

    \foreach \px/\py in {1.9/1.6, 2.25/2.05, 1.7/1.3, 2.3/1.35, 3.1/1.45} {
        \fill[black, opacity=0.1] (\px, \py) circle (\R);
        \draw[black!35, line width=0.5pt]
            (\px-0.065, \py-0.065) -- (\px+0.065, \py+0.065);
        \draw[black!35, line width=0.5pt]
            (\px+0.065, \py-0.065) -- (\px-0.065, \py+0.065);
    }

    \node[sub] at (2.1, -0.15) {Buffer excludes near-boundary\\samples from all folds};
    \node[annot, goodgreen, font=\sffamily\scriptsize\bfseries] at (2.1, -0.65) {$\rightarrow$ Realistic evaluation};
\end{scope}

\begin{scope}[shift={(0, -1.4)}]
    \fill[trainblue, opacity=0.75] (1.5, 0) circle (0.1);
    \node[legendlabel] at (1.55, 0) {Train};
    \fill[testred, opacity=0.75] (2.5, 0) circle (0.1);
    \node[legendlabel] at (2.55, 0) {Test};

    \fill[foldA, opacity=0.75] (5.5, 0) circle (0.1);
    \node[legendlabel] at (5.55, 0) {Fold 1};
    \fill[foldB, opacity=0.75] (7.0, 0) circle (0.1);
    \node[legendlabel] at (7.05, 0) {Fold 2};
    \fill[foldC, opacity=0.75] (8.5, 0) circle (0.1);
    \node[legendlabel] at (8.55, 0) {Fold 3};
    \fill[foldD, opacity=0.75] (10, 0) circle (0.1);
    \node[legendlabel] at (10.05, 0) {Fold 4};

    \fill[pattern=north east lines, pattern color=black!15]
        (11.5, -0.12) rectangle (11.85, 0.12);
    \draw[black!25, line width=0.3pt] (11.5, -0.12) rectangle (11.85, 0.12);
    \node[legendlabel] at (11.95, 0) {Buffer};
    
    \fill[black, opacity=0.1] (13.0, 0) circle (0.08);
    \draw[black!35, line width=0.4pt] (12.94, -0.06) -- (13.06, 0.06);
    \draw[black!35, line width=0.4pt] (13.06, -0.06) -- (12.94, 0.06);
    \node[legendlabel] at (13.16, 0) {Excluded};

\end{scope}


\end{tikzpicture}
\end{document}

%% file: uq_flowchart.tikz
\begin{tikzpicture}[
    >={Stealth[length=3mm]},
    question/.style={
        rectangle, rounded corners=3pt, draw=qborder, fill=qgray,
        text width=4.4cm, minimum height=1.1cm, align=center, font=\small
    },
    method/.style={
        rectangle, rounded corners=3pt, draw=#1, fill=#1!12,
        text width=4.4cm, minimum height=1cm, align=center, font=\small
    },
    method/.default=dkblue,
    secref/.style={font=\scriptsize\itshape, text=dkred},
    elabel/.style={font=\scriptsize, fill=white, inner sep=2pt}
]

\definecolor{dkblue}{HTML}{2B4C7E}
\definecolor{ltblue}{HTML}{D6E4F0}
\definecolor{dkgreen}{HTML}{2E6B4F}
\definecolor{ltgreen}{HTML}{D4EDDA}
\definecolor{dkorange}{HTML}{8B5E3C}
\definecolor{ltorange}{HTML}{FDEBD0}
\definecolor{dkred}{HTML}{7B2D3B}
\definecolor{ltred}{HTML}{F5D6D6}
\definecolor{dkpurple}{HTML}{5B3A6E}
\definecolor{ltpurple}{HTML}{E8D5F5}
\definecolor{qgray}{HTML}{F5F5F5}
\definecolor{qborder}{HTML}{888888}

\node[question, draw=dkblue, line width=1pt, fill=ltblue, text width=5.5cm, minimum height=1.3cm] 
    (start) {\textbf{What will you do with\\the uncertainty?}};
 
\node[question] (mask) at (-8, -3) {Mask unreliable pixels\\before release or aggregation};
\node[question] (report) at (-2.7, -3) {Report calibrated\\prediction intervals};
\node[question] (aggregate) at (3.2, -3) {Estimate uncertainty\\over spatial aggregates};
\node[question] (change) at (8.5, -3) {Estimate uncertainty\\for temporal change};
 
\node[method=dkgreen] (mask_method) at (-8, -5.8) 
    {\textbf{Any rough proxy}\\[2pt]
    Heteroskedastic head,\\ensemble spread, or\\area of applicability};
\node[secref] at (-8, -7.25) {Calibration desirable, not required};
 
\node[question] (budget) at (-2.7, -5.8) {What is your\\compute budget?};
 
\node[method=dkorange] (cheap) at (-5, -8.8) 
    {\textbf{Low budget}\\[2pt]
    Conformalized quantile\\regression (CQR)};
\node[secref] at (-5, -10) {Post-hoc; marginal coverage only};
 
\node[method=dkpurple] (expensive) at (-0.2, -8.8) 
    {\textbf{Higher budget}\\[2pt]
    Gaussian/neural process, ensembles + spatial block};
\node[secref] at (-0.2, -10) {Verify calibration};
 
\node[method=dkred] (agg_method) at (3.2, -5.8)
    {\textbf{Spatial error correlation}\\[2pt]
    Block kriging on model\\residuals; pixel-level UQ\\alone is insufficient};
\node[secref] at (3.2, -7.25) {Wadoux \& Heuvelink (2023)};
 
\node[method=dkblue] (change_method) at (8.5, -5.8) 
    {\textbf{Propagate through\\differencing}\\[2pt]
    Temporal error correlation between epochs};
\node[secref] at (8.5, -7.25) {correlated vs.\ independent errors};
 
\draw[thick, ->] (start.south) -- ++(0,-0.6) -| (mask.north);
\node[elabel, above=0.1cm of mask] {Filtering};
\draw[thick, ->] (start.south) -- ++(0,-0.6) -| (report.north);
\node[elabel, above=0.1cm of report] {Reporting};
\draw[thick, ->] (start.south) -- ++(0,-0.6) -| (aggregate.north);
\node[elabel, above=0.1cm of aggregate] {Aggregation};
\draw[thick, ->] (start.south) -- ++(0,-0.6) -| (change.north);
\node[elabel, above=0.1cm of change] {Change};

\draw[thick, ->] (mask) -- (mask_method);
\draw[thick, ->] (report) -- (budget);
\draw[thick, ->] (budget.south) -- ++(0,-0.6) -| (cheap.north)
    node[elabel, pos=0.25, above] {Low};
\draw[thick, ->] (budget.south) -- ++(0,-0.6) -| (expensive.north)
    node[elabel, pos=0.25, above] {High};
\draw[thick, ->] (aggregate) -- (agg_method);
\draw[thick, ->] (change) -- (change_method);
 
\draw[thick, dashed, ->, dkred!55] 
    (agg_method.west) -- (budget.east)
    node[elabel, midway, above=2pt, text=dkred!70] {\emph{+ pixel-level UQ}};
 
\end{tikzpicture}

%% file: patches.tex
\begin{tikzpicture}[
    >=Stealth,
    every node/.style={font=\sffamily\small},
    patch/.style={rounded corners=1pt, line width=1pt},
    annot/.style={font=\sffamily\scriptsize, text opacity=0.7},
    heading/.style={font=\sffamily\small\bfseries, text opacity=0.8},
]

\node[heading, font=\sffamily\bfseries] at (2.8, 5.8) {(a) Overlapping patches + blending};
\node[annot, font=\sffamily\scriptsize\bfseries] at (1.5, 5.2) {1.\ Overlapping grid};
\fill[patchA, opacity=0.12] (0, 1.7) rectangle (2.8, 4.5);
\draw[patchA, patch] (0, 1.7) rectangle (2.8, 4.5);
\node[annot, patchA] at (1.1, 3.0) {Patch $i$};
\fill[patchB, opacity=0.12] (1.8, 1.7) rectangle (4.6, 4.5);
\draw[patchB, patch] (1.8, 1.7) rectangle (4.6, 4.5);
\node[annot, patchB] at (3.5, 3.0) {Patch $i{+}1$};
\fill[black, opacity=0.04] (1.8, 1.7) rectangle (2.8, 4.5);
\draw[decorate, decoration={brace, amplitude=4pt, mirror}, black!40, line width=0.6pt]
    (1.8, 1.55) -- (2.8, 1.55);
\node[annot, black!50] at (2.3, 1.2) {overlap};
\draw[->, black!30, line width=1pt] (4.9, 3.1) -- (5.6, 3.1);
\node[annot, font=\sffamily\scriptsize\bfseries] at (7.1, 5.2) {2.\ Weighted merge};
\draw[black!25, line width=0.8pt, rounded corners=1pt] (5.8, 1.7) rectangle (8.6, 4.5);
\foreach \i in {0,...,27} {
    \pgfmathsetmacro{\x}{5.8 + \i * 0.1}
    \pgfmathsetmacro{\opA}{max(0, 0.18 - \i * 0.006)}
    \pgfmathsetmacro{\opB}{max(0, \i * 0.006 - 0.0)}
    \fill[patchA, opacity=\opA] (\x, 1.7) rectangle (\x+0.1, 4.5);
    \fill[patchB, opacity=\opB] (\x, 1.7) rectangle (\x+0.1, 4.5);
}
\draw[black!15, line width=0.4pt] (5.8, 1.2) -- (8.6, 1.2);
\draw[patchA, opacity=0.6, line width=1pt]
    (5.8, 1.5) .. controls (6.5, 1.45) and (7.5, 1.25) .. (8.6, 1.2);
\draw[patchB, opacity=0.6, line width=1pt]
    (5.8, 1.2) .. controls (7.0, 1.25) and (7.8, 1.45) .. (8.6, 1.5);
\node[annot, black!45] at (7.2, 0.85) {weight profiles (e.g.,\ cosine)};

\node[annot, anchor=west, kept] at (0, 0.8) {\textbullet\ Smooth transitions};
\node[annot, anchor=west, black!60] at (0, 0.35) {\textbullet\ Blurs predictions in overlap zone};
\node[annot, anchor=west, black!60] at (0, -0.1) {\textbullet\ Complicates uncertainty propagation};

\node[heading, font=\sffamily\bfseries] at (13.5, 5.8) {(b) Padding + crop to center};

\node[annot, font=\sffamily\scriptsize\bfseries] at (11.2, 5.2) {1.\ Expand with context};

\fill[black, opacity=0.03] (9.8, 1.7) rectangle (12.8, 4.7);
\draw[black!20, line width=0.6pt, dashed, rounded corners=1pt] (9.8, 1.7) rectangle (12.8, 4.7);

\fill[kept, opacity=0.1] (10.4, 2.2) rectangle (12.2, 4.2);
\draw[kept, line width=1.2pt, rounded corners=1pt] (10.4, 2.2) rectangle (12.2, 4.2);

\node[annot, kept] at (11.3, 3.2) {patch};
\node[annot, black!40] at (11.3, 4.5) {padding};

\draw[->, black!30, line width=1pt] (13.1, 3.2) -- (13.8, 3.2);

\node[annot, font=\sffamily\scriptsize\bfseries] at (15.5, 5.2) {2.\ Predict $\rightarrow$ crop};

\fill[black, opacity=0.03] (14.1, 1.7) rectangle (17.1, 4.7);
\draw[black!15, line width=0.6pt, dashed, rounded corners=1pt] (14.1, 1.7) rectangle (17.1, 4.7);

\draw[cropred, line width=1.2pt, dashed] (14.7, 1.5) -- (14.7, 4.9);
\draw[cropred, line width=1.2pt, dashed] (16.5, 1.5) -- (16.5, 4.9);
\draw[cropred, line width=1.2pt, dashed] (13.9, 2.2) -- (17.3, 2.2);
\draw[cropred, line width=1.2pt, dashed] (13.9, 4.2) -- (17.3, 4.2);

\fill[kept, opacity=0.1] (14.7, 2.2) rectangle (16.5, 4.2);
\draw[kept, line width=1.2pt, rounded corners=1pt] (14.7, 2.2) rectangle (16.5, 4.2);

\node[annot, kept] at (15.6, 3.2) {kept};
\node[annot, cropred, opacity=0.7] at (15.6, 4.5) {discarded};

\node[annot, anchor=west, kept] at (9.8, 0.8) {\textbullet\ No blending --- preserves model output};
\node[annot, anchor=west, kept] at (9.8, 0.35) {\textbullet\ Uncertainty maps remain faithful};
\node[annot, anchor=west, black!60] at (9.8, -0.1) {\textbullet\ Residual artifacts if receptive field $<$ padding};

\end{tikzpicture}

%% file: divedeeper.tex
\section{Sentinel-1 data gaps}
\label{app:s1-gaps}

\begin{figure}[!b]
\centering
\includegraphics[trim={1cm 3cm 1cm 3cm},clip,width=\columnwidth]{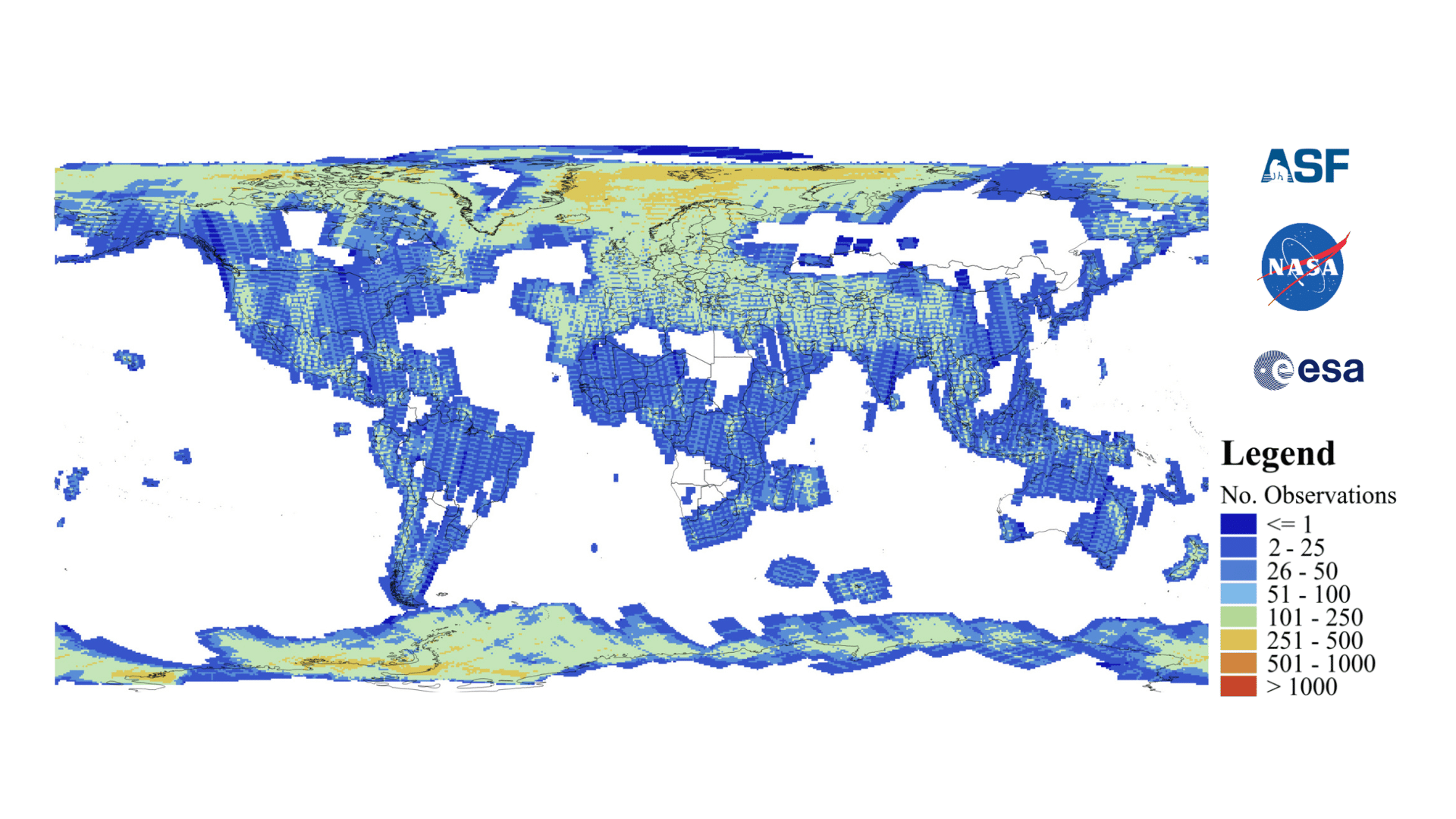}
\caption{Sentinel-1 Level 1 GRD data coverage, as provided by the Alaska Satellite Facility \citep{ASFS1}. The map shows the number of Sentinel-1A and Sentinel-1B GRD observations acquired globally, in April 2024.}
\label{fig:s1-coverage}
\end{figure}

The failure of the Sentinel-1B sensor reduced the number of available radar images and introduced gaps in global coverage (see Figure~\ref{fig:s1-coverage}) from December 2021 until Sentinel-1C became operational in early 2025. This loss was particularly significant because synthetic aperture radar (SAR) can image through clouds and in darkness, making it an invaluable complement to optical imagery, especially in persistently cloud-covered regions such as Gabon. Furthermore, SAR provides sensing capabilities unavailable from other instruments and has proven particularly useful for various tasks, such as flood detection \citep{Misra2025}, vegetation monitoring \citep{PARAG2024101107}, or ocean object detection \citep{10234538}. Rather than viewing these data gaps as a limitation, we encourage readers to explore complementary or alternative SAR sources (such as ALOS PALSAR/PALSAR-2 yearly mosaics) and to develop models capable of handling multiple or missing modalities (see Section~\ref{subsec:model-design}). 

\section{Sentinel-1 and Sentinel-2 processing across access pathways}
\label{app:processing-pathways}

The same nominal Sentinel-1 or Sentinel-2 product can differ depending on where it is accessed. CDSE distributes the authoritative ESA outputs through three principal interfaces (direct download, Sentinel Hub, and openEO) each exposing a different subset of preprocessing options. Google Earth Engine (GEE) ingests from CDSE but applies its own preprocessing chain, particularly for Sentinel-1, and refreshes its catalog on its own schedule. This appendix summarizes the resulting differences and discusses aspects that are not self-evident from the tables alone.

\subsection{Sentinel-2}
\label{app:s2-pathways}

For Sentinel-2, the differences between CDSE and GEE are summarized in Table~\ref{tab:s2-pathways}. The atmospheric correction is identical between the two platforms, so observed differences in surface reflectance might arise from baseline currency or radiometric offset handling.

\begin{table}[!t]
\centering
\caption{Processing differences between native Sentinel-2 products on CDSE and the corresponding GEE collections. DN: digital number. SR: surface reflectance.}
\label{tab:s2-pathways}
\footnotesize
\renewcommand{\arraystretch}{1.2}
\begin{tabularx}{\columnwidth}{@{}P{1.9cm}LL@{}}
\toprule
\textbf{Aspect} & \textbf{CDSE (native)} & \textbf{GEE} \\
\midrule
Radiometric offset (post PB 04.00) & $+1000$ DN baked in & Subtracted in \texttt{HARMONIZED} collections; passed through unchanged in non-harmonized collections \\
Atmospheric correction & Sen2Cor & Same (ingested from CDSE) \\
Processing baseline currency & Authoritative; updated on ESA's schedule & May lag CDSE during reprocessing campaigns \\
QA60 cloud mask & Discontinued by ESA after February 2022 & Reconstructed from classification data starting February 2024 \\
\bottomrule
\end{tabularx}
\end{table}

The most consequential subtlety beyond the table is the \emph{timing of baseline reprocessing}. CDSE has carried out a Collection-1 reprocessing campaign in stages. Phase~1 brought all data through the end of 2021 to baseline 05.00 (which uses the 30\,m Copernicus DEM in place of the older 90\,m one) and harmonizes radiometry between Sentinel-2A and Sentinel-2B. Phase~2 reprocessed January 2022 through December 2023 to baseline 05.10. Baseline 05.11 (July 2024) introduced further refinements to geometric registration, cloud screening (with parallax handling), and snow/ice detection, and renamed the SCL class \texttt{DARK\_FEATURES} to \texttt{CAST\_SHADOW}. Because GEE ingests from CDSE, it inherits these improvements but with a refresh lag whose duration depends on the granule volume. During such windows, CDSE and GEE may serve different baseline versions of the same granule.

The \emph{QA60 reconstruction} introduced by GEE in February 2024 is similarly worth flagging. ESA discontinued the legacy polygon-based cloud mask after February 2022, leaving a two-year gap in QA60 availability on GEE. The reconstructed band, derived from classification outputs, is intended to preserve backward compatibility with pipelines that depend on QA60 but is not part of the native CDSE distribution and should not be assumed to be bit-identical to pre-2022 QA60 values.

\subsection{Sentinel-1}
\label{app:s1-pathways}

For Sentinel-1, the differences are more substantial because GEE applies a full preprocessing chain (thermal noise removal, data calibration, multi-looking and range-doppler terrain correction) rather than redistributing the GRD product, and the three CDSE pathways themselves expose different defaults and degrees of configurability. Table~\ref{tab:s1-pathways} summarizes the four options side by side.

\begin{table*}[!t]
\centering
\caption{Sentinel-1 GRD processing options across CDSE access pathways and GEE. ``Configurable'' indicates that the option is exposed through the API and can be enabled or disabled by the user.}
\label{tab:s1-pathways}
\footnotesize
\renewcommand{\arraystretch}{1.2}
\begin{tabular}{@{}P{0.20\textwidth}P{0.16\textwidth}P{0.20\textwidth}P{0.16\textwidth}P{0.18\textwidth}@{}}
\toprule
 & \multicolumn{3}{c}{\textbf{CDSE}} & \multirow{2}{=}{\textbf{Google Earth Engine}} \\
\cmidrule(lr){2-4}
 & \textbf{OData} & \textbf{Sentinel Hub} & \textbf{openEO} & \\
\midrule
Backscatter convention & User choice ($\beta^0/\sigma^0/\gamma^0$) & Configurable (default $\gamma^0$ ellipsoid) & $\sigma^0$ ellipsoid (fixed) & $\sigma^0$ only \\
Thermal noise removal & $\times$ & \checkmark & \checkmark & \checkmark \\
Speckle filtering & $\times$ & Configurable & $\times$ & $\times$ \\
Orthorectification & $\times$ & Configurable, with Range--Doppler terrain correction & \checkmark & \checkmark \\
DEM for orthorectification & User choice & Copernicus 10\,m / 30\,m / 90\,m, or Mapzen (regional defaults differ) & Copernicus 30\,m & SRTM 30\,m, with ASTER fallback above $\pm 60^\circ$ latitude \\
Radiometric terrain correction & $\times$ & Available via \texttt{gamma0\_terrain} + orthorectification & $\times$ & $\times$ \\
Output units & DN & Linear power; dB via evalscript & Linear power & dB (\texttt{S1\_GRD}) or linear power (\texttt{S1\_GRD\_FLOAT}) \\
SLC products & \checkmark & $\times$ & $\times$ & $\times$ \\
\bottomrule
\end{tabular}
\end{table*}

Several details in Table~\ref{tab:s1-pathways} have downstream consequences worth flagging. The default backscatter coefficient on Sentinel Hub is $\gamma^0$ ellipsoid, which contrasts with both the openEO default ($\sigma^0$ ellipsoid) and GEE's $\sigma^0$-only output. Radiometric terrain correction on Sentinel Hub is not selected by a single flag: it requires setting \texttt{backCoeff} to \texttt{gamma0\_terrain} \emph{and} enabling orthorectification, with users selecting \texttt{gamma0\_ellipsoid} alone obtaining only the ellipsoid-referenced coefficient. The DEM defaults on Sentinel Hub also differ between deployment regions (Copernicus 30\,m in EU-Central-1, Mapzen in US-West-2), so reproducibility across deployments requires setting \texttt{demInstance} explicitly.

The DEM choice is also where the most significant divergence between Sentinel Hub and GEE arises. GEE uses SRTM 30\,m, with an ASTER fallback above $\pm 60^\circ$ latitude, while Sentinel Hub recommends the Copernicus DEM. The resulting geometric and radiometric differences are typically small over flat terrain but can become significant in regions with strong relief, especially when co-locating Sentinel-1 with optical imagery.

Speckle filtering is not applied by default on any pathway: it is left to the user. GEE provides only GRD-level data, so any workflow requiring phase information for interferometry must use CDSE direct download (or another canonical provider such as ASF), as SLC products are not available through Sentinel Hub, openEO, or GEE.

\paragraph{Sentinel-1 RTC} CDSE also distributes a Sentinel-1 Radiometric Terrain Corrected (RTC) backscatter product, processed from GRD inputs and compliant with the CEOS Analysis Ready Data (CEOS-ARD, formerly CARD4L) specifications for Normalized Radar Backscatter (NRB). RTC products are orthorectified using the Copernicus DEM and delivered without speckle filtering. The NRB standard~--- which combines range--Doppler geocoding with radiometric terrain correction~--- is increasingly adopted across providers and may eventually supersede GRD as the default Sentinel-1 distribution format for analysis-ready use cases.

\subsection*{Further reading}

ESA's Sentinel-2 User Handbook \citep{esas2handbook} and the Sen2Cor Algorithm Theoretical Basis Document \citep{sen2coratbd} provide the authoritative description of the L2A processing chain. The Sentinel Hub Sentinel-1 GRD processing documentation \citep{sentinelhubs1grd} details the full set of \texttt{demInstance} and \texttt{backCoeff} options exposed through the API. The GEE Sentinel-1 algorithm \citep{gees1grd} documents the preprocessing applied to the GRD collections. \citet{Mullissa2021} provide a community-validated reference implementation for analysis-ready Sentinel-1 preprocessing on GEE, which is the recommended starting point for users who require more control than the default GEE collections provide.

\section{Identifying Sentinel-2 tiling artifacts}
\label{app:s2-artifacts}

Section~\ref{subsubsec:optical} describes the general problem: a product's distribution unit need not align with its acquisition unit, so a nominally complete tile may be partial, may need mosaicking, or may be duplicated. For Sentinel-2 the three cases are distinguished from the product metadata as follows.

\begin{figure}[!t]
\centering
\includegraphics[trim={0 8cm 0 0},clip,width=\columnwidth]{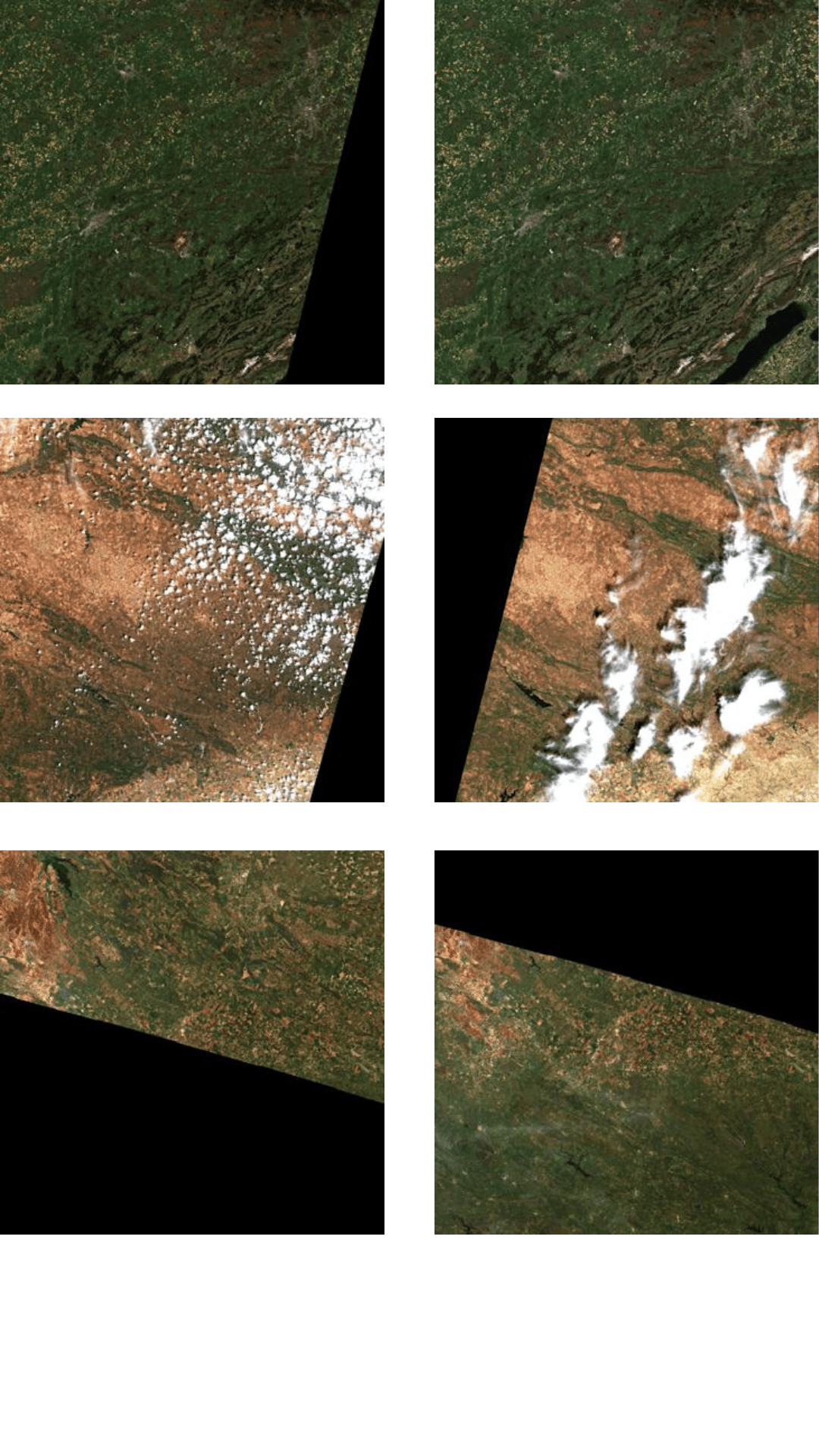}
\caption{Illustration of Sentinel-2 artifacts. \textit{Top}: Sentinel-2 L2A products for tile 31TGN, from sensor A, orbit 008 (left) and sensor B, orbit 108 (right) for April 8 2025. \textit{Middle}: Sentinel-2 L2A products for tile 30SUH, sensor B on June 4 2025, orbit 137  (left) vs. sensor C on June 6 2025, orbit 094 (right). \textit{Bottom}: Sentinel-2C L2A products for tile 30STH, acquired on April 5 2026, in the same orbit (137). \textit{The images were acquired using the Copernicus Browser.}}
\label{fig:s2-combined}
\end{figure}

\paragraph{Partial and full orbit coverage} Each tile is covered by multiple orbits, some of which encompass it entirely. In Figure~\ref{fig:s2-combined} (top), tile 31TGN is partially covered by orbit 008 and fully by orbit 108. The covering orbit is identified directly from the \texttt{MTD\_MSIL2A.xml} file: \texttt{NODATA\_PIXEL\_PERCENTAGE} is 0.0 for the orbit that fully covers the tile and greater than 0.0 for the others.

\paragraph{Tiles requiring mosaicking} Most tiles are not fully covered by any single orbit and require mosaicking products from several orbits, which may have been acquired on different days. Tile 30SUH is typical: two products acquired two days apart are needed for complete coverage (Figure~\ref{fig:s2-combined}, middle).

\paragraph{Duplicate products} A tile that should be fully covered by one orbit sometimes yields two products sharing date, orbit and footprint. If the processing baselines differ, retain the higher one. If they are identical, the duplication indicates a data strip split: a single continuous observation broken into separate segments during ground processing, slicing the tile into two partial files that must be mosaicked (Figure~\ref{fig:s2-combined}, bottom).

\section{Selecting a SAR viewing geometry}
\label{app:sar-geometry}

Section~\ref{subsubsec:sar} notes that data volume and the need for geometric consistency (required for analyses such as interferometric stacks, amplitude change detection, or feature extraction) push toward selecting a single track. For Sentinel-1 in Interferometric Wide (IW) mode, three factors define this viewing geometry: the relative orbit number or track (the satellite's repeating ground path), the sub-swath (which sets the incidence angle range), and the pass direction (ascending or descending). Ascending and descending acquisitions of the same area are complementary (slopes that suffer layover under one geometry are often well-imaged under the other) and combining the two is a standard strategy for filling distortion gaps in mountainous regions. More generally, several tracks are typically available over any target area, each contributing hundreds of acquisitions over the mission's lifetime. ESA's Sentinel-1 Burst ID Map \citep{sarmpc2022burstid} provides the metadata needed to make this choice. It consists of global polygons that divide each sub-swath into smaller units called bursts, each annotated with its track and sub-swath identifier. Together with the pass direction, these identifiers specify the viewing geometry for any given location, and three approaches of increasing detail can be used to select among the candidates. The simplest uses the burst grid alone: candidates are filtered by sub-swath, typically with a preference for mid-range incidence angles, which is sufficient for flat or moderate terrain. For rougher terrain, combining the burst grid with a digital elevation model allows candidates to be ranked by the alignment between the radar's look direction and the dominant slope aspect. For a more exact comparison, tools such as \href{https://step.esa.int/main/wp-content/help/versions/9.0.0/snap-toolboxes/org.esa.s1tbx.s1tbx.op.sar.processing.ui/operators/SARSimulationOp.html}{SNAP's SAR Simulation Operator} can simulate the per-pixel layover and shadow masks that would arise from each candidate, though the per-candidate simulation cost makes this approach impractical at scale.

\section{Radiometric calibration and terrain correction of SAR data}
\label{app:sar-rtc}

This appendix expands on the radiometric processing chain summarized in Section~\ref{subsubsec:sar-processing}, focusing on the choice of backscatter convention and the role of terrain in radiometric normalization, following the taxonomy of \citet{Small2011}. 

\begin{figure}[!t]
    \centering
    \includegraphics[width=0.9\linewidth]{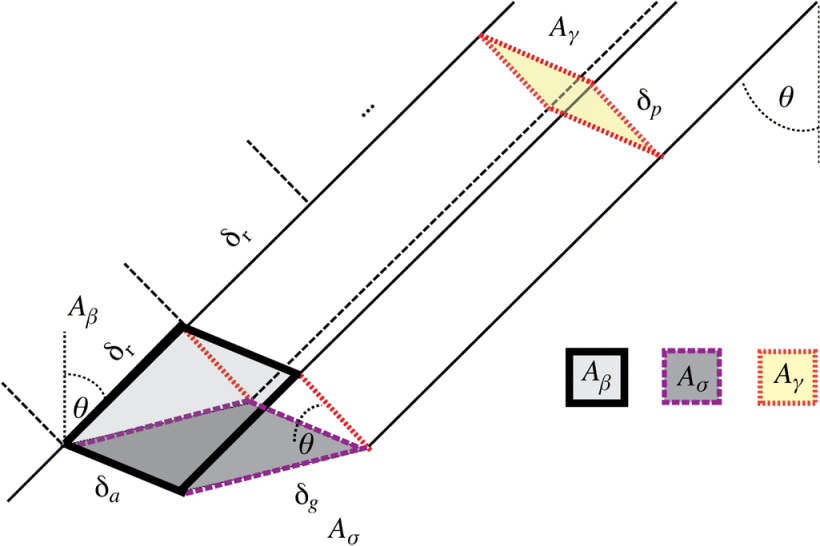}
    \caption{Normalization areas for SAR backscatter. Reproduced from \citep{Small2011}.}
    \label{fig:backscatter}
\end{figure}

\paragraph{From digital numbers to calibrated backscatter} A SAR system records a complex signal with an amplitude, indicating the amount of signal returning to the sensor, and a phase. Detected products such as GRD keep only the amplitude, stored as digital numbers (DN); intensity is its square. Radiometric calibration converts DN to a physically meaningful backscatter coefficient by normalizing by a reference area to obtain backscatter per unit area. The simplest calibrated quantity is the radar brightness $\beta^0$, defined as the calibrated intensity divided by the slant-range pixel area $A_\beta$ (see Figure \ref{fig:backscatter}). Because $\beta^0$ assumes no Earth model, it is a property of the radar measurement alone and forms the conceptual root from which all other conventions are derived \citep{Raney1994}.

\paragraph{Ellipsoid-referenced conventions} Two conventions normalize $\beta^0$ by an area defined under a flat-ellipsoid assumption. The $\sigma^0_E$ convention divides by the ground-projected area $A_\sigma$, giving $\sigma^0_E = \beta^0 \cdot \frac{A_{\beta}}{A_{\sigma}} = \beta^0 \cdot \sin\theta_E$, where $\theta_E$ is the incidence angle with respect to the ellipsoid. The $\gamma^0_E$ convention divides by the area perpendicular to the line of sight $A_\gamma$, giving $\gamma^0_E = \beta^0 \cdot \tan\theta_E$. Note that a model of the Earth is required to gain knowledge of the incident angle $\theta_E$ to calculate $\sigma^0$ or $\gamma^0$.

\paragraph{The problem with the ellipsoid assumption} None of $\beta^0$, $\sigma^0_E$, or $\gamma^0_E$ accounts for local relief: the areas they normalize by are derived from the ellipsoid, but the true illuminated area of a pixel depends on the terrain. A slope tilted toward the sensor exposes more surface area than the ellipsoid assumes and therefore appears brighter; a slope tilted away exposes less and appears darker. Pixel intensity becomes partly a function of topography rather than ground properties, biasing any task that compares backscatter across relief: land cover classification, biomass retrieval, change detection, or the mosaicking of ascending and descending passes.

\paragraph{Terrain-referenced conventions} Radiometric terrain correction (RTC) replaces the ellipsoid-derived reference area with one computed from a digital elevation model (DEM). Two terrain-referenced conventions are in common use. The $\sigma^0_T$ convention applies a per-pixel scalar correction based on the local incidence angle (LIM), $\sigma^0_T = \sigma^0_E \cdot \sin\theta_{\text{LIM}}/\sin\theta_E$, where $\theta_{\text{LIM}}$ is derived from the DEM. This correction assumes that a single planar facet contributes to each pixel and only adjusts for its tilt. \citet{Small2011} argues that such angle-based normalizations are flawed due to their failure to account for many important properties of radar backscatter in regions with significant topographic variation. He introduced the $\gamma^0_T$ convention that instead integrates the illuminated facet areas rigorously over the DEM, accumulating contributions from all terrain facets that map to a given SAR pixel. This handles cases of steep or complex relief where multiple facets contribute per pixel, which a local-angle scalar correction cannot represent. $\gamma^0_T$ is therefore preferred over $\sigma^0_T$ whenever significant relief is present, and is the default terrain-flattened output of modern processors, such as ESA SNAP \citep{esasnap}. 

\paragraph{Relationship to geometric terrain correction} Radiometric terrain correction is distinct from geometric terrain correction (GTC), which uses a DEM and the Range-Doppler algorithm to relocate pixels from radar coordinates to map coordinates. GTC corrects \emph{where} pixels sit; RTC corrects \emph{what their values mean}. The two operations share the same geometric mapping between SAR and DEM coordinates and are typically computed together, but they are conceptually independent: an image can be geometrically correct while remaining radiometrically biased on slopes, and backscatter can be terrain-flattened in radar geometry before being geocoded. GTC alone is adequate for visual interpretation over relatively flat terrain; quantitative analysis across significant relief generally requires both corrections, with $\gamma^0_T$ as the natural input convention for any downstream task that compares backscatter across topographically varied scenes.

\section{Analysis-ready SAR products}
\label{app:sar-ard}
 
The four corrections described in Section~\ref{subsubsec:sar-processing} can be inherited rather than performed. Two products apply them over the whole globe and distribute the result free of charge, and comparing them is instructive: both arrive at terrain-corrected $\gamma^0$, both are defensible, and they answer the four questions differently at every stage (Table~\ref{tab:sar-ard-products}).
 
\paragraph{Sentinel-1 monthly mosaics} The Copernicus Data Space Ecosystem publishes global Sentinel-1 mosaics as a Level-3 product, built by calibrating GRD inputs to $\beta^0$, removing thermal noise, flattening to $\gamma^0_T$ by integrating the illuminated facet areas over the Copernicus DEM, and applying Range-Doppler geometric correction against the same DEM. Each month is then reduced to a weighted mean in which observations at higher local resolution carry more weight \citep{Small2012} and radar-shadow observations are discarded. Because the compositing is what suppresses speckle, the product commits its user to a monthly cadence.
 
\paragraph{ALOS PALSAR/PALSAR-2 mosaics} Beyond Sentinel-1, another widely used SAR data source is the ALOS PALSAR/PALSAR-2 annual mosaic, a seamless global L-band SAR image created by mosaicking strips of PALSAR and PALSAR-2 imagery. Each strip is orthorectified and slope-corrected using the ALOS World 3D--30\,m digital surface model, and polarization data can be converted to $\gamma^0$ values. Because these mosaics are delivered as annual composites with terrain and radiometric corrections already applied, they require less user-side preprocessing than Sentinel-1 GRD products.
 
\begin{table*}[!t]
\centering
\caption{Two analysis-ready SAR products, and the four corrections of Section~\ref{subsubsec:sar-processing} as each of them resolves them. Neither is wrong; adopting either means adopting its answers.}
\label{tab:sar-ard-products}
\footnotesize
\renewcommand{\arraystretch}{1.2}
\begin{tabularx}{\textwidth}{@{}>{\raggedright\arraybackslash}p{2.6cm}LL@{}}
\toprule
 & \textbf{Sentinel-1 monthly mosaics} (CDSE)
 & \textbf{PALSAR/PALSAR-2 yearly mosaic} (JAXA) \\
\midrule
 
Thermal noise
 & Thermal noise removal applied.
 & Not documented as a separate step; radiometric balancing between adjacent
   paths is derived from their overlap regions, over land only. \\[3pt]
 
Radiometric calibration
 & Calibrated to $\beta^0$, then flattened to $\gamma^0_T$ by integrating the
   illuminated facet areas over the DEM.
 & Radiometric slope correction, with $\gamma^0$ recovered from the stored
   digital numbers as $10\log_{10}(\mathrm{DN}^2) - 83.0$\,dB. \\[3pt]
 
Speckle reduction
 & Temporal averaging: a weighted monthly mean, weights set by each
   observation's local resolution \citep{Small2012}, radar-shadow observations
   discarded.
 & Power averaged over neighboring pixels; the annual composite is assembled
   from strips selected by visual inspection, favoring those with the least
   surface-moisture response. \\[3pt]
 
Terrain correction
 & Radiometric correction as above, plus Range-Doppler geometric correction,
   both against the Copernicus DEM --- 10\,m over Europe, 30\,m elsewhere.
 & Ortho-rectified and slope-corrected against the ALOS World 3D 30\,m surface
   model (AW3D30). \\
 
\midrule
 
Grid and resolution
 & 20\,m for IW (VV, VH), 40\,m for DH (HH, HV); 16-bit cloud-optimized GeoTIFF
   on a UTM grid.
 & 25\,m; HH and HV (VV and VH over Japan); $1^\circ\times1^\circ$ tiles in
   geographic coordinates, with layover and shadow flagged in a quality
   band. \\[3pt]
 
Cadence and coverage
 & Monthly, global, 2020 onward.
 & Annual, global, 2007--2010 and 2015 onward; no mosaic exists for
   2011--2014, between the two missions. \\
 
\bottomrule
\end{tabularx}
\end{table*}
 
\paragraph{What is inherited} The case for starting from either is the case for analysis-ready data generally: the work is done once, by people who know the sensor, and the result is consistent across space and time in a way an in-house pipeline rarely is. What is given up is granularity. The cadence is fixed, so the temporal decision of Section~\ref{subsubsec:temporal-distribution} is made on the user's behalf and individual acquisitions --- with their orbit direction and incidence angle --- are no longer addressable. The normalization is fixed too, and the two products differ in kind rather than in quality here: the Sentinel-1 mosaics integrate facet areas over the DEM, while the PALSAR mosaic applies a slope correction and a fixed calibration factor.

\section{Cloud masking methods for optical data}
\label{app:s2-cloud-masking}

Cloud and cloud-shadow detection is the single most consequential preprocessing step for any optical workflow that operates on individual scenes rather than aggregated composites. The choice of method involves a trade-off along three axes: detection quality (does it correctly flag thin cirrus, cloud edges, and cast shadows?), computational cost (can it run at archive scale on available infrastructure?), and operational availability (is it pre-computed and shipped with the product, or must it be run separately?). No single method dominates on all three axes, and the appropriate choice depends on the use case: a global mapping pipeline running monthly composites has different requirements from a single-scene change-detection workflow where every false negative becomes a visible artifact.

Table~\ref{tab:s2-cloud-masking} compares the methods in current use. We distinguish the algorithmic family (rule-based, classical ML, or deep learning), the form of the output (binary mask versus continuous quality score), and the practical considerations that govern whether a given method is realistic to deploy at scale.

\begin{table}[!t]
\centering
\caption{Cloud and cloud-shadow masking methods for Sentinel-2 data.}
\label{tab:s2-cloud-masking}
\footnotesize
\renewcommand{\arraystretch}{1.2}
\begin{tabularx}{\columnwidth}{@{}L P{1.9cm} P{1.9cm} c@{}}
\toprule
\textbf{Method} & \textbf{Family} & \textbf{Output} & \textbf{\shortstack{Pre-\\computed}} \\
\midrule
SCL \citep{louis2016sentinel} & Rule-based & Multi-class label & \checkmark \\
Fmask \citep{QIU2019111205} & Physics-based and ML & Multi-class label & $\times$ \\
Cloud Score+ \citep{Pasquarella2023} & Deep learning & Continuous probability & \checkmark$^*$ \\
s2cloudless \citep{zupanc2017improving} & ML & Continuous probability & \checkmark$^{\dagger}$ \\
KappaMask \citep{Domnich2021} & Deep learning & Multi-class label & $\times$  \\
CloudSen12 \citep{Aybar2022} & Deep learning & Multi-class label & $\times$  \\
OmniCloudMask \citep{Wright2025} & Deep learning & Multi-class label & $\times$  \\
\bottomrule
\end{tabularx}
\vspace{1ex}
  \begin{minipage}{\columnwidth}
    \footnotesize
    \centering
    $^*$On Google Earth Engine.\quad $^{\dagger}$On Sentinel Hub.
  \end{minipage}
\end{table}

A few practical considerations are worth highlighting. The SCL layer is the path of least resistance: it ships with every L2A product, requires no additional compute, and is sufficient for the dominant use case of cleaning cloudy pixels before median compositing. Its known weaknesses~--- conservative cirrus detection, missed cloud edges, occasional confusion between bright surfaces and clouds~--- become problematic when working with single scenes or when the application is sensitive to false negatives in cloud detection. Fmask, originally developed for Landsat, offers a more physics-informed and well-validated alternative, but is slower than SCL and must be run separately. s2cloudless occupies the opposite end of the efficiency spectrum: it is lightweight, fast on CPU, and widely used in operational pipelines.

Cloud Score+ \citep{Pasquarella2023} occupies an unusual position because of its continuous output. Rather than committing to a binary mask, it returns a per-pixel quality score that the downstream user can threshold according to the sensitivity of the application. This is convenient when the same archive must serve multiple workflows with different tolerances. The principal limitation is operational: Cloud Score+ is exposed natively only through Google Earth Engine, so workflows running outside that ecosystem cannot use it directly. CloudSen12, trained on hand-labeled patches stratified across biomes, and OmniCloudMask address the same need with self-hostable models at the cost of having to run inference oneself.

The deep learning methods (KappaMask, CloudSen12, OmniCloudMask) generally outperform the rule-based and classical-ML alternatives on hand-labeled benchmarks, but the magnitude of the gain depends on the region, the scene composition, and the specific failure modes the benchmark is designed to expose. \citet{Wright2025} report computational efficiency for OmniCloudMask against several alternatives on an NVIDIA RTX 4090, providing a useful reference point for relative throughput on consumer-grade hardware. Whether the accuracy gain justifies the additional compute and infrastructure complexity is a judgment that should be made against the requirements of the specific mapping task: for global-scale composite-based workflows, the marginal benefit over SCL is often modest; for single-scene applications or studies in cloud-prone regions such as the humid tropics, it can be decisive.

\paragraph{Landsat} The methods in Table~\ref{tab:s2-cloud-masking} are tuned for Sentinel-2, but the lineage runs the other way: Fmask was developed for Landsat first, and its C implementation is what produces the operational Landsat mask. Collection~2 ships a bit-packed \texttt{QA\_PIXEL} band generated during Level-1 processing and carried unchanged into the Level-2 package --- a structural difference from Sentinel-2, where the SCL is an output of the Level-2 processor and therefore exists only at Level-2. Its design also sits between the two families in Table~\ref{tab:s2-cloud-masking}: alongside high-confidence binary flags for cloud, cloud shadow, cirrus, snow and water, it carries a dilated-cloud bit that provides a buffer around detected cloud, and two-bit confidence fields for cloud, cloud shadow, snow/ice and cirrus. A user can therefore threshold on confidence rather than accept a binary verdict, which is the flexibility that makes Cloud Score+ attractive for Sentinel-2, available here without leaving the operational product. Level-2 adds further layers: \texttt{SR\_QA\_AEROSOL} for Landsat~8 and~9, \texttt{SR\_CLOUD\_QA} for Landsat~4 to~7, and \texttt{ST\_QA}, which reports per-pixel surface temperature uncertainty in Kelvin. One trap is specific to long archives: the \texttt{QA\_PIXEL} bit assignments are not identical across Landsat generations, so a pipeline spanning Landsat~5 through~9 must decode the band per sensor rather than apply one bit mask throughout.

Finally, none of these methods is sensor-agnostic by default with the exception of OmniCloudMask. Methods originally developed for Sentinel-2 can be retrained on other optical sensors, but doing so requires either a labeled dataset for the target sensor or a sufficiently large labeled Sentinel-2 dataset together with a domain adaptation strategy. Practitioners working across multiple optical missions should weigh this carefully, as the cost of maintaining one cloud-masking pipeline per sensor can rapidly exceed the cost of the inference itself.

\section{Radiometric conventions for optical data}
\label{app:s2-radiometry}

Section~\ref{subsubsec:optical-processing} notes that reflectance scaling is a convention rather than a constant. This appendix gives the Sentinel-2 and Landsat cases in full, as a worked example of the kind of detail such a convention involves.

\paragraph{Sentinel-2}

Since the deployment of Sentinel-2 Processing Baseline (PB) 04.00 in January 2022, a radiometric offset is added to the digital number values of all bands. For Sentinel-2 L1C data, the conversion to reflectance is given by Equation \ref{eq:1}, where $\rho_\mathtt{TOA}$ is the top-of-atmosphere (TOA) reflectance, $DN$ is the digital number, $\mathtt{RADIO\_ADD\_OFFSET}$ is the radiometric offset ($-1000$ for all bands for PB$\geq04.00$, $0$ otherwise), and $\mathtt{QUANTIFICATION\_VALUE} = 10{,}000$. For Sentinel-2 L2A data, the conversion is given by Equation \ref{eq:2}, where $\rho_{BOA}$ is the bottom-of-atmosphere (BOA) reflectance, $\mathtt{BOA\_ADD\_OFFSET}$ is the corresponding offset ($-1000$ for all bands for PB$\geq04.00$, $0$ otherwise), and $\mathtt{BOA\_QUANTIFICATION\allowbreak\_VALUE} = 10{,}000$. All values can be retrieved from the product metadata files. The offset applies to any product with Processing Baseline $\geq$ 04.00, which should be determined from the \texttt{PROCESSING\_BASELINE} field in the metadata rather than the acquisition date: as part of the Collection-1 reprocessing campaign completed in 2024, ESA reprocessed the entire historical archive, so even early acquisitions now carry the offset convention. This procedure applies to products distributed by ESA (e.g.,\ through CDSE). Some third-party platforms (such as GEE's harmonized collections) apply the offset upstream, in which case it must not be applied a second time.

{\small
\begin{equation}
\label{eq:1}
    \rho_\mathtt{TOA}=\frac{\mathtt{DN}+\mathtt{RADIO\_ADD\_OFFSET}}{\mathtt{QUANTIFICATION\_VALUE}}
\end{equation}
}

{\small
\begin{equation}
\label{eq:2}
    \rho_\mathtt{BOA}=\frac{\mathtt{DN}+\mathtt{BOA\_ADD\_OFFSET}}{\mathtt{BOA\_QUANTIFICATION\_VALUE}}
\end{equation}
}

\vspace{1em}

While Equations \ref{eq:1}--\ref{eq:2} use a quantification value of $10{,}000$ as specified in the product metadata, ML pipelines sometimes substitute alternative divisors: \citet{kerner2025fields} use $3{,}000$ and \citet{pauls2025capturing,pauls2026echosat} apply band-specific divisors. These linear schemes target a bounded range such as $[0,1]$ or $[-1,1]$, typically with further clipping. A less conventional family of approaches instead applies nonlinear transformations to handle the long-tailed distribution of reflectance values. \citet{Brown_2022} log-transform reflectance values and remap percentiles of the result onto a sigmoid function, bounding the data to $(0,1)$ without hard truncation. Similarly, \citet{brown2025alphaearthfoundations,Pasquarella2023} compress the distribution as $x'=\log(x+1)/10$.

\paragraph{Landsat} Landsat Collection~2 applies the same idea with different numbers, and changed them between collections. Level-2 surface reflectance is stored as an unsigned 16-bit integer and converted by Equation~\ref{eq:l8sr}, with a scale factor of $0.0000275$ and an additive offset of $-0.2$; valid digital numbers run from $7{,}273$ to $43{,}636$. Surface temperature in the same package converts to Kelvin by Equation~\ref{eq:l8st}, with a scale factor of $0.00341802$ and an offset of $149.0$, over a valid range of $293$ to $65{,}535$.

{\small
\begin{equation}
\label{eq:l8sr}
    \rho = \mathtt{DN} \times 0.0000275 - 0.2
\end{equation}
}
 
{\small
\begin{equation}
\label{eq:l8st}
    T_{\mathrm{K}} = \mathtt{DN} \times 0.00341802 + 149.0
\end{equation}
}
 
\vspace{1em}
 
Collection~1 used a scale factor of $0.0001$ with no offset, so a pipeline written against Collection~1 and pointed at Collection~2 returns reflectance values about 3.6 times their true magnitude, shifted by a constant\footnote{\url{https://www.usgs.gov/faqs/why-are-fill-values-and-scaling-factors-landsat-collection-2-level-2-products-different-those}}. The fill value changed in the same transition, and the change deserves more attention than it usually receives: Collection~1 used $-9999$, which is conspicuous in any histogram and fatal to any arithmetic that reaches it, whereas Collection~2 uses $0$, which converts to a reflectance of $-0.2$ and passes through most pipelines looking like an unremarkable dark pixel. As with the Sentinel-2 offset, neither change raises an error; both silently shift the input distribution.

Conventions are not the only thing that changes beneath a pipeline. Errors are found and corrected too, sometimes long after acquisition. A radiometric error in Landsat~2 and~3 MSS Level-1 products, affecting more than 44{,}000 scenes, was identified in April 2025, with reprocessing under way through 2026\footnote{\url{https://www.usgs.gov/faqs/what-are-some-known-issues-users-might-find-landsat-data}}; separately, framing errors traced to faulty ancillary data in a subset of 2023--2024 Landsat~8 scenes prompted a reprocessing campaign announced in June 2025. Two users holding what is nominally the same scene can therefore hold different data, depending only on when they downloaded it. This is the practical argument for recording the processing version alongside every input, rather than the product identifier alone: the identifier is stable, and the pixels behind it are not.

\section{Combining Landsat and Sentinel-2}
\label{app:combining-missions}
 
Products from different missions are rarely directly comparable. This appendix gives the Landsat and Sentinel-2 case, which is the best-supported instance and the one most often encountered.
 
\paragraph{What a mature processing chain delivers} Landsat benefits from a longer-standing and more standardized processing pipeline than Sentinel-2, and the user-side burden is correspondingly lower. Collection~2 Level-2 Science Products provide atmospherically corrected surface reflectance together with surface temperature, delivered scene by scene in UTM projection. Every scene carries a bit-packed \texttt{QA\_PIXEL} band generated during Level-1 processing, so the same quality information travels with the Level-1 and Level-2 products alike. The Landsat Analysis Ready Data product standardizes the geometric framework further by tiling scenes onto a fixed grid in Albers Equal Area projection, removing the need to manage overlapping scene boundaries, though the underlying radiometric processing is identical to Collection~2 Level-2.
 
\paragraph{Harmonizing across Landsat generations} Researchers working with multi-decadal archives should be aware that spectral response functions differ across Landsat generations. Landsat~8 and~9 use narrower bands that avoid atmospheric absorption features present in the Landsat~5 and~7 bandpasses, producing systematic offsets that bias time-series analysis if left uncorrected. \citet{Roy2016b} published spectral bandpass adjustment coefficients that harmonize across sensors. BRDF normalization improves temporal consistency in the same way it does for Sentinel-2, and \citet{Roy2016a} developed the standard route for Landsat, applying MODIS BRDF spectral model parameters to normalize surface reflectance to a nadir view at a fixed solar zenith angle. Neither correction is applied operationally by USGS, so both remain the user's responsibility.

\paragraph{What the long archive costs} The depth that makes Landsat attractive is also where its defects accumulate. USGS maintains a catalog of known issues running to more than twenty distinct anomalies\footnote{\url{https://www.usgs.gov/landsat-missions/landsat-known-issues}}, the large majority of them affecting Landsat~1 through~7: banding and detector striping, coherent noise, memory effect, optical leaks, oversaturation. The single largest discontinuity is the failure of the Landsat~7 scan line corrector on 31 May 2003, after which roughly 22\% of every ETM+ scene is lost to wedge-shaped gaps. The remaining data are usable and gap-filled products exist, but a time series crossing that date changes character partway through. None of this is an argument against the long archive. It is an argument for establishing, before committing to a span of years, which anomalies fall inside it --- a check that costs an afternoon and is almost never performed.

\paragraph{Harmonizing across missions} The Harmonized Landsat Sentinel-2 (HLS) dataset \citep{Claverie2018} offers a convenient starting point for studies that combine both missions. HLS provides 30\,m surface reflectance from Landsat~8 and~9 and Sentinel-2A, 2B and 2C, with atmospheric correction, BRDF normalization, cloud masking and spectral bandpass adjustment applied consistently across all of them. Products are delivered on the Sentinel-2 MGRS tile grid, in tiles of 109.8\,km overlapping by 4.9\,km on each side; with five satellites contributing, the combined revisit is now under two days. For large-scale mapping, HLS removes most of the cross-sensor harmonization burden, at a cost that should be stated plainly: Sentinel-2's native 10\,m resolution is sacrificed in favor of cross-mission consistency, and every pixel is resampled.
 
The complementary direction is taken by sen2like\footnote{\url{https://github.com/senbox-org/sen2like}}, which harmonizes Landsat toward Sentinel-2 rather than the reverse, preserving the finer grid at the cost of upsampling the coarser sensor. Which direction is appropriate depends on whether spatial detail or archive length matters more for the task: harmonizing toward Landsat buys a record reaching back to the 1980s at 30\,m, while harmonizing toward Sentinel-2 keeps 10\,m detail over a decade.